\documentclass[sigconf]{acmart}
\usepackage{multirow}
\usepackage{colortbl}
\usepackage[capitalize,noabbrev]{cleveref}
\usepackage{subcaption}
\usepackage{mathtools}
\usepackage{enumitem}
\usepackage[most]{tcolorbox}
\usepackage{marvosym}
\usepackage{algorithm}
\usepackage{algorithmic}
\usepackage{placeins}

\AtBeginDocument{%
  }

\copyrightyear{2026}
\acmYear{2026}
\setcopyright{cc}
\setcctype{by}
\acmConference[MM '26]{Proceedings of the 34th ACM International Conference on Multimedia}{November 10--14, 2026}{Rio de Janeiro, Brazil}
\acmBooktitle{Proceedings of the 34th ACM International Conference on Multimedia (MM '26), November 10--14, 2026, Rio de Janeiro, Brazil}
\acmISBN{979-8-4007-2213-4/2026/11}
\acmDOI{10.1145/3767308.3836185}
\begin{document}

\title[ForgeryVCR: Visual-Centric Reasoning for Image Forgery Detection and Localization]{ForgeryVCR: Visual-Centric Reasoning via Efficient Forensic Tools in MLLMs for Image Forgery Detection and Localization}

\author{Youqi Wang}
\authornote{Equal contribution.}
\authornote{Work done during an internship at Tencent Youtu Lab.}
\orcid{0009-0009-6072-6755}
\affiliation{%
  \institution{Shenzhen University}
  \city{Shenzhen}
  \state{Guangdong}
  \country{China}}
\affiliation{%
  \institution{Tencent Youtu Lab}
  \city{Shanghai}
  \country{China}}
\email{2400042020@mails.szu.edu.cn}

\author{Shen Chen}
\authornotemark[1]
\authornote{Project leader.}
\orcid{0000-0001-9140-194X}
\affiliation{%
  \institution{Tencent Youtu Lab}
  \city{Shanghai}
  \country{China}}
\email{kobeschen@tencent.com}

\author{Haowei Wang}
\orcid{0009-0006-0289-9672}
\affiliation{%
  \institution{Tencent Youtu Lab}
  \city{Shanghai}
  \country{China}}
\email{asucawang@tencent.com}

\author{Rongxuan Peng}
\orcid{0009-0007-3739-652X}
\affiliation{%
  \institution{Shenzhen University}
  \city{Shenzhen}
  \state{Guangdong}
  \country{China}}
\email{2350433004@email.szu.edu.cn}

\author{Taiping Yao}
\orcid{0000-0002-2359-1523}
\affiliation{%
  \institution{Tencent Youtu Lab}
  \city{Shanghai}
  \country{China}}
\email{taipingyao@tencent.com}

\author{Shunquan Tan}
\correspondingauthor
\orcid{0000-0002-7457-3691}
\affiliation{%
  \institution{Shenzhen University}
  \city{Shenzhen}
  \state{Guangdong}
  \country{China}}
\email{tansq@szu.edu.cn}

\author{Changsheng Chen}
\orcid{0000-0003-4857-6810}
\affiliation{%
  \institution{Shenzhen University}
  \city{Shenzhen}
  \state{Guangdong}
  \country{China}}
\email{cschen@szu.edu.cn}

\author{Bin Li}
\orcid{0000-0002-2613-5451}
\affiliation{%
  \institution{Shenzhen University}
  \city{Shenzhen}
  \state{Guangdong}
  \country{China}}
\email{libin@szu.edu.cn}

\author{Shouhong Ding}
\correspondingauthor
\orcid{0000-0002-3175-3553}
\affiliation{%
  \institution{Tencent Youtu Lab}
  \city{Shanghai}
  \country{China}}
\email{ericshding@tencent.com}

\renewcommand{\shortauthors}{Wang et al.}

\begin{abstract}
  Existing Multimodal Large Language Models (MLLMs) for image forgery detection and localization predominantly operate under a text-centric Chain-of-Thought (CoT) paradigm. However, forcing these models to textually characterize imperceptible low-level tampering traces inevitably leads to hallucinations, as linguistic modalities are insufficient to capture such fine-grained pixel-level inconsistencies. To overcome this, we propose ForgeryVCR, a framework that incorporates a forensic toolbox to materialize imperceptible traces into explicit visual intermediates via Visual-Centric Reasoning. To enable efficient tool utilization, we introduce a Strategic Tool Learning post-training paradigm, encompassing gain-driven trajectory construction for Supervised Fine-Tuning (SFT) and subsequent Reinforcement Learning (RL) optimization guided by a tool utility reward. This paradigm empowers the MLLM to act as a proactive decision-maker, learning to spontaneously invoke multi-view reasoning paths including local zoom-in for fine-grained inspection and the analysis of invisible inconsistencies in compression history, noise residuals, and frequency domains. Extensive experiments reveal that ForgeryVCR achieves state-of-the-art (SOTA) performance in both detection and localization tasks, demonstrating superior generalization and robustness with minimal tool redundancy.
  The code is available at https://github.com/youqiwong/ForgeryVCR.
\end{abstract}


\maketitle

\section{Introduction}

\begin{figure}[t]
  \centering
  \includegraphics[width=\linewidth]{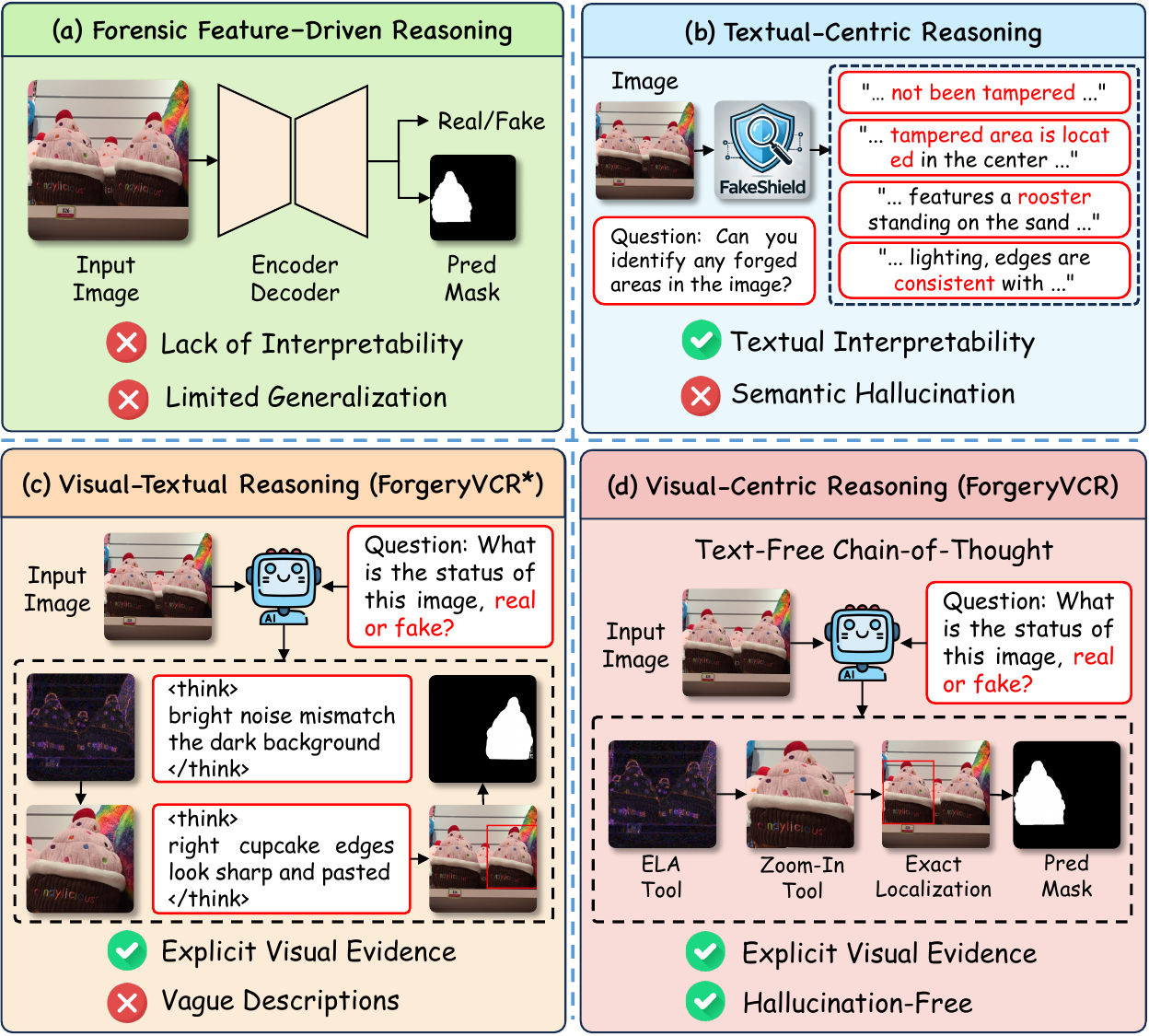}
  \caption{\textbf{Motivation.} Unlike semantically biased methods, ForgeryVCR employs \textbf{Visual-Centric Reasoning}, grounding the verdict in visual evidence rather than vague descriptions.}
  \Description{Motivation figure comparing prior methods with ForgeryVCR's Visual-Centric Reasoning approach for image forgery detection.}
  \label{fig:motivation}
\end{figure}

The rapid evolution of editing tools has created a deluge of hyper-realistic forgeries, establishing Image Forgery Detection and Localization (IFDL) as a crucial defense. Deep learning approaches predominantly capture subtle traces via frequency volumetric representations~\cite{Kwon2022CATNetIJCV}, noise degradation modeling~\cite{Wu2022IF-OSN}, or semantic-noise decoupling~\cite{Dong2023MVSSNetTPAMI, Han2024HDFNetTPAMI}. Advanced methods further employ learned noise fingerprints~\cite{Guillaro2023TruForCVPR}, sequential decision-making~\cite{Peng2024CoDETIFS}, masked attention~\cite{Kong2025PIMTPAMI}, or adapted vision foundation models~\cite{Kwon2025SAFIREAAAI, Peng2025ForensicsSAM}. However, these specialized networks operate as black boxes lacking interpretability and generalization, as illustrated in Fig.~\ref{fig:motivation}(a).

To bridge interpretability and generalization gaps, recent efforts have shifted toward Multimodal Large Language Models (MLLMs). Forensic adaptations such as ForgeryGPT~\cite{Liu2025ForgeryGPT}, FakeShield~\cite{Xu2025FakeShield}, and SIDA~\cite{Huang2025SIDA} extend MLLMs to image forgery analysis by exploiting high-level semantic knowledge. However, these approaches face two critical limitations: 1) \textbf{Semantic Hallucinations.} Existing methods rely on a text-centric Chain-of-Thought (CoT) paradigm, yet linguistic modalities inherently suffer from information loss. As visualized in Fig.~\ref{fig:motivation}(b), linguistic descriptions fail to accurately align with specific tampering clues, leading to hallucinations where irrelevant scene context is misidentified as forgery evidence. 2) \textbf{Insensitivity to Low-Level Artifacts.} While low-level features have proven effective in traditional solutions, current MLLM architectures focus solely on high-level semantic extraction. This inherent bias renders them insensitive to low-level forensic information, limiting their ability to perceive subtle manipulation traces.

To mitigate the discrepancy between high-level semantics and low-level forensic artifacts, we introduce \textbf{ForgeryVCR}, which reformulates the MLLM as a proactive decision-maker that plans, executes, and revises forensic analysis. We equip it with a Hybrid Forensics Toolbox via lightweight code execution, combining zoom-in inspection with specialized spatial-, frequency-, and noise-domain operators to materialize imperceptible inconsistencies as explicit visual intermediates. As illustrated in Fig.~\ref{fig:motivation}(c), vague textual interpretations of forensic patterns can misguide decision-making. ForgeryVCR therefore adopts the Visual-Centric Reasoning paradigm shown in Fig.~\ref{fig:motivation}(d), deriving the final decision directly from visual intermediates to reduce interference from ambiguous descriptions and improve generalization in detection and localization.

Our main contributions are summarized as follows:
\begin{itemize}
  \item We introduce ForgeryVCR, the first framework in the IFDL domain to implement \textbf{Visual-Centric Reasoning} via forensic analysis tools. Our work reveals that this paradigm significantly outperforms models relying on textual explanations in capturing subtle manipulation traces, effectively mitigating hallucinations caused by semantic over-reliance.
  \item We propose an efficient tool utilization strategy to capture low-level artifacts essential for forensics. By employing \textbf{gain-driven tool selection} during SFT to construct effective reasoning trajectories, followed by RL optimization with a \textbf{tool utility reward}, we empower the MLLM to spontaneously invoke tools only when they yield decisive visual evidence.
  \item Extensive experiments demonstrate that ForgeryVCR achieves SOTA detection and localization performance with superior generalization across diverse benchmarks.
\end{itemize}

\begin{figure*}[t!]
  \centering
  \includegraphics[width=\textwidth]{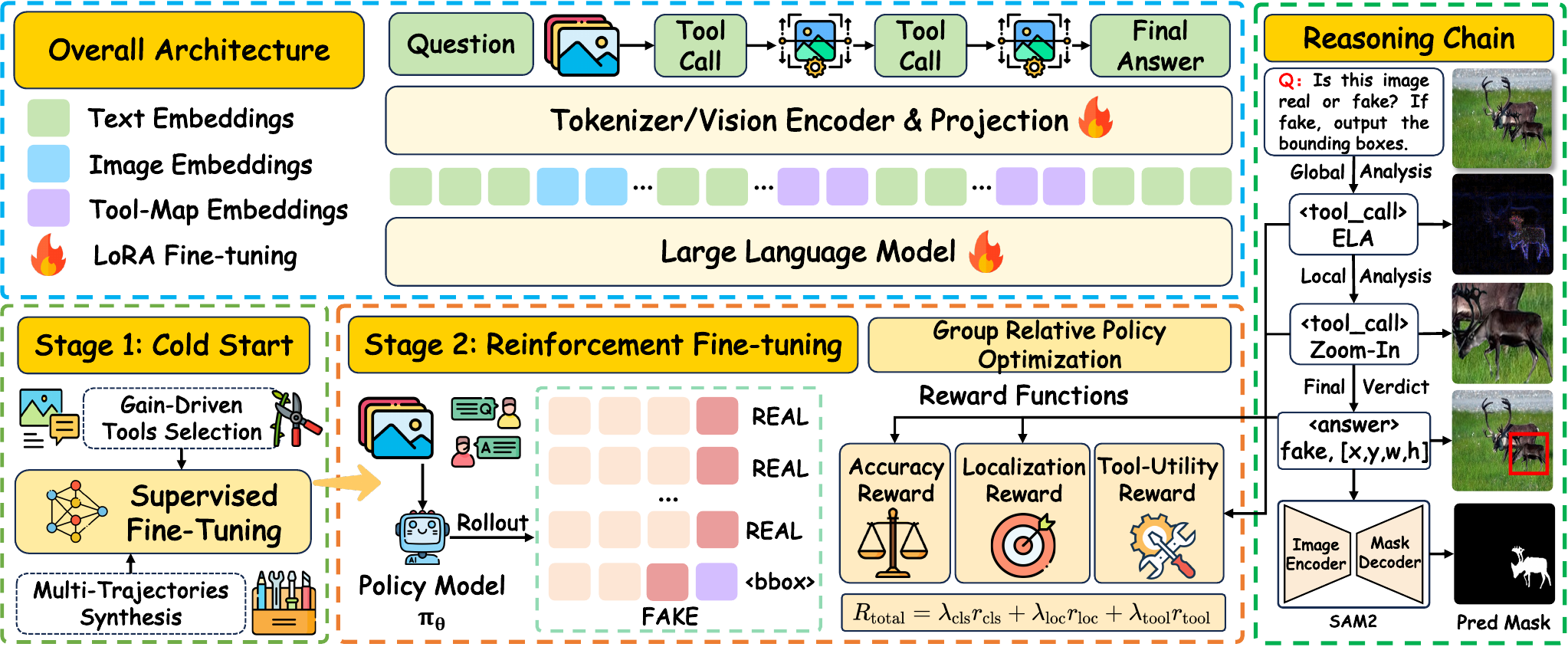}
  \caption{\textbf{Overview of the ForgeryVCR Framework.} The top panel depicts the architecture. The training pipeline: (1) Stage 1 uses Gain-Driven Tool Selection and Multi-Trajectories Synthesis to construct diverse reasoning paths; (2) Stage 2 optimizes the policy via GRPO with Tool-Utility Reward to foster strategic tool usage. The right panel shows the reasoning chain invoking tools to expose subtle artifacts for precise localization, guiding SAM2 to generate the fine-grained mask.}
  \Description{Overview diagram of the ForgeryVCR framework showing the architecture and two-stage training pipeline.}
  \label{fig:training_pipeline}
\end{figure*}

\section{Related Work}
\subsection{Image Forgery Detection and Localization}

Existing approaches for Image Forgery Detection and Localization can be broadly categorized into deep learning-based specialist networks and Multimodal Large Language Model (MLLM) adaptations.

\textbf{Deep Learning-based Approaches.} Traditional deep forensic networks operate as specialized experts designed to scrutinize low-level statistical anomalies. To capture invisible traces, methods like CAT-Net~\cite{Kwon2022CATNetIJCV} and IF-OSN~\cite{Wu2022IF-OSN} utilize dual-stream architectures or noise-aware schemes to analyze frequency domains and noise residuals. Moving towards fusing semantics with physical traces, TruFor~\cite{Guillaro2023TruForCVPR} and MVSS-Net~\cite{Dong2023MVSSNetTPAMI} combine learned noise fingerprints with edge supervision to model boundary discontinuities. Recent advancements further refine localization precision by synergizing RGB features with steganalysis models (HDF-Net~\cite{Han2024HDFNetTPAMI}), employing sequential decision-making (CoDE~\cite{Peng2024CoDETIFS}), or utilizing advanced segmentation designs (PIM~\cite{Kong2025PIMTPAMI}, SAFIRE~\cite{Kwon2025SAFIREAAAI}).

\textbf{MLLM-based Approaches.} To address the interpretability and generalization issues of specialized networks, recent research has pivoted towards adapting MLLMs for forensics. ForgeryGPT~\cite{Liu2025ForgeryGPT} employs a three-stage training paradigm, introducing a mask-aware forgery extractor to explicitly align fine-grained tampering traces with linguistic features. Similarly, FakeShield~\cite{Xu2025FakeShield} utilizes a domain-tag generator and a tamper comprehension module to bridge the gap between semantic understanding and pixel-level localization for explainable diagnosis. SIDA~\cite{Huang2025SIDA} extends MLLM capabilities to social media deepfakes by incorporating specialized detection and segmentation tokens to unify authenticity assessment with mask generation. To address the lack of fine-grained perception in MLLMs, ProposeAndRectify~\cite{Zhang2025ProposeAndRectify} explicitly introduces low-level forensic information into the reasoning loop, employing a rectification module to validate semantic proposals using multi-scale physical artifacts.

\textbf{Limitations.} Despite their successes, both paradigms have inherent limitations. Specialized deep networks operate as black boxes and generalize poorly to novel attacks because they rely on fixed feature distributions. MLLM-based approaches offer better explainability but focus heavily on high-level semantics, which can lead to context-driven hallucinations and poor grounding in subtle low-level forensic artifacts.

\subsection{Visual Chain-of-Thought}

The paradigm of Chain-of-Thought (CoT) has recently evolved from textual reasoning to the multimodal domain, aiming to bridge the gap between visual perception and logical deduction. Recent studies unifiedly interpret this as integrating intermediate visual representations into the reasoning chain~\cite{Cheng2025VisualThoughts}. In contrast to text-centric paradigms which rely solely on linguistic rationales, visual-interleaved strategies explicitly generate or retrieve visual evidence to ground the reasoning process.

Active-perception methods broadly follow two directions. Chain-of-Focus (CoF)~\cite{Zhang2025ChainOfFocus} and DeepEyes~\cite{Zheng2025DeepEyes} perform localized visual search through iterative zooming, whereas VCTP~\cite{Chen2024VisualCoT}, VACoT~\cite{Xu2025VACoT}, and DeepEyesV2~\cite{Hong2025DeepEyesV2} extend reasoning with visual verification, image operations, or external tools, respectively.

However, adapting these frameworks to IFDL presents unique challenges. Even with general visual processing tools, MLLM visual encoders remain blind to imperceptible forensic artifacts beyond dominant semantic content, and MLLMs lack the capability to strategically invoke appropriate tools. To bridge this gap, ForgeryVCR repurposes the Visual-CoT paradigm for the forensic domain by integrating specialized tools that expose low-level visual patterns and employing a gain-driven mechanism for efficient tool invocation, grounding the model's judgment in explicit visual evidence rather than semantic biases.

\section{Methodology}
\label{sec:method}

We begin by outlining the overall architecture and the investigative workflow in Section~\ref{subsec:architecture}. Subsequently, Section~\ref{subsec:toolbox} details the hybrid forensics toolbox utilized to expose imperceptible artifacts. Section~\ref{subsec:data_generation_pipeline} introduces the trajectories synthesis pipeline, incorporating Gain-Driven Tool Selection and Multi-Trajectories Diversification. Finally, Section~\ref{subsec:two_stage_training_pipeline} describes the Strategic Tool Learning Pipeline, comprising SFT and RL via Group Relative Policy Optimization (GRPO)~\cite{Guo2025DeepSeek-R1} to foster strategic tool usage.

\subsection{Architecture}
\label{subsec:architecture}

ForgeryVCR establishes a Visual-Centric Reasoning architecture. In this paradigm, the MLLM functions as a proactive agent that constructs a reasoning chain grounded in explicit forensic mappings. Crucially, this design bypasses the reliance on linguistic descriptors, enabling the model to directly perceive and process the generated visual outputs to expose hidden manipulation artifacts.

As illustrated in Fig.~\ref{fig:training_pipeline}, the inference workflow begins with a preliminary visual assessment. When the initial evidence is ambiguous, the model dynamically invokes tools from the hybrid forensics toolbox. These tools materialize latent artifacts into discernible visual cues, which are then re-injected into the context window to ground the reasoning process. Upon concluding the investigation with a forgery verdict, the model predicts a bounding box to localize the manipulated region. Finally, to ensure pixel-level precision, we feed this bounding box as a visual prompt into the Segment Anything Model 2 (SAM2)~\cite{Ravi2024SAM2}, generating a high-quality segmentation mask.

\subsection{Hybrid Forensics Toolbox}
\label{subsec:toolbox}
To bridge the perceptual gap in MLLMs for image forgery analysis, we introduce the Hybrid Forensics Toolbox designed to materialize imperceptible artifacts into explicit visual evidence. Although numerous forensic tools exist across spatial, frequency, and generative domains, integrating them indiscriminately proves suboptimal. Our analysis indicates that significant feature redundancy exists among these tools, where stacking excessive operators leads to performance saturation rather than linear improvement. As evidenced by the quantitative analysis in Appendix, adding redundant operators introduces unnecessary complexity to the tool selection policy without yielding tangible benefits. Therefore, we strategically select a sufficient set of tools that exhibit strong complementarity and necessity. This ensures comprehensive capture of diverse manipulation traces while maintaining an efficient decision space.

\paragraph{Forensic Analysis Tools.}
We integrate three representative forensic algorithms to convert invisible statistical inconsistencies into distinct visual maps. Since raw tool outputs encode low-level information in compression, noise, or frequency domains that are difficult for MLLM visual encoders to interpret directly, we apply RGB-space enhancement to remap these maps into perceptually accessible representations while preserving their forensic content. This enables the MLLM to acquire basic perceptual ability over tool outputs, establishing an effective foundation for subsequent post-training.
(1) \textbf{Error Level Analysis (ELA):} This tool ~\cite{Lu2025MFDFIMLCSCWD_ELA} visualizes discrepancies in compression levels. It highlights regions with different compression histories as distinct high-response areas compared to the background.
(2) \textbf{Noise Print++ (NPP):} This tool ~\cite{Guillaro2023TruForCVPR} extracts camera model fingerprints by analyzing noise residuals. It effectively reveals manipulation traces where the original sensor noise pattern is disrupted or inconsistent.
(3) \textbf{Fast Fourier Transform (FFT):} This tool ~\cite{Kashiani2025FreqDebiasCVPR_FFT} transforms the spatial signal into the frequency spectrum. It is particularly effective at exposing grid-like artifacts and spectral anomalies introduced by resampling and interpolation, providing complementary frequency-domain clues distinct from spatial ones.

\paragraph{Visual Refinement Tool.}
Identifying tampering artifacts that occupy only a small proportion of the image requires fine-grained visual perception. However, under global scale perception, such subtle traces are often overwhelmed by global context, making them difficult to distinguish.
To address this challenge, we introduce a Zoom-In tool for coarse-to-fine inspection. After global forensic analysis, the MLLM identifies a suspicious region and the tool directly crops the raw patch without any interpolation or resizing as a new visual input. This preserves original pixel-level details, enabling focused local perception for subsequent prediction.

\subsection{Visual-Centric Trajectory Construction}
\label{subsec:data_generation_pipeline}

\begin{figure}[t]
  \centering
  \includegraphics[width=0.98\linewidth]{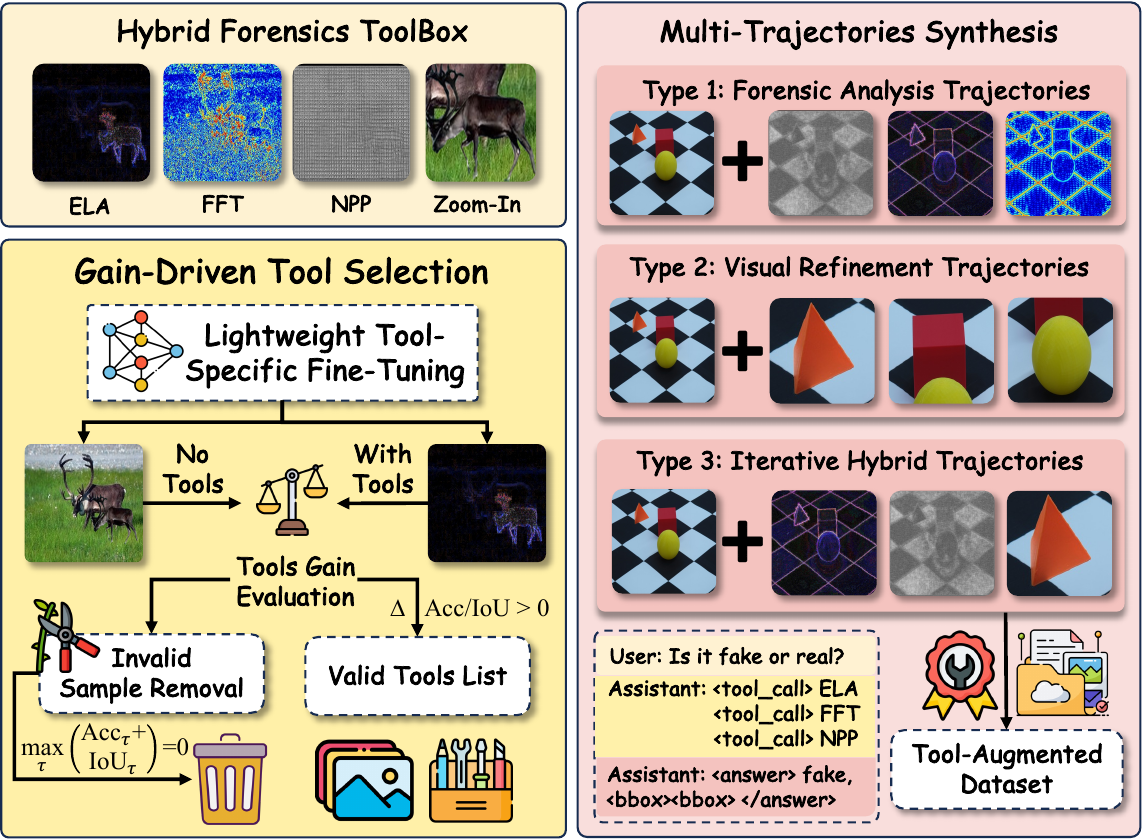}
  \caption{\textbf{Pipeline of Visual-Centric Trajectory Construction.}}
  \Description{Pipeline diagram showing the visual-centric trajectory construction process.}
  \label{fig:cot_pipeline}

\end{figure}

Constructing high-quality reasoning trajectories is pivotal for aligning semantic understanding with forensic execution. Indiscriminate training on all available tools often leads to redundancy and inefficient tools dependency. To address this, we introduce a trajectory synthesis pipeline, as illustrated in Fig.~\ref{fig:cot_pipeline}, which rigorously filters effective tools and generates diverse reasoning paths. This design ensures the model invokes forensic analysis and visual refinement tools only when they yield tangible information gains.

\paragraph{Gain-Driven Tool Selection.}
To mitigate redundancy induced by indiscriminate tool usage, we establish a mechanism to identify forensic operations yielding tangible gain. We begin with Lightweight Tool-Specific Fine-Tuning on the Cold-Start training set (CASIA v2) for 5 epochs, where a tool-free baseline and independent single-tool experts (ELA, NPP, FFT, and Zoom-In) are each trained to directly predict classification and localization bounding boxes without reasoning. During fine-tuning, the LLM is frozen while only the ViT and aligner remain trainable (details in the Appendix). Specifically, the tool-free baseline takes the original image as input, while each single-tool expert receives the original image concatenated with the corresponding forensic tools. All specialist models are then evaluated on the same Cold-Start set to obtain per-sample performance metrics that guide subsequent gain-driven tool selection.

Based on the inference results, we first execute Intractable Sample Removal (ISR), filtering out samples where all models (including the baseline) fail to yield correct classification or valid localization, thereby retaining only diagnosable instances. The Cold-Start set (CASIA v2) contains 5,123 balanced authentic and tampered images. After ISR, 4,961 authentic and 3,987 tampered images are retained.

For the remaining samples, we construct a sample-specific gain-driven set $T_{valid}$ by identifying candidate tools that outperform the tool-free baseline. We quantify the gain using a metric $P$ conditioned on the sample type: localization IoU for manipulated images, and the normalized probability of the ground-truth class token for authentic ones. With the baseline performance $P_{base}$ and a validity threshold $\tau = 0.5$, the selection criterion is:
\begin{equation}
T_{valid} = \{ t \in T_{pool} \mid P_t > \max(P_{base}, \tau) \}
\end{equation}
Here $\tau = 0.5$ serves as a natural quality gate: for tampered samples, an IoU below 0.5 indicates inadequate localization; for authentic samples, a classification probability below 0.5 implies uncertain prediction. Tools must exceed both this intrinsic quality floor and the baseline performance to be retained.
To enable gain-prioritized reasoning, we derive the final tool chain $T_{rank} = [t_1, \dots, t_M]$ by sorting the selected tools in descending order of their performance gain. If no tool surpasses the baseline ($T_{valid} = \emptyset$), $T_{rank}$ remains empty, directing the model to perform direct visual assessment without tool invocation.

\paragraph{Multi-Trajectories Synthesis.}
Relying solely on a fixed trajectory prevents the MLLM from adaptively adjusting its tool invocation strategy based on sample complexity, often leading to the loss of direct visual assessment capabilities. To overcome this limitation, we employ a \textit{Multi-Trajectory Synthesis (MTS)} strategy based on the ranked sequence $T_{rank}$. To ensure data balance during SFT, we truncate $T_{rank}$ to the top-$K$ candidates. We then construct the augmented reasoning set $P_{aug}$ by unifying direct judgment, independent checks, and accumulative evidence:
\begin{equation}
    P_{aug} = \underbrace{\{\emptyset\}}_{\text{No-Tool}} \cup \underbrace{\big\{ [t_i] \big\}_{i=1}^{K'}}_{\text{Single-Tool}} \cup \underbrace{\big\{ [t_1, \dots, t_k] \big\}_{k=2}^{K'}}_{\text{Multi-Tools}}
\end{equation}
where $K' = \min(|T_{rank}|, K)$. We apply a dynamic $K$ to balance the SFT data: $K{=}4$ for tampered samples and $K{=}2$ for authentic ones, since the lightweight specialists inherit the MLLM's intrinsic perception on real images, making large tool combinations redundant for authentic samples. After MTS, the final SFT dataset comprises 16,645 authentic and 17,408 tampered reasoning trajectories.

\subsection{Strategic Tool Learning Pipeline}
\label{subsec:two_stage_training_pipeline}

\paragraph{SFT-based Cold Start.}
To instantiate the visual-centric reasoning capability, we perform SFT using the structured trajectory dataset $P_{aug}$ constructed in Sec.~\ref{subsec:data_generation_pipeline}. The primary objective of this stage is to instill the fundamental protocol of Visual-CoT, aligning the model with the correct syntax for tool invocation and the sequential logic of visual evidence accumulation. We model this as a standard autoregressive generation task minimizing the cross-entropy loss:
\begin{equation}
    \mathcal{L}_{SFT} = -\mathbb{E}_{(V, X, Y) \sim P_{aug}} \sum_{t=1}^{T} \log \pi_\theta(y_t \mid V, X, y_{<t})
\end{equation}
where $\pi_\theta$ is the policy, $(V, X, Y)$ represent the visual input, textual instruction, and target response, and $T$ is the sequence length.

\paragraph{Reinforcement Learning Optimization.}
While SFT ensures structural formatting, the MLLM often lacks adaptive tool selection and genuine perceptual grounding. To bridge this gap, we employ RL via GRPO to optimize the policy. For each query $q$, we sample a group of $G$ outputs $\{o_i\}_{i=1}^G$ from the old policy $\pi_{\theta_{old}}$. The objective maximizes the advantage-weighted likelihood while constraining deviation from the reference policy $\pi_{ref}$:
\begin{equation}
    \mathcal{J}(\theta) = \mathbb{E}_{q} \left[ \frac{1}{G} \sum_{i=1}^G \left( \mathcal{L}_{clip}(o_i, A_i) - \beta \mathbb{D}_{KL}(\pi_\theta || \pi_{ref}) \right) \right]
\end{equation}
where $\mathcal{L}_{clip}$ denotes the clipped surrogate loss and $\beta$ controls the KL penalty. Crucially, GRPO computes the advantage $A_i$ by normalizing the total reward $R_i$ within the group: $A_i = (R_i - \text{mean}(R)) / \text{std}(R)$. The reward $R_i$ employs a composite function to guide forensic rigorousness, comprising three components: 1) \textbf{Classification Reward} ($R_{cls}$): We reward correct classification of the image. 2) \textbf{Localization Reward ($R_{loc}$)}: We reward precise localization using the Intersection-over-Union (IoU) between predicted and ground-truth bounding boxes. 3) \textbf{ Tool Utility Reward ($R_{tool}$)}: we foster efficient execution by granting rewards only when tool invocation leads to correct classification or accurate localization.

Let $c$, $\hat{c}$ denote the ground-truth and predicted labels, $\mathcal{B}_{gt}$, $\hat{\mathcal{B}}$ the ground-truth and predicted bounding box sets, and $u_t \in \{0,1\}$ indicate whether any tool was invoked. The reward components are defined as:
\begin{equation}
    R_{cls} = \mathbb{I}(\hat{c} = c) \cdot \begin{cases}
    1, & \text{if } c = \text{real} \\
    \mathbb{I}(\hat{\mathcal{B}} \neq \emptyset), & \text{if } c = \text{fake}
    \end{cases}
\end{equation}
For manipulated samples, $R_{cls}$ requires at least one valid bounding box to prevent hallucinated ``fake'' predictions without localization evidence. The localization reward uses Hungarian matching to find the optimal one-to-one assignment between predicted and ground-truth boxes, computing the mean IoU of matched pairs:
\begin{equation}
    R_{loc} = \text{HungarianIoU}(\hat{\mathcal{B}}, \mathcal{B}_{gt}) \cdot \mathbb{I}(c = \text{fake})
\end{equation}
The tool utility reward grants credit for tool invocation only when it leads to a correct verdict (for real images) or accurate localization with $\text{HungarianIoU} > \tau_{\text{IoU}}$ (for manipulated images, $\tau_{\text{IoU}} = 0.5$):
\begin{equation}
  R_{tool} = u_t \left( \mathbb{I}_{c=\text{real}}\, R_{cls} + \mathbb{I}_{c=\text{fake}}\, \mathbb{I}_{\text{HungarianIoU} > \tau_{\text{IoU}}} \right)
\end{equation}
The total reward is $R_{total} = \lambda_{cls} R_{cls} + \lambda_{loc} R_{loc} + \lambda_{tool} R_{tool}$, where $\{\lambda_{cls}, \lambda_{loc}, \lambda_{tool}\} = \{1.0, 2.0, 0.5\}$. Ablations on these coefficients are provided in the Appendix.

\section{Experiments}
\label{sec:experiments}

\subsection{Experimental Setup}

\paragraph{Baselines.}
To evaluate the efficacy of ForgeryVCR, we compare against three categories: 1) SOTA Traditional IFDL Methods. We select representative models focusing on low-level features, including MVSS-Net~\cite{Dong2023MVSSNetTPAMI}, TruFor~\cite{Guillaro2023TruForCVPR}, CoDE~\cite{Peng2024CoDETIFS}, HDF-Net~\cite{Han2024HDFNetTPAMI}, PIM~\cite{Kong2025PIMTPAMI}, and SAFIRE~\cite{Kwon2025SAFIREAAAI}. 2) Forensic-tuned MLLM methods. We compare against emerging MLLMs adapted for forensics, namely FakeShield \cite{Xu2025FakeShield} and SIDA \cite{Huang2025SIDA}. For fair comparison, all selected methods are retrained on the same training set as ours using their officially released code, with hyperparameters following their original papers. 3) General-purpose MLLMs. As detailed in Section~\ref{sec:mllm_comparison}, we also include open-source MLLMs fine-tuned on the same protocol.

\paragraph{Datasets.}
Our experimental setup adopts distinct datasets for training and evaluation. The training procedure follows a two-stage strategy, including SFT and RL. During SFT, we use the CASIA v2 dataset~\cite{Dong2013CASIADatabase} with balanced authentic and tampered samples. In the RL stage, we further incorporate IMD2020~\cite{Novozamsky2020IMD2020} and a subset of FantasticReality~\cite{Kniaz2019PointFantasticReality}. 

We evaluate generalization on benchmarks including CASIA v1~\cite{Dong2013CASIADatabase}, Columbia~\cite{Hsu2006ColumbiaICME}, Coverage~\cite{Wen2016COVERAGEICIP}, CocoGlide~\cite{Guillaro2023TruForCVPR}, DSO~\cite{DeCarvalho2013DSOTIFS}, Korus~\cite{Korus2017TIFS}, In-the-wild~\cite{Huh2018InthewildECCV}, NIST16~\cite{Guan2019NIST16WACVW}, and SID-Set~\cite{Huang2025SIDA}, comprising 25{,}588 samples in total. These datasets cover diverse manipulation types, including copy-move, splicing, object removal, and AIGC-based inpainting. Detailed statistics for each benchmark are provided in the Appendix.

\paragraph{Implementation Details.}
Our model is built on Qwen3-VL-4B-Instruct~\cite{Bai2025Qwen3} as the backbone MLLM. During the Cold-Start stage, we apply LoRA fine-tuning to the ViT, Aligner, and LLM layers with rank/alpha of 32/64, training for 2{,}700 steps. In the RL stage, only the LLM layers are fine-tuned with rank/alpha of 128/256, training for 776 steps. For ForgeryVCR$^*$, the SFT textual CoT annotations are generated by Qwen2.5-VL-72B-Instruct~\cite{Bai2025Qwen25VL}. Further details and ablation configurations are provided in the Appendix.

\paragraph{Evaluation Metrics.}
To ensure a fair and standardized comparison, we strictly adhere to the evaluation protocols established in FakeShield~\cite{Xu2025FakeShield}. For image-level forgery detection, we report the standard Accuracy (ACC) and F1-score. For traditional forensic networks that generate continuous probability maps, we apply a fixed threshold of 0.5 to binarize, classifying an image as tampered if it contains any positive pixels. For pixel-level localization, we employ the F1-score and Intersection over Union (IoU). We additionally report B-IoU (BBox-IoU), computed as in Eq.~(6), to assess region-level localization without SAM2, which also isolates the segmentation module's contribution. For specialist methods that output probability masks, we apply morphological closing and small-region filtering, then extract and merge connected components into bounding boxes before computing B-IoU.

\begin{table*}[tb!]
  \centering
  \small
  \caption{Quantitative comparison of image-level forgery detection performance (F1 and Accuracy). Methods are separated by category (Specialist vs. MLLMs methods). Best and second-best results are marked in \textbf{bold} and \underline{underlined}, respectively. The method marked with * denotes ForgeryVCR evaluated with visual and textual CoT. The column ``Weighted Avg.'' denotes the weighted average performance.}
  \label{tab:detection_comparison}

  \setlength{\tabcolsep}{1.5pt}

  \resizebox{\textwidth}{!}{
  \begin{tabular}{@{}c cc cc cc cc cc cc cc cc cc cc@{}}
      \toprule
      \multirow{2}{*}{Method} & \multicolumn{2}{c}{CASIA v1} & \multicolumn{2}{c}{Coverage} & \multicolumn{2}{c}{CocoGlide} & \multicolumn{2}{c}{NIST16} & \multicolumn{2}{c}{Korus} & \multicolumn{2}{c}{DSO} & \multicolumn{2}{c}{Columbia} & \multicolumn{2}{c}{In-the-wild} & \multicolumn{2}{c}{SID-Set} & \multicolumn{2}{c}{Weighted Avg.} \\
      \cmidrule(lr){2-3} \cmidrule(lr){4-5} \cmidrule(lr){6-7} \cmidrule(lr){8-9} \cmidrule(lr){10-11} \cmidrule(lr){12-13} \cmidrule(lr){14-15} \cmidrule(lr){16-17} \cmidrule(lr){18-19} \cmidrule(lr){20-21}
       & F1 & ACC & F1 & ACC & F1 & ACC & F1 & ACC & F1 & ACC & F1 & ACC & F1 & ACC & F1 & ACC & F1 & ACC & F1 & ACC \\
      \midrule
      MVSS-Net (\textit{TPAMI'22}) & 0.7323 & 0.7599 & 0.6738 & 0.5577 & 0.6657 & 0.5605 & 0.5514 & 0.4166 & 0.6310 & 0.5379 & 0.6405 & 0.4712 & 0.7541 & 0.6767 & 0.9829 & 0.9663 & 0.6257 & 0.5101 & 0.6349 & 0.5302 \\
      TruFor (\textit{CVPR'23})   & 0.7888 & 0.7552 & 0.6269 & 0.6394 & 0.5637 & 0.6523 & 0.5808 & 0.5342 & 0.6622 & \underline{0.6629} & \underline{0.7279} & 0.6298 & \underline{0.9211} & \underline{0.9151} & 0.9521 & 0.9087 & 0.7773 & 0.7681 & 0.7611 & 0.7488 \\
      CoDE (\textit{TIFS'24})     & 0.7046 & 0.5535 & 0.6688 & 0.5048 & 0.6675 & 0.5049 & 0.5665 & 0.4027 & \underline{0.6667} & 0.5067 & 0.6667 & 0.5000 & 0.6679 & 0.5068 & \textbf{1.0000} & \textbf{1.0000} & 0.6604 & 0.4990 & 0.6627 & 0.5018 \\
      HDF-Net (\textit{TPAMI'24}) & 0.6963 & 0.5360 & 0.6645 & 0.5000 & 0.6662 & 0.5000 & 0.5550 & 0.4139 & 0.6465 & 0.5045 & 0.6775 & 0.5240 & 0.7002 & 0.5753 & 0.9976 & 0.9952 & 0.6632 & 0.5011 & 0.6641 & 0.5037 \\
      PIM (\textit{TPAMI'25})     & 0.6208 & 0.7017 & 0.5200 & 0.5385 & 0.5248 & 0.5791 & 0.4650 & 0.6176 & 0.4324 & 0.5781 & 0.5236 & 0.5625 & \textbf{0.9217} & \textbf{0.9180} & 0.7942 & 0.6587 & 0.6173 & 0.6400 & 0.6106 & 0.6421 \\
      SAFIRE (\textit{AAAI'25})   & 0.6972 & 0.5355 & 0.6667 & 0.5000 & 0.6667 & 0.5000 & 0.5622 & 0.3910 & \textbf{0.6677} & 0.5022 & 0.6667 & 0.5000 & 0.6630 & 0.4959 & \textbf{1.0000} & \textbf{1.0000} & 0.6664 & 0.5000 & 0.6665 & 0.5002 \\
      \midrule
      FakeShield (\textit{ICLR'25}) & 0.7193 & 0.7169 & 0.4485 & 0.5450 & 0.6190 & 0.6562 & 0.5380 & 0.6456 & 0.5942 & 0.4227 & 0.4605 & 0.5900 & 0.7678 & 0.7300 & 0.9046 & 0.8259 & 0.4088 & 0.5556 & 0.4570 & 0.5794 \\
      SIDA (\textit{CVPR'25})       & 0.6381 & 0.7099 & 0.1524 & 0.5550 & 0.4636 & 0.6475 & 0.2432 & 0.6715 & 0.2080 & 0.5500 & 0.2909 & 0.6100 & 0.8867 & 0.9028 & 0.7500 & 0.6000 & 0.1798 & 0.5436 & 0.2410 & 0.5723 \\
      \midrule
      ForgeryVCR$^*$ & \underline{0.8880} & \underline{0.8878} & \underline{0.7465} & \underline{0.7250} & \underline{0.7304} & \underline{0.7383} & \underline{0.6558} & \underline{0.6764} & 0.6173 & 0.6591 & 0.6536 & \underline{0.7350} & 0.8551 & 0.8320 & 0.9666 & 0.9353 & \underline{0.7827} & \underline{0.8155} & \underline{0.7812} & \underline{0.8066} \\
      \rowcolor[HTML]{CCE6FF} \textbf{ForgeryVCR} & \textbf{0.9193} & \textbf{0.9180} & \textbf{0.7963} & \textbf{0.7800} & \textbf{0.7835} & \textbf{0.7842} & \textbf{0.7271} & \textbf{0.7556} & 0.6284 & \textbf{0.6909} & \textbf{0.8497} & \textbf{0.8550} & 0.9045 & 0.8953 & 0.9719 & 0.9453 & \textbf{0.7887} & \textbf{0.8224} & \textbf{0.7985} & \textbf{0.8237} \\
      \bottomrule
  \end{tabular}
  }
\end{table*}

\begin{table*}[tb!]
  \centering
  \small
  \caption{Quantitative comparison of forgery localization performance (B-IoU and P-IoU). B-IoU denotes region-level BBox-IoU and P-IoU denotes pixel-level IoU. Methods are separated by category (Specialist vs. MLLMs methods).}
  \label{tab:localization_comparison}

  \setlength{\tabcolsep}{1.8pt}

  \resizebox{\textwidth}{!}{
  \begin{tabular}{@{}c cc cc cc cc cc cc cc cc cc cc@{}}
      \toprule
      \multirow{2}{*}{Method} & \multicolumn{2}{c}{CASIA v1} & \multicolumn{2}{c}{Coverage} & \multicolumn{2}{c}{CocoGlide} & \multicolumn{2}{c}{NIST16} & \multicolumn{2}{c}{Korus} & \multicolumn{2}{c}{DSO} & \multicolumn{2}{c}{Columbia} & \multicolumn{2}{c}{In-the-wild} & \multicolumn{2}{c}{SID-Set} & \multicolumn{2}{c}{Weighted Avg.} \\
      \cmidrule(lr){2-3} \cmidrule(lr){4-5} \cmidrule(lr){6-7} \cmidrule(lr){8-9} \cmidrule(lr){10-11} \cmidrule(lr){12-13} \cmidrule(lr){14-15} \cmidrule(lr){16-17} \cmidrule(lr){18-19} \cmidrule(lr){20-21}
       & B-IoU & P-IoU & B-IoU & P-IoU & B-IoU & P-IoU & B-IoU & P-IoU & B-IoU & P-IoU & B-IoU & P-IoU & B-IoU & P-IoU & B-IoU & P-IoU & B-IoU & P-IoU & B-IoU & P-IoU \\
      \midrule
      MVSS-Net (\textit{TPAMI'22}) & 0.3765 & 0.3789 & 0.3404 & 0.3786 & 0.2285 & 0.2564 & 0.2368 & 0.2476 & 0.0646 & 0.0747 & 0.1778 & 0.1837 & 0.4868 & 0.5552 & 0.1839 & 0.2157 & 0.0738 & 0.0816 & 0.1192 & 0.1290 \\
      TruFor (\textit{CVPR'23})   & 0.4135 & 0.4176 & 0.2562 & 0.2776 & 0.2129 & 0.2148 & 0.2004 & 0.2233 & 0.1673 & 0.2017 & \textbf{0.7058} & \textbf{0.7913} & 0.7679 & \underline{0.8206} & 0.3920 & 0.4433 & 0.0939 & 0.1025 & 0.1478 & 0.1589 \\
      CoDE (\textit{TIFS'24})     & 0.6046 & \underline{0.6360} & 0.2186 & 0.3579 & 0.3493 & 0.3861 & 0.2973 & 0.3394 & 0.1592 & 0.2140 & 0.1957 & 0.2650 & 0.7914 & \textbf{0.8529} & 0.3723 & 0.4492 & 0.2044 & 0.2426 & 0.2532 & 0.2933 \\
      HDF-Net (\textit{TPAMI'24}) & 0.1521 & 0.2391 & 0.0838 & 0.1422 & 0.1864 & 0.2201 & 0.1385 & 0.1685 & 0.0850 & 0.1274 & 0.0768 & 0.1325 & 0.1583 & 0.2103 & 0.1175 & 0.1925 & 0.1056 & 0.1433 & 0.1138 & 0.1557 \\
      PIM (\textit{TPAMI'25})     & 0.3130 & 0.3058 & 0.1056 & 0.1251 & 0.1052 & 0.1122 & 0.1579 & 0.1617 & 0.0662 & 0.0900 & 0.0949 & 0.1236 & 0.6694 & 0.6747 & 0.2741 & 0.3080 & 0.1178 & 0.1277 & 0.1421 & 0.1512 \\
      SAFIRE (\textit{AAAI'25})   & 0.2644 & 0.2593 & 0.4835 & 0.5095 & \underline{0.4499} & \textbf{0.4258} & 0.3582 & 0.3624 & \textbf{0.2026} & \textbf{0.2505} & 0.4296 & 0.4369 & 0.7306 & 0.7073 & 0.4004 & 0.4052 & 0.2483 & 0.2429 & 0.2740 & 0.2694 \\
      \midrule
      FakeShield (\textit{ICLR'25}) & 0.4598 & 0.4843 & 0.1933 & 0.1930 & \textbf{0.4577} & \underline{0.4229} & 0.1978 & 0.2187 & 0.0829 & 0.0955 & 0.3712 & 0.3940 & 0.7413 & 0.6996 & 0.4255 & 0.4195 & 0.2340 & 0.2189 & 0.2659 & 0.2551 \\
      SIDA (\textit{CVPR'25})       & 0.3246 & 0.3446 & 0.1396 & 0.1990 & 0.3503 & 0.3327 & 0.2138 & 0.2500 & 0.0898 & 0.1303 & 0.1561 & 0.1756 & 0.5223 & 0.4664 & 0.3084 & 0.3361 & 0.2594 & \textbf{0.2900} & 0.2655 & 0.2927 \\
      \midrule
      ForgeryVCR$^*$ & \underline{0.6499} & 0.6312 & \textbf{0.5963} & \textbf{0.6034} & 0.3601 & 0.3375 & \textbf{0.4481} & \textbf{0.4532} & \underline{0.1993} & \underline{0.2418} & 0.3731 & 0.3984 & \underline{0.9020} & 0.7628 & \underline{0.6009} & \underline{0.5735} & \underline{0.2667} & 0.2798 & \underline{0.3224} & \underline{0.3293} \\
      \rowcolor[HTML]{CCE6FF} \textbf{ForgeryVCR} & \textbf{0.7031} & \textbf{0.6619} & \underline{0.5882} & \underline{0.5850} & 0.3895 & 0.3571 & \underline{0.4015} & \underline{0.4115} & 0.1522 & 0.1902 & \underline{0.4806} & \underline{0.5148} & \textbf{0.9259} & 0.7652 & \textbf{0.6202} & \textbf{0.5910} & \textbf{0.2829} & \underline{0.2893} & \textbf{0.3386} & \textbf{0.3380} \\
      \bottomrule
  \end{tabular}
  }
\end{table*}

\subsection{Image Forgery Detection Evaluation}
\label{sec:sota_detection}

Table~\ref{tab:detection_comparison} presents the quantitative comparison for image-level forgery detection. ForgeryVCR achieves SOTA performance, attaining the best weighted average F1-score of 0.7985 and Accuracy of 0.8237, with significant margins on challenging benchmarks such as Coverage, NIST16, and DSO. Furthermore, ForgeryVCR consistently outperforms ForgeryVCR$^*$ which additionally incorporates textual CoT, confirming that purely visual reasoning avoids the semantic hallucinations introduced by linguistic descriptions and yields higher detection accuracy. Notably, while pure localization methods attain perfect scores on the all-manipulated In-the-wild dataset, their low accuracy on balanced benchmarks exposes a severe false-positive bias, whereas ForgeryVCR maintains consistent metrics across diverse datasets, confirming superior generalization.
\subsection{Image Forgery Localization Evaluation}
\label{sec:sota_localization}

Beyond detection, we evaluate localization performance in \cref{tab:localization_comparison} using both region-level B-IoU and pixel-level P-IoU. ForgeryVCR achieves the best weighted B-IoU of 0.3386 and, by further leveraging SAM2 with predicted bounding boxes as visual prompts, the best P-IoU of 0.3380, attaining SOTA on both localization metrics. The performance advantage over ForgeryVCR$^*$ underscores that the purely Visual-Centric Reasoning mechanism is essential for high-precision localization, as the inclusion of textual rationales tends to weaken the spatial precision required for guiding segmentation. Meanwhile, specialist networks often exhibit domain-specific variances, whereas ForgeryVCR delivers superior performance across both metrics and diverse manipulation types.

\subsection{Comparison with General MLLMs}
\label{sec:mllm_comparison}

To evaluate the effectiveness of ForgeryVCR against general-purpose architectures, we benchmark representative open-source MLLMs. For fairness, these baselines are fine-tuned on the identical dataset with the same SFT and RL training steps as ForgeryVCR, but without any chain-of-thought reasoning, including both textual CoT and forensic tool use. We enforce a standardized evaluation pipeline to ensure consistency, applying the same image-level metric and utilizing SAM2 to generate segmentation masks from the predicted bounding boxes, strictly adhering to the protocols in Table~\ref{tab:detection_comparison} and Table~\ref{tab:localization_comparison}. As presented in Table~\ref{tab:mllm_finetune_comparison}, general MLLMs exhibit a distinct deficit compared to ForgeryVCR, even when aligned with identical forensic objectives. This gap stems from standard pre-training, which prioritizes high-level semantics over the high-frequency sensitivity required for forensics. This confirms that visual-centric reasoning is indispensable for capturing subtle forgery artifacts.
\begin{table}[t]
  \centering
  \small
  \caption{Performance comparison of general MLLMs fine-tuned via the SFT and RL training pipeline.}
  \label{tab:mllm_finetune_comparison}

  \setlength{\tabcolsep}{4pt}

  \resizebox{\columnwidth}{!}{
  \begin{tabular}{ccccc}
    \toprule
    \multirow{2}{*}{Method} & \multicolumn{2}{c}{Detection} & \multicolumn{2}{c}{Localization} \\
    \cmidrule(lr){2-3} \cmidrule(lr){4-5}
     & F1 & ACC & F1 & IoU \\
    \midrule

    InternVL3.5-8B &
    0.6856 & 0.7105 &
    0.3874 & 0.3482 \\

    Llama-3.2-11B-Vision &
    0.7085 & 0.7090 &
    0.4575 & 0.4076 \\

    LLaVA-OneVision-1.5-8B &
    0.6671 & 0.5005 & 0.1711 & 0.1216 \\

    Qwen-3-VL-2B &
    0.7057 & 0.7568 & 0.4597 & 0.4218 \\

    Qwen-3-VL-4B &
    \underline{0.7351} & 0.7561 & \underline{0.4814} & \underline{0.4357} \\

    Qwen-3-VL-8B &
    0.7180 & \underline{0.7686} & 0.4454 & 0.4053 \\
    \midrule
    \rowcolor[HTML]{CCE6FF}
    \textbf{ForgeryVCR} &
    \textbf{0.8271} & \textbf{0.8261} &
    \textbf{0.5881} & \textbf{0.5306} \\

    \bottomrule
  \end{tabular}
  }
\end{table}

\subsection{Ablation Studies}
\label{sec:ablation_studies}

Unless otherwise noted, all ablation experiments report weighted average metrics over the eight evaluation benchmarks (excluding SID-Set). SID-Set is excluded from ablation evaluation as its large scale would dominate the weighted average and obscure the fine-grained differences across other benchmarks. The full evaluation including SID-Set is reported in Table~\ref{tab:detection_comparison} and Table~\ref{tab:localization_comparison}.

\paragraph{Impact of Training Stages.}
Table \ref{tab:ablation_training_stages} delineates the contribution of each training phase. The SFT stage significantly bridges the domain gap, with both \textit{Multi-Trajectories Synthesis (MTS)} and \textit{Intractable Sample Removal (ISR)} proving essential for diverse trajectories and data quality. The subsequent RL stage yields the most substantial gains, confirming that reward-driven optimization effectively refines decision boundaries and localization precision.
\begin{table}[t]
  \centering
  \caption{Ablation study on the effectiveness of training stages.}
  \label{tab:ablation_training_stages}
  \resizebox{\columnwidth}{!}{
  \begin{tabular}{lcccc}
  \toprule
  \multirow{2}{*}{Method} & \multicolumn{2}{c}{Image-level Detection} & \multicolumn{2}{c}{Pixel-level Localization} \\
  \cmidrule(lr){2-3} \cmidrule(lr){4-5}
   & F1 & ACC & F1 & IoU \\
  \midrule
  Base Model &
  0.5200 & 0.6600 &
  0.2193 & 0.2742 \\

  + SFT &
  0.6289 & 0.5936 &
  0.3948 & 0.3573 \\

  + SFT + ISR &
  0.6499 & 0.6589 &
  0.3953 & 0.3596 \\

  + SFT + ISR + MTS &
  \underline{0.6903} & \underline{0.6893} &
  \underline{0.4464} & \underline{0.4049} \\

  \rowcolor[HTML]{CCE6FF}
  \textbf{ForgeryVCR} &
  \textbf{0.8271} & \textbf{0.8261} &
  \textbf{0.5881} & \textbf{0.5306} \\
  \bottomrule
  \end{tabular}
  }
  \end{table}

\begin{table}[t]
  \centering
  \caption{Ablation study on different CoT modalities.}
  \label{tab:ablation_tools_cot}
  \resizebox{\columnwidth}{!}{
  \small
  \begin{tabular}{cccccc}
  \toprule
  \multicolumn{2}{c}{CoT Modality} & \multicolumn{2}{c}{Image-level Detection} & \multicolumn{2}{c}{Pixel-level Localization} \\
  \cmidrule(lr){1-2} \cmidrule(lr){3-4} \cmidrule(lr){5-6}
  Visual & Textual & F1 & ACC & F1 & IoU \\
  \midrule

  $\times$ & $\times$ &
  0.7351 & 0.7561 &
  0.4814 & 0.4357 \\

  $\times$ & $\checkmark$ &
  0.6431 & 0.7215 &
  0.4514 & 0.4107 \\

  $\checkmark$ & $\checkmark$ &
  \underline{0.7756} & \underline{0.7711} &
  \underline{0.5805} & \underline{0.5249} \\

  \rowcolor[HTML]{CCE6FF}
  $\checkmark$ & $\times$ &
  \textbf{0.8271} & \textbf{0.8261} &
  \textbf{0.5881} & \textbf{0.5306} \\

  \bottomrule
  \end{tabular}
  }
\end{table}

\begin{table}[t!]
  \centering
  \caption{Ablation on reward components in RL training.}
  \label{tab:ablation_reward_functions}

  \resizebox{\columnwidth}{!}{
  \small
  \begin{tabular}{ccccccc}
      \toprule
      \multicolumn{3}{c}{Reward Functions} &
      \multicolumn{2}{c}{Image-level Detection} &
      \multicolumn{2}{c}{Pixel-level Localization} \\

      \cmidrule(lr){1-3} \cmidrule(lr){4-5} \cmidrule(lr){6-7}

      $R_{\text{cls}}$ & $R_{\text{loc}}$ & $R_{\text{tool}}$ &
      F1 & ACC & F1 & IoU \\

      \midrule

      \checkmark & & &
      \underline{0.7891} & \underline{0.7944} &
      0.5227 & 0.4688 \\

      \checkmark & \checkmark & &
      0.7544 & 0.7339 &
      \underline{0.5554} & \underline{0.5005} \\

      \rowcolor[HTML]{CCE6FF}
      \checkmark & \checkmark & \checkmark &
      \textbf{0.8271} & \textbf{0.8261} &
      \textbf{0.5881} & \textbf{0.5306} \\

      \bottomrule
  \end{tabular}
}
\end{table}

\begin{table}[t!]
  \centering
  \caption{Ablation study on toolbox components.}
  \label{tab:ablation_toolbox_components}

  \resizebox{\columnwidth}{!}{
  \small
  \begin{tabular}{cccccc}
    \toprule
    \multicolumn{2}{c}{Toolbox} & \multicolumn{2}{c}{Image-level Detection} & \multicolumn{2}{c}{Pixel-level Localization} \\
    \cmidrule(lr){1-2} \cmidrule(lr){3-4} \cmidrule(lr){5-6}

    \begin{tabular}{@{}c@{}}Forensics Tools\end{tabular} &
    \begin{tabular}{@{}c@{}}Zoom-In Tool\end{tabular} &
    F1 & ACC & F1 & IoU \\

    \midrule

    $\checkmark$ & &
    \underline{0.7342} & \underline{0.7700} &
    \underline{0.4843} & \underline{0.4420} \\

    & $\checkmark$ &
    0.7118 & 0.6981 &
    0.4574 & 0.4099 \\

    \rowcolor[HTML]{CCE6FF}
    $\checkmark$ & $\checkmark$ &
    \textbf{0.8271} & \textbf{0.8261} &
    \textbf{0.5881} & \textbf{0.5306} \\

    \bottomrule
  \end{tabular}
  }
\end{table}

\paragraph{Impact of Reasoning Strategies.}
Table~\ref{tab:ablation_tools_cot} compares distinct reasoning paradigms. The results show that Visual-CoT significantly boosts image-level Accuracy from 0.7561 to 0.8261 compared to the baseline. This improvement indicates that explicitly materializing forensic artifacts into visual intermediate steps is essential for capturing subtle manipulation traces. In contrast, the inclusion of Textual-CoT results in performance degradation across both detection and localization tasks. This suggests that intermediate linguistic rationales introduce hallucinations, whereas direct visual alignment offers superior precision for forensic analysis.

\paragraph{Impact of Reward Functions.}
Table \ref{tab:ablation_reward_functions} analyzes the contribution of distinct reward components. We observe that relying solely on the classification reward yields suboptimal localization, whereas adding the localization reward improves pixel-level metrics but degrades detection accuracy, indicating an optimization conflict. The integration of the Tool Utility Reward mitigates this issue by guiding strategic tool invocation. Consequently, the comprehensive reward formulation attains superior performance across all metrics, demonstrating that incentivizing efficient tool usage effectively harmonizes the trade-off between global forgery detection and fine-grained forgery localization.

\paragraph{Impact of Toolbox Components.}
Table \ref{tab:ablation_toolbox_components} evaluates the contribution of the toolbox components. We observe that utilizing Zoom-In mechanism alone yields inferior performance compared to forensic tools. Since MLLMs inherently lack perception of low-level statistical inconsistencies, mere visual magnification fails to capture imperceptible traces. Consequently, the substantial performance gain from their combination stems from the forensic tools' ability to materialize imperceptible traces, directing the MLLM to suspicious regions for fine-grained inspection.

\paragraph{Impact of Tool Selection Strategy.}
Table \ref{tab:ablation_tool_selection} investigates the necessity of adaptive tool selection via four variants: (1) \textit{All Tools, Arbitrary Order}---invoke all tools in arbitrary sequence; (2) \textit{All Tools, Optimal Order}---invoke all tools in gain-ranked order from SFT; (3) \textit{Random Selection \& Order}---randomly sample a subset and shuffle order; and (4) the full \textit{ForgeryVCR} with RL-driven adaptive selection.

\begin{table}[t]
  \centering
  \caption{Ablation study on tool selection strategies.}
  \label{tab:ablation_tool_selection}
  \resizebox{\columnwidth}{!}{
  \begin{tabular}{lcccc}
  \toprule
  \multirow{2}{*}{Strategy} & \multicolumn{2}{c}{Image-level Detection} & \multicolumn{2}{c}{Pixel-level Localization} \\
  \cmidrule(lr){2-3} \cmidrule(lr){4-5}
   & F1 & ACC & F1 & IoU \\
  \midrule
  All Tools, Arbitrary Order &
  0.6879 & 0.7545 &
  0.4255 & 0.3900 \\

  All Tools, Optimal Order &
  \underline{0.7602} & \underline{0.7928} &
  \underline{0.5060} & \underline{0.4606} \\

  Random Selection \& Order &
  0.7563 & 0.7831 &
  0.5056 & 0.4596 \\

  \rowcolor[HTML]{CCE6FF}
  \textbf{ForgeryVCR} &
  \textbf{0.8271} & \textbf{0.8261} &
  \textbf{0.5881} & \textbf{0.5306} \\
  \bottomrule
  \end{tabular}
  }
  \end{table}

Indiscriminate invocation (Arbitrary Order) yields the worst results, as redundant or conflicting signals degrade performance. Imposing gain-ranked order improves substantially, yet exhaustive invocation still underperforms, showing that per-tool optimality does not guarantee combinatorial optimality. Random sampling performs comparably at the image level but falls short in localization, indicating that random tool sampling cannot substitute for structured gain-driven selection. The full ForgeryVCR achieves the best results by a considerable margin, validating that the RL-driven strategy effectively learns \textit{which} tools to invoke and \textit{in what order}.

\begin{figure}[t]
  \centering
  \includegraphics[width=\linewidth]{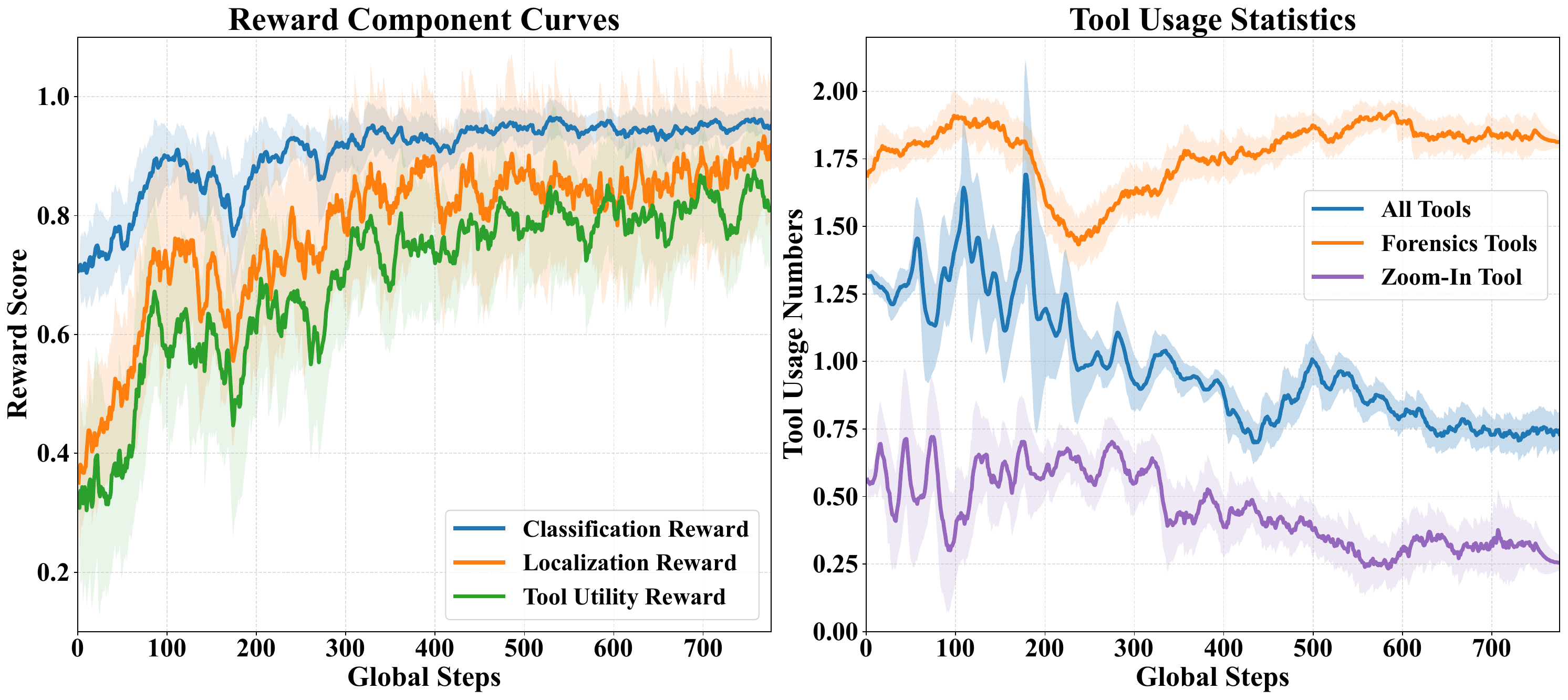}
  \caption{Evolution of training dynamics during the RL stage.}
  \Description{Line charts showing the evolution of training dynamics during the reinforcement learning stage.}
  \label{fig:stacked_metrics}
\end{figure}

\subsection{Analysis}
\label{sec:analysis}
\paragraph{Training Dynamics.}
Fig.~\ref{fig:stacked_metrics} tracks model dynamics during the RL stage. The steady ascent in total reward contrasts with the stabilization of interaction turns at a lower level, indicating that the MLLM evolves from random exploration to adaptive execution. Crucially, this reduction in tool usage does not compromise performance; instead, it reflects a learned selectivity where the model invokes forensic tools only when necessary. This confirms that reward-driven optimization fosters a task-aware strategy, balancing detection and localization precision with efficiency.

\paragraph{Tool Usage.}

\begin{figure}[t]
  \centering
  \includegraphics[width=\linewidth]{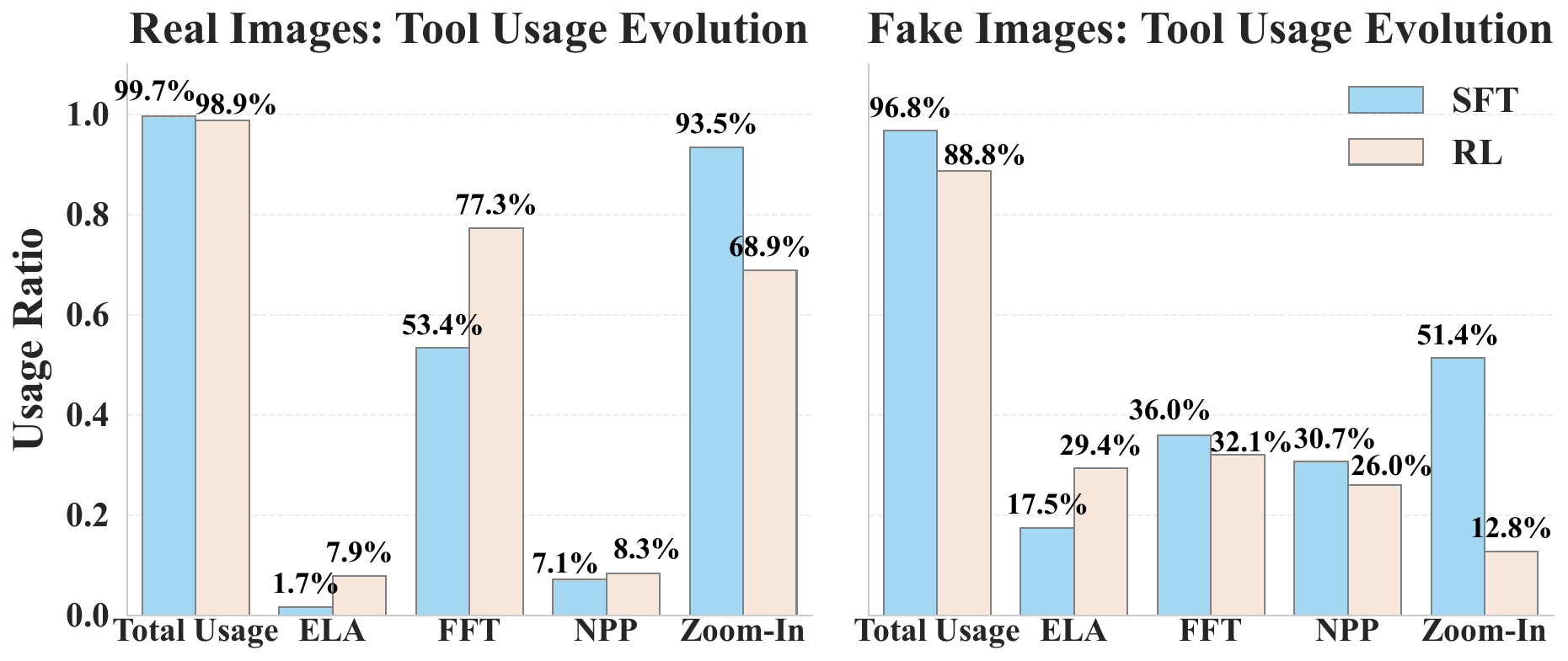}
  \caption{Comparison of specific tool usage ratios between SFT and RL stages.}
  \Description{Bar charts comparing tool usage ratios between SFT and RL stages.}
  \label{fig:tool_usage_distribution}

\end{figure}

We further analyze tool invocation distributions in Fig.~\ref{fig:tool_usage_distribution}. The SFT stage exhibits indiscriminate tool usage, notably applying Zoom-In frequently even on authentic images.

\paragraph{Hallucination Analysis.}
A key motivation for Visual-Centric Reasoning is suppressing semantic hallucinations introduced by linguistic rationales. To quantify this, we adopt an LLM-as-a-Judge~\cite{Zheng2023JudgingNeurIPS} protocol using Qwen3-VL-32B-Instruct to score the reasoning text of ForgeryVCR$^*$ across three dimensions: (1) faithfulness of the global visual description; (2) accuracy of the forensic tool output description; and (3) coherence of the reasoning logic. Among all 25,588 evaluation samples, 9,881 are flagged as hallucinated, and ForgeryVCR achieves a 35.2\% correction rate on this subset — the fraction where ForgeryVCR predicts correctly while ForgeryVCR$^*$ does not — as shown in Table~\ref{tab:hallucination_analysis}, confirming that removing linguistic rationales substantially reduces hallucination-induced errors.

\begin{table}[t]
  \centering
  \small
  \caption{Performance on hallucinated samples.}
  \label{tab:hallucination_analysis}
  \setlength{\tabcolsep}{4pt}
  \resizebox{\columnwidth}{!}{
  \begin{tabular}{lcccccc}
    \toprule
    \multirow{2}{*}{Strategy} & \multicolumn{2}{c}{Image-Level Detection} & \multicolumn{2}{c}{Pixel-Level Localization} \\
    \cmidrule(lr){2-3} \cmidrule(lr){4-5}
     & F1 & ACC & F1 & IoU \\
    \midrule
    ForgeryVCR$^*$ & 0.6718 & 0.5107 & 0.2844 & 0.2504 \\
    \rowcolor[HTML]{CCE6FF} \textbf{ForgeryVCR} & \textbf{0.7416} & \textbf{0.6138} & \textbf{0.3234} & \textbf{0.2857} \\
    \bottomrule
  \end{tabular}
  }
\end{table}

\paragraph{Robustness Analysis.}
ForgeryVCR demonstrates strong robustness under common post-processing distortions, including JPEG compression, Gaussian noise, Gaussian blur, and image resizing. Detailed analysis is provided in the Appendix.

\section{Conclusion}
In this paper, we propose ForgeryVCR, which establishes a Visual-Centric Reasoning paradigm to empower MLLMs with the capability to perceive imperceptible tampering traces. By converting statistical inconsistencies into explicit visual evidence, our approach effectively mitigates semantic hallucinations caused by linguistic descriptions. To ensure rigorous tool utilization, we introduce a Strategic Tool Learning pipeline. The gain-driven SFT stage constructs diverse reasoning trajectories to retain high-quality investigative paths, while the subsequent RL optimization, guided by a Tool Utility Reward, fosters an efficient execution policy that maximizes detection accuracy and localization precision. Extensive experiments demonstrate that ForgeryVCR achieves SOTA performance, exhibiting superior generalization and exceptional robustness against post-processing degradations.

\begin{acks}
This work was supported by the Shenzhen Science and Technology Program
(JCYJ20250604181211016 and SYSPG20241211174032004).
\end{acks}

\bibliographystyle{ACM-Reference-Format}
\balance
\bibliography{references}

\clearpage
\appendix
\section{Overview}
\label{app:overview}

In this appendix, we provide a comprehensive supplement to the main paper, detailing the dataset composition, the rationale behind forensic tools and reasoning trajectory construction, implementation protocols, and extended experimental analysis of ForgeryVCR. The content is organized as follows:

\begin{itemize}
    \item \textbf{Section~\ref{app:datasets_and_scale} (Dataset Composition and Scale)} details the specific composition of training and evaluation datasets used in ForgeryVCR, including dataset statistics for all benchmarks.

    \item \textbf{Section~\ref{app:toolbox_construction} (Hybrid Forensics Toolbox Construction)} presents a rigorous analysis of the forensic tool selection process. This includes both quantitative validation and qualitative rationale for pruning the candidate pool to the selected set (ELA, FFT, NPP, and Zoom-In), highlighting their mutual orthogonality and effectiveness.
    
    \item \textbf{Section~\ref{app:trajectory_construction} (Visual-Centric Trajectory Synthesis Details)} elaborates on the data generation algorithms described in the main text, specifically the Gain-Driven Tool Selection and Multi-Trajectories Synthesis modules. We also provide statistical breakdowns of the synthesized reasoning paths.

    \item \textbf{Section~\ref{app:training_implementation_and_reward_formulation} (Training Implementation and Reward Formulation)} provides the precise mathematical definitions of the reward components utilized in the RL stage. Additionally, we list the exact training hyperparameters and model configurations for the main results, ablation studies, and the preliminary Lightweight Tool-Specific Fine-Tuning.

    \item \textbf{Section~\ref{app:extended_experiments} (Extended Experimental Analysis)} presents supplementary results on reward-weight sensitivity, tool-usage evolution, computational cost, and robustness under image degradations.

    \item \textbf{Section~\ref{app:qualitative_analysis} (Qualitative Visualization and Analysis)} offers visual insights into the model's decision-making process. This covers the evolution of tool usage policies after RL optimization, a visual comparison between Visual-Centric and Visual-Textual reasoning paradigms, and qualitative examples of predicted masks against baseline methods.

    \item \textbf{Section~\ref{app:prompt_templates} (Prompt Templates)} lists the exact system prompts and template structures used for constructing the Visual-Centric Chain-of-Thought data, ensuring the reproducibility of our data generation pipeline.
\end{itemize}

\section{Dataset Composition and Scale}
\label{app:datasets_and_scale}

\subsection{Dataset Composition}
\label{app:dataset_statistics}

\textbf{Training Data.} The Strategic Tool Learning Pipeline involves two phases: Supervised Fine-Tuning (SFT) and Reinforcement Learning (RL).
\begin{itemize}
    \item \textbf{SFT} We strictly utilize a balanced subset of the CASIA v2 dataset~\cite{Dong2013CASIADatabase}, consisting of 5,123 authentic and 5,123 tampered images. This equilibrium prevents the model from developing a bias toward authenticity during initialization.
    \item \textbf{RL} To enhance policy robustness, we expand the training distribution by incorporating the IMD2020 dataset~\cite{Novozamsky2020IMD2020} and a subset of the FantasticReality dataset~\cite{Kniaz2019PointFantasticReality}. For FantasticReality, we sample a balanced set of 5,000 authentic and 5,000 tampered images to maintain training efficiency.
\end{itemize}

\textbf{Evaluation Data.} We employ a comprehensive suite of nine benchmarks to assess generalization across Splicing (SP), Copy-Move (CM), and Removal/Inpainting (INP). These include CASIA v1~\cite{Dong2013CASIADatabase}, Coverage~\cite{Wen2016COVERAGEICIP}, CocoGlide~\cite{Guillaro2023TruForCVPR}, NIST16~\cite{Guan2019NIST16WACVW}, Korus~\cite{Korus2017TIFS}, DSO-1~\cite{DeCarvalho2013DSOTIFS}, Columbia~\cite{Hsu2006ColumbiaICME}, In-the-wild~\cite{Huh2018InthewildECCV}, and SID-Set~\cite{Huang2025SIDA}. Notably, the In-the-wild dataset consists exclusively of manipulated images, providing a rigorous test for false negative rates.

\section{Hybrid Forensics Toolbox Construction}
\label{app:toolbox_construction}

In this section, we provide a detailed analysis of the forensic tool selection process. We visualize the outputs of the full candidate pool and articulate the rationale for excluding specific descriptors based on redundancy, visual distinctiveness, and their compatibility with the visual encoders of MLLMs.

\subsection{Initial Candidate Pool}
\label{subsec:candidate_tools}
Our initial investigation explored a diverse set of \textbf{8 forensic descriptors} spanning spatial, frequency, and generative domains. As illustrated in Fig.~\ref{fig:all_tools}, we visualize the outputs of these tools alongside the ground truth (GT). The candidates include:

\begin{figure}[h]
    \centering
    \includegraphics[width=\linewidth]{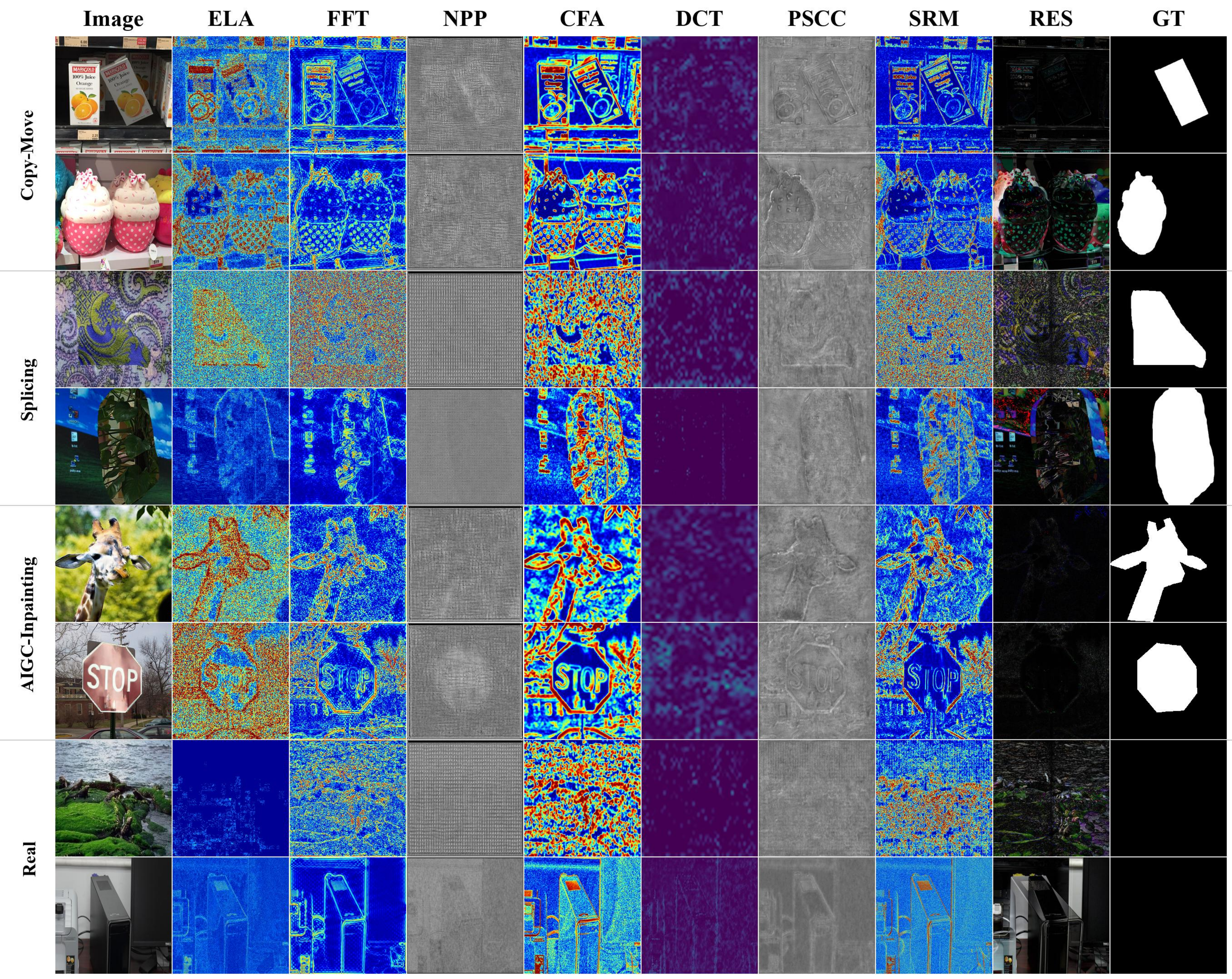}
    \caption{Visual comparison of the initial candidate pool of forensic tools across splicing, copy-move, removal and Inpainting manipulations. Our final selection (ELA, FFT, NPP, and Zoom-In) retains the most visually distinct features, whereas others exhibit redundancy or lack perceptible visual cues for the MLLM.}
    \label{fig:all_tools}
\end{figure}

\begin{itemize}
  \item \textbf{Spatial \& Statistical Consistency Tools:}
  \begin{itemize}
      \item \textbf{Error Level Analysis (ELA)}~\cite{Lu2025MFDFIMLCSCWD_ELA} (Selected): 
      This method highlights discrepancies in compression artifacts. By computing the difference between the original image and a re-compressed version (typically at fixed JPEG quality), ELA reveals regions that have undergone different compression histories compared to the background.
      
      \item \textbf{Noise Print++ (NPP)}~\cite{Guillaro2023TruForCVPR} (Selected): 
      A learned noise extractor that captures camera model fingerprints (PRNU) and anomalous noise residuals. It excels at exposing inconsistencies where the local sensor noise pattern deviates from the global camera signature, often creating a sharp contrast in manipulated areas.
      
      \item \textbf{Progressive Spatio-Channel Correlation (PSCC)}~\cite{Liu2022PSCCNetTCSVT} (Excluded): 
      Derived from the PSCC-Net architecture, this tool utilizes a Spatio-Channel Correlation Module to capture dense feature correlations. It produces a probability map intended to highlight manipulation masks by analyzing both spatial and channel-wise inconsistencies in a coarse-to-fine manner.
      
      \item \textbf{Spatial Rich Models (SRM)}~\cite{Su2024NovelIHMMSECSRM} (Excluded): 
      Originally designed for steganalysis, these are a set of high-pass filters that extract local pixel dependencies. SRM visualizes high-frequency residuals to expose disruptions in local texture statistics that are invisible to the naked eye.
      
      \item \textbf{Color Filter Array (CFA)}~\cite{Zhou2018RichFeaturesCVPRCFA} (Excluded): 
      This tool visualizes statistical traces of the demosaicing process. Since most cameras use a Bayer filter, image manipulation often disrupts the periodic interpolation patterns. The tool aims to expose these disruptions as visual artifacts in the probability map.
  \end{itemize}

  \item \textbf{Frequency Domain Tools:}
  \begin{itemize}
      \item \textbf{Fast Fourier Transform (FFT)}~\cite{Kashiani2025FreqDebiasCVPR_FFT} (Selected): 
      Transforms the spatial image signal into the frequency spectrum. It is particularly effective at revealing periodic anomalies, such as grid-like artifacts left by GAN upsampling or checkerboard patterns from resizing, which appear as distinct high-frequency spectral peaks.
      
      \item \textbf{Discrete Cosine Transform (DCT)}~\cite{Kwon2022CATNetIJCV} (Excluded): 
      Analyzes frequency coefficients within local blocks (e.g., $8\times8$). It is widely used to detect double JPEG compression by analyzing the histogram of coefficients, visualizing block-level inconsistencies in the frequency domain.
  \end{itemize}
  
  \item \textbf{Generative Domain Tools:}
  \begin{itemize}
      \item \textbf{VAE Reconstruction Residuals (RES)} (Excluded): 
      Inspired by the concept of reconstruction-based detection~\cite{Wang2023DIREICCV}, we utilize the \textbf{KL-regularized Autoencoder (VAE)} from \textbf{Stable Diffusion v2.1-base} to compute pixel-wise residuals. Specifically, we encode the image into the latent space and decode it back to the pixel space. The resulting map visualizes the absolute difference between the original input and the VAE reconstruction. The hypothesis is that the VAE, pre-trained on large-scale authentic images, yields higher reconstruction errors in regions containing unnatural manipulation artifacts or high-frequency anomalies that fall outside its learned latent distribution.
  \end{itemize}
\end{itemize}

\subsection{Qualitative Analysis and Tool Selection Rationale}
\label{subsec:qualitative_rationale}

As observed in Fig.~\ref{fig:all_tools}, simply stacking all forensic visualizations introduces significant noise and redundancy. Based on the visual evidence from the initial candidate pool, we derived the following rationale for our pruning process, ensuring the selected tools provide the distinctest visual signals for the MLLM.

\textbf{Redundancy in Spatial Residuals (CFA \& SRM).} 
While CFA and SRM are staples in traditional steganalysis, their visual outputs in Fig.~\ref{fig:all_tools} appear as high-frequency visual static (random noise). For instance, in the first row (spliced box), although CFA captures some traces, the signal is buried within the background noise. This noisy texture is semantically similar to natural high-frequency details (like gravel or foliage), making it difficult for general-purpose vision encoders (e.g., CLIP, SigLIP) to distinguish forgery from texture. In contrast, the selected \textbf{FFT} transforms these invisible periodicities into distinct, star-like geometric patterns (Row 2), and \textbf{NPP} highlights the manipulated region with significantly higher contrast and cleaner background suppression. We exclude CFA and SRM to avoid flooding the model with ambiguous noise patterns.

\textbf{Functional Overlap in Consistency Checks (PSCC vs. NPP).} 
Both tools aim to identify inconsistencies in source fingerprints or lighting. However, a visual comparison in Fig.~\ref{fig:all_tools} reveals a stark difference in clarity. \textbf{PSCC} (Column 6) tends to produce muddy or diffuse gray-scale heatmaps where the manipulated region blends into the background with soft edges. Conversely, \textbf{NPP} (Column 4) produces a sharp, almost binary-like mask where the tampered region (e.g., the white silhouette in Row 4) stands out vividly against a suppressed gray background. This high Signal-to-Noise Ratio in NPP allows the MLLM to perform precise localization without the ambiguity found in PSCC's outputs. Thus, we retain NPP as the superior consistency checker.

\textbf{Visual Ineffectiveness (DCT and RES).} 
Despite their theoretical soundness, these tools fail to produce human- or model-perceptible visual cues in many scenarios.
\begin{itemize}
    \item \textbf{DCT:} As shown in the DCT column of Fig.~\ref{fig:all_tools}, the resulting maps are predominantly black or dark purple, dominated by grid artifacts. The crucial coefficients that indicate forgery are mathematically present but visually imperceptible to a frozen vision encoder. The lack of salient visual features makes DCT an inefficient token consumer for Visual-CoT.
    \item \textbf{RES (Generative Residuals):} The RES maps rely on reconstruction errors from a VAE. While effective for diffusion-generated artifacts, they suffer from severe texture leak in traditional manipulations. As seen in Row 3 (the group of girls), RES highlights the complex textures of the hair and faces of authentic people just as strongly as potential forgeries. This creates visual confusion, leading the MLLM to hallucinate forgeries on legitimate complex textures. 
\end{itemize}

Consequently, our final \textbf{Hybrid Forensics Toolbox} retains only ELA, NPP, FFT, and the Zoom-In mechanism. This configuration maximizes forensic coverage—capturing compression, sensor noise, and frequency anomalies—while ensuring every input visual token carries a distinct, high-contrast forensic signal.

\subsection{Quantitative Ablation Study}
\label{subsec:quantitative_ablation}

To determine the effective configuration for the Hybrid Forensics Toolbox, we conducted a rigorous two-step quantitative analysis on the CASIA v2 dataset and generalization benchmarks. This process avoids arbitrary selection by evaluating both the individual capability of each tool and the potential gain from their combination.

\textbf{Step 1: Individual Tool Effectiveness.}
First, we assessed the intrinsic value of each forensic descriptor. Following the Lightweight Tool-Specific Fine-tuning protocol mentioned in Section 3.3, we trained separate SFT models for each candidate tool. Each model takes the original RGB image and one specific forensic view (e.g., ELA) as input. We then evaluated these specialist models on the test set. 

Table~\ref{tab:single_tool_perf} presents the performance of each single-tool expert. The results indicate that ELA, FFT, NPP, and the Zoom-In mechanism individually provide the strongest forensic signals, whereas tools like DCT and RES yield lower performance, suggesting they offer less perceptible information to the MLLM.

\begin{table}[h]
  \centering
  \caption{Performance of single-tool specialist models. Each row represents an MLLM fine-tuned exclusively with the original image and the specified tool. \textbf{Zoom-In}, \textbf{NPP}, \textbf{ELA} and \textbf{FFT} demonstrate the most significant individual contributions.}
  \label{tab:single_tool_perf}
  \setlength{\tabcolsep}{6pt}
  \begin{tabular}{ccccc}
      \toprule
      \multirow{2}{*}{Forensic Tool} & \multicolumn{2}{c}{Image-level Detection} & \multicolumn{2}{c}{Pixel-level Localization} \\
      \cmidrule(lr){2-3} \cmidrule(lr){4-5}
      & F1 & ACC & F1 & IoU \\
      \midrule
      Baseline & 0.5672 & 0.5926 & 0.2746 & 0.2465 \\
      \midrule
      + ELA     & \textbf{0.6973} & \underline{0.6420} & \textbf{0.4167} & \textbf{0.3731} \\
      + FFT     & 0.6002 & 0.6146 & 0.3276 & 0.2890 \\
      + NPP     & 0.5819 & 0.5956 & 0.3082 & 0.2775 \\
      + CFA     & 0.5959 & 0.5906 & 0.3294 & 0.2968 \\
      + DCT     & 0.5240 & 0.5715 & 0.2590 & 0.2355 \\
      + PSCC    & 0.5025 & 0.5701 & 0.2542 & 0.2311 \\
      + RES     & 0.5105 & 0.5698 & 0.2538 & 0.2300 \\
      + SRM     & 0.5902 & 0.6015 & 0.3304 & 0.2983 \\
      + Zoom-In & \underline{0.6045} & \textbf{0.6615} & \underline{0.3569} & \underline{0.3261} \\
      \bottomrule
  \end{tabular}
\end{table}

\textbf{Step 2: Theoretical Optimal Performance Analysis.}
Merely stacking tools can introduce redundancy. To identify the point of diminishing returns, we calculated the \textit{Theoretical Optimal Performance} (Upper Bound) for cumulative tool combinations. 

\textit{Calculation Method:} For a given set of available tools $T_{set}$, and a specific test sample $x$, we simulate a Perfect Selector. We aggregate the predictions from all single-tool models trained in Step 1 corresponding to tools in $T_{set}$. The sample is considered correctly detected/localized if \textit{at least one} tool in $T_{set}$ produces a correct classification or a high-quality mask. Specifically, the theoretical IoU for sample $x$ is defined as $IoU_{best} = \max_{t \in T_{set}} (IoU_t)$.

Table~\ref{tab:theoretical_bound} shows the trajectory of this theoretical upper bound as we incrementally add tools. We observe that combining \textbf{ELA, FFT, NPP, and Zoom-In} yields a significant performance leap. However, adding subsequent tools (CFA, DCT, etc.) results in marginal gains or saturation. This confirms that the selected subset covers the necessary forensic modalities (compression, frequency, noise, and detail) effectively, while further additions offer redundant information.

\begin{table*}[h]
  \centering
  \caption{\textit{Theoretical Optimal Performance} analysis with cumulative tool integration. The metrics represent the upper bound achievable if the model perfectly selects the best tool from the available subset for each sample. The performance gain saturates after the inclusion of the first four tools.}
  \label{tab:theoretical_bound}
  \resizebox{\linewidth}{!}{
    \begin{tabular}{ccccccccc|cc|cc}
      \toprule
      \multicolumn{9}{c}{Tool Configuration} & \multicolumn{2}{c}{Image-level Detection} & \multicolumn{2}{c}{Pixel-level Localization} \\
      \cmidrule(lr){1-9} \cmidrule(lr){10-11} \cmidrule(lr){12-13}
      ELA & FFT & NPP & Zoom & CFA & DCT & PSCC & RES & SRM & F1 & ACC & F1 & IoU \\
      \midrule
       & & & & & & & & & 
       0.5672 & 0.5926 & 0.2746 & 0.2465 \\
      
      \checkmark & & & & & & & & & 
      0.6920 \scriptsize{(+12.5\%)} & 0.6420 \scriptsize{(+4.9\%)} & 0.4167 \scriptsize{(+14.2\%)} & 0.3731 \scriptsize{(+12.7\%)} \\
      
      \checkmark & \checkmark & & & & & & & & 
      0.7678 \scriptsize{(+7.6\%)} & 0.7479 \scriptsize{(+10.6\%)} & 0.4538 \scriptsize{(+3.7\%)} & 0.4064 \scriptsize{(+3.3\%)} \\
      
      \checkmark & \checkmark & \checkmark & & & & & & & 
      0.8009 \scriptsize{(+3.3\%)} & 0.7869 \scriptsize{(+3.9\%)} & 0.4801 \scriptsize{(+2.6\%)} & 0.4296 \scriptsize{(+2.3\%)} \\
      
      \rowcolor[HTML]{CCE6FF} 
      \checkmark & \checkmark & \checkmark & \checkmark & & & & & & 
      0.8773 \scriptsize{(+7.6\%)} & 0.8776 \scriptsize{(+9.1\%)} & 0.4972 \scriptsize{(+1.7\%)} & 0.4450 \scriptsize{(+1.5\%)} \\
      
      \checkmark & \checkmark & \checkmark & \checkmark & \checkmark & & & & & 
      0.8805 \scriptsize{(+0.3\%)} & 0.8807 \scriptsize{(+0.3\%)} & 0.4981 \scriptsize{(+0.1\%)} & 0.4457 \scriptsize{(+0.1\%)} \\
      
      \checkmark & \checkmark & \checkmark & \checkmark & \checkmark & \checkmark & & & & 
      0.8812 \scriptsize{(+0.1\%)} & 0.8823 \scriptsize{(+0.2\%)} & 0.4985 \scriptsize{(+0.0\%)} & 0.4462 \scriptsize{(+0.1\%)} \\
      
      \checkmark & \checkmark & \checkmark & \checkmark & \checkmark & \checkmark & \checkmark & & & 
      0.8812 \scriptsize{(+0.0\%)} & 0.8823 \scriptsize{(+0.0\%)} & 0.4992 \scriptsize{(+0.1\%)} & 0.4466 \scriptsize{(+0.0\%)} \\
      
      \checkmark & \checkmark & \checkmark & \checkmark & \checkmark & \checkmark & \checkmark & \checkmark & & 
      0.8812 \scriptsize{(+0.0\%)} & 0.8823 \scriptsize{(+0.0\%)} & 0.4996 \scriptsize{(+0.0\%)} & 0.4469 \scriptsize{(+0.0\%)} \\
      
      \checkmark & \checkmark & \checkmark & \checkmark & \checkmark & \checkmark & \checkmark & \checkmark & \checkmark & 
      0.8817 \scriptsize{(+0.1\%)} & 0.8839 \scriptsize{(+0.2\%)} & 0.5003 \scriptsize{(+0.1\%)} & 0.4475 \scriptsize{(+0.1\%)} \\
      \bottomrule
  \end{tabular}
  }
\end{table*}

\begin{table*}[h]
  \centering
  \caption{Detailed ablation study on specific combinations within the selected toolset. We evaluate permutations of ELA, FFT, and NPP, both with and without the Zoom-In mechanism. The study encompasses 15 distinct configurations, ranging from single-tool baselines to the comprehensive \textbf{Full Suite} (highlighted in blue), which yields the best performance.}
  \label{tab:granular_ablation}
  \setlength{\tabcolsep}{5pt}
  \begin{tabular}{cccccccc}
      \toprule
      \multicolumn{4}{c}{Toolbox Components} & \multicolumn{2}{c}{Image-level Detection} & \multicolumn{2}{c}{Pixel-level Localization} \\
      \cmidrule(r){1-4} \cmidrule(lr){5-6} \cmidrule(l){7-8}
      ELA & FFT & NPP & Zoom-In & F1 & ACC & F1 & IoU \\
      \midrule
      
      \checkmark & & & \checkmark & 0.6617 & 0.6786 & 0.4404 & 0.3994 \\
       & \checkmark & & \checkmark & 0.6814 & 0.6958 & 0.4714 & 0.4281 \\
       & & \checkmark & \checkmark & 0.6643 & 0.6764 & 0.4482 & 0.4076 \\
       
      \checkmark & \checkmark & & \checkmark & 0.6648 & 0.6788 & 0.4428 & 0.4031 \\
      \checkmark & & \checkmark & \checkmark & 0.6526 & 0.6811 & 0.4433 & 0.4051 \\
       & \checkmark & \checkmark & \checkmark & 0.6860 & 0.6822 & 0.4618 & 0.4192 \\
       
      \rowcolor[HTML]{CCE6FF} \checkmark & \checkmark & \checkmark & \checkmark & \textbf{0.7242} & \textbf{0.6893} & \textbf{0.5215} & \textbf{0.4699} \\
      \midrule
      
      \checkmark & & & & 0.6742 & 0.6881 & 0.4235 & 0.3844 \\
       & \checkmark & & & 0.6588 & 0.6830 & 0.3951 & 0.3596 \\
       & & \checkmark & & 0.6543 & 0.6756 & 0.4027 & 0.3656 \\
       
      \checkmark & \checkmark & & & 0.6188 & 0.6578 & 0.3735 & 0.3383 \\
      \checkmark & & \checkmark & & 0.6322 & 0.6584 & 0.3951 & 0.3597 \\
       & \checkmark & \checkmark & & 0.5896 & 0.6523 & 0.3512 & 0.3201 \\
       
      \checkmark & \checkmark & \checkmark & & 0.6903 & 0.6875 & 0.4464 & 0.4049 \\
      \bottomrule
  \end{tabular}
\end{table*}

Having established that ELA, FFT, NPP, and Zoom-In constitute the theoretically optimal subset above, we now proceed to empirically validate their synergy. We conduct a fine-grained SFT ablation study to confirm that these components offer complementary forensic information and that the full suite outperforms any partial combination.

\textbf{Ablation on Selected Tool Combinations.}
To validate the internal synergy of the proposed toolbox, we perform a detailed breakdown of the components. While Table 7 in the main text establishes the superiority of the full framework, this granular analysis isolates the contribution of each forensic descriptor (ELA, FFT, NPP) and their interaction with the Zoom-In mechanism.

Table~\ref{tab:granular_ablation} presents the SFT performance across different subsets, revealing the intrinsic synergy within our toolbox. The results demonstrate a strong complementarity among forensic tools, where the combination of ELA, FFT, and NPP consistently outperforms single-tool baselines by covering diverse manipulation traces—ranging from compression artifacts and frequency anomalies to sensor noise. Crucially, this forensic foundation is further amplified by the Zoom-In mechanism, which universally lifts performance across all metrics. This confirms that visual refinement acts as a powerful attention guide, enabling the MLLM to apply global forensic insights to precise local regions, with the complete suite achieving the best overall performance.

\section{Visual-Centric Trajectory Synthesis Details}
\label{app:trajectory_construction}
\subsection{Trajectory Synthesis Pipeline}
\label{app:trajectory_synthesis_pipeline}

The construction of the Cold-Start SFT dataset involves a rigorous process of model-based filtration and trajectory synthesis. We first establish a foundation by training lightweight specialist models to assess the intrinsic utility of each forensic tool.

\paragraph{Lightweight Tool-Specific Fine-Tuning.}
To obtain the performance metrics required for our gain-driven tool selection, we train a set of independent specialist models on the Cold-Start training dataset. These models include one \textbf{Tool-Free Baseline} and separate \textbf{Single-Tool Experts} for each candidate tool (ELA, NPP, FFT, and Zoom-In). 
The precise data formatting for these models—\textbf{Data Structure~1: Tool-Free Baseline} and \textbf{Data Structure~2: Single-Tool Expert (e.g., FFT)}—is provided at the end of this appendix in \S\ref{app:prompt_templates}, alongside all other prompt templates, to keep the present discussion focused on the methodology. The Tool-Free Baseline relies solely on the original image, whereas the Single-Tool Experts incorporate specific forensic visualizations alongside expert-guided prompts to facilitate visual diagnostics.

The detailed hyperparameters for this lightweight fine-tuning stage are provided in Table~\ref{tab:experimental_configurations}. Notably, we employ a short training schedule (5 epochs) and strictly visual-input-only prompting to evaluate the raw perceptibility of the forensic features without the aid of linguistic and visual-centric reasoning chains.

\paragraph{Algorithm.}
\label{app:algorithms}

To ensure the instruction-tuning dataset fosters genuine visual-centric reasoning rather than hallucination or tool redundancy, we formalize the data construction process into two cooperative phases as detailed in Algorithm \ref{alg:combined_pipeline}. The primary objective is to curate a set of reasoning trajectories that are both diagnostically valid and diverse in complexity. In the first phase, denoted as Gain-Driven Tool Selection, we rigorously filter the candidate pool $T_{pool}$ to retain only those forensic operators capable of revealing imperceptible artifacts. By evaluating the performance gain $P_t$ of each tool-specific expert against a tool-free baseline $P_{base}$ and a validity threshold $\tau$, we isolate the discriminative signals essential for the specific sample $(x, y)$. This process simultaneously executes Intractable Sample Removal to discard instances where all models fail to extract useful features, returning an optimized and ranked tool sequence $T_{rank}$. Subsequently, the Multi-Trajectories Synthesis phase utilizes this ranked sequence to construct a comprehensive set of reasoning paths $P_{aug}$. To cultivate an adaptive policy capable of handling varying difficulty levels, we synthesize trajectories ranging from direct visual assessment to independent evidence collection and progressive multi-view verification. This structured diversification ensures the model encounters specific tool invocations only when they yield tangible information gains, thereby mitigating dependency on ineffective operations.

Regarding the truncation parameter $K$ mentioned in the main paper, we apply a dynamic setting to balance the SFT training data: $K=4$ for tampered samples and $K=2$ for authentic ones. Since the lightweight specialist models inherit the MLLM's intrinsic perception on real images, achieving high accuracy with minimal assistance, restricting $K$ for authentic samples prevents redundant tool invocations while preserving valuable instruction data.

\textbf{Notation.} In Algorithm~\ref{alg:combined_pipeline}, $(x, y)$ denotes an image-label pair; $M_{base}$ is the tool-free baseline model and $\{M_t\}$ are single-tool expert models; $V_t$ represents the forensic tool map generated by tool $t$; the operator $\oplus$ denotes concatenation; $\text{Eval}(\cdot)$ returns the performance metric $P$; and $\mathcal{S}_{reject}$ indicates sample rejection via ISR.

\begin{algorithm}[!ht]
  \caption{Data Construction Pipeline: Tool Selection \& Trajectory Synthesis}
  \label{alg:combined_pipeline}
  
  
  \begin{minipage}[t]{0.48\textwidth}
    \textbf{Stage 1:} Gain-Driven Tool Selection \& Ranking
    \hrule
    \small 
    \begin{algorithmic}[1]
      \STATE \textbf{Input:} Sample $(x, y)$, Pool $T_{pool}$, Models $M_{base} \cup \{M_t\}$, Threshold $\tau$
      \STATE \textbf{Output:} Ranked tool sequence $T_{rank}$ or Reject Signal $\mathcal{S}_{reject}$
      
      \STATE Calc baseline $P_{base} \leftarrow \text{Eval}(M_{base}, x)$
      \STATE Calc tool perfs $P_{t} \leftarrow \text{Eval}(M_{t}, x \oplus V_t)$ $\forall t$
      \STATE $P_{max} \leftarrow \max(P_{base}, \max_{t} P_t)$
      
      \STATE \textit{// Execute Intractable Sample Removal}
      \IF{$P_{max} < \tau$}
          \STATE \textbf{Return} $\mathcal{S}_{reject}$
      \ENDIF
      
      \STATE \textit{// Select \& Rank Gain-Driven Tools}
      \STATE Initialize $T_{rank} \leftarrow \emptyset$
      \FOR{each tool $t \in T_{pool}$}
          \IF{$P_t > \max(P_{base}, \tau)$}
              \STATE $T_{rank} \leftarrow T_{rank} \cup \{t\}$
          \ENDIF
      \ENDFOR
      \STATE Sort $T_{rank}$ by $P_t$ in descending order
      
      \STATE \textbf{Return} $T_{rank}$

    \end{algorithmic}
  \end{minipage}
  \hfill
  \begin{minipage}[t]{0.48\textwidth}
    \hrule
    \vspace{0.3em}
    \textbf{Stage 2:} Multi-Trajectories Synthesis
    \hrule
    \small 
    \begin{algorithmic}[1]
      \STATE \textbf{Input:} Ranked Sequence $T_{rank}$, Truncation $K$
      \STATE \textbf{Output:} Reasoning Trajectories $P_{aug}$
      
      \STATE Init trajectories $P_{aug} \leftarrow \{ \emptyset \}$ \textit{// Direct Assessment}
      
      \IF{$T_{rank} \neq \emptyset$}
          \STATE $T'_{rank} \leftarrow T_{rank}[1 : \min(|T_{rank}|, K)]$ \textit{// Top-K Truncation}
          
          \STATE \textit{// Type 1: Independent Evidence Paths}
          \FOR{each tool $t_i \in T'_{rank}$}
              \STATE $P_{aug} \leftarrow P_{aug} \cup \{ [t_i] \}$
          \ENDFOR
          
          \STATE \textit{// Type 2: Accumulative Evidence Paths}
          \IF{$|T'_{rank}| \geq 2$}
              \FOR{$k = 2$ \textbf{to} $|T'_{rank}|$}
                  \STATE $S_{accum} \leftarrow [t_1, \dots, t_k]$ \textit{// Sequence up to k}
                  \STATE $P_{aug} \leftarrow P_{aug} \cup \{ S_{accum} \}$
              \ENDFOR
          \ENDIF
      \ENDIF
      
      \STATE \textbf{Return} $P_{aug}$

    \end{algorithmic}
  \end{minipage}
  \vspace{0.1cm}
  \hrule
\end{algorithm}

\subsection{Statistics of Synthesized Reasoning Trajectories}
\label{app:data_analysis}

\begin{figure}[h!]
  \centering
  \includegraphics[width=1.0\linewidth]{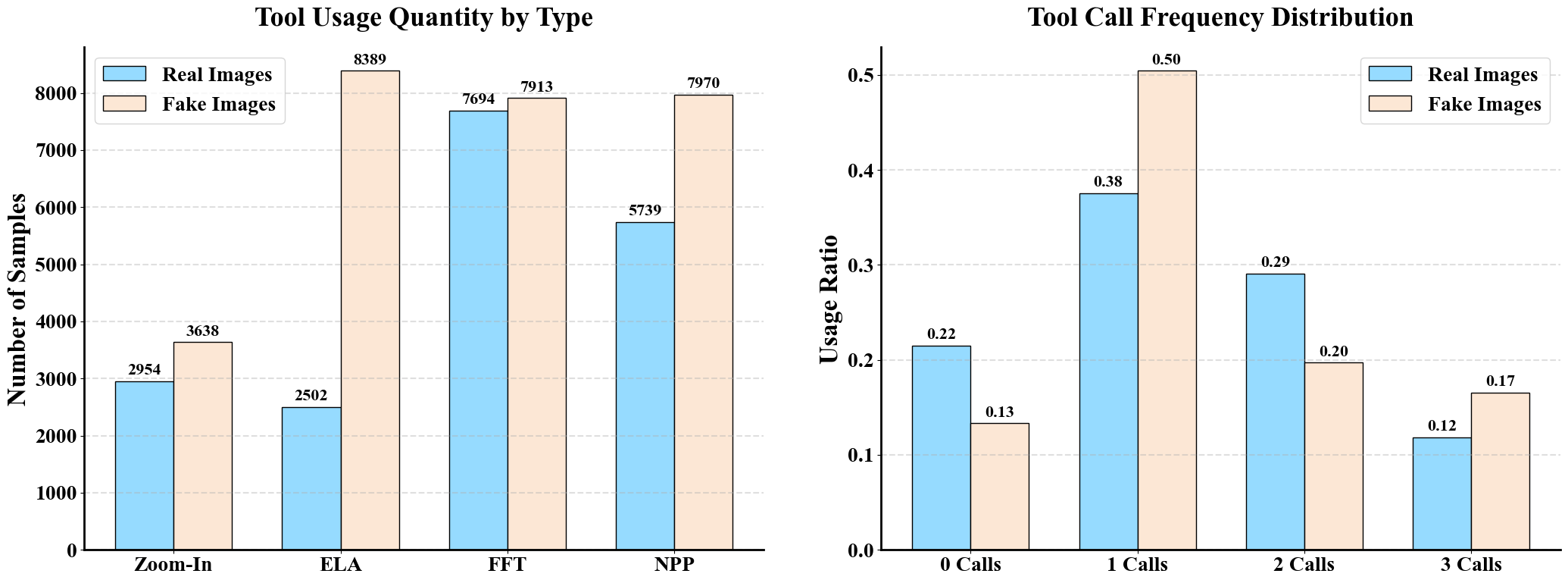}
  \vspace{-10pt}
  \caption{Statistical analysis of the synthesized SFT trajectories. \textbf{Left:} The frequency of specific tool invocations, showing high selectivity for ELA on manipulated samples. \textbf{Right:} The distribution of reasoning chain lengths, indicating that manipulated images generally require deeper multi-tool verification compared to authentic ones.}
  \label{fig:tool_usage_analysis}
\end{figure}

\begin{table}[h]
  \centering
  \caption{Statistics of the SFT dataset after applying the visual-centric trajectories synthesis pipeline. \textbf{Unique Source Images} refers to the distinct samples retained after Intractable Sample Removal (ISR). \textbf{Reasoning Trajectories} denotes the total number of logical chains generated via Multi-Trajectories Synthesis (MTS), effectively expanding the training data.}
  \label{tab:sample_counts_isr_mts}
  \small
  \setlength{\tabcolsep}{6pt}
  \begin{tabular}{lcc}
  \toprule
  \textbf{Category} & \textbf{Authentic} & \textbf{Tampered} \\
  \midrule
  Unique Source Images (Post-ISR) & 4,961 & 3,987 \\
  Reasoning Trajectories (Post-MTS) & 16,645 & 17,408 \\
  \bottomrule
  \end{tabular}
\end{table}

To contextualize the statistics presented here, we refer to the Trajectory Synthesis Pipeline detailed in Appendix~\ref{app:trajectory_synthesis_pipeline}. The initial specialist models were evaluated on the CASIA v2 training set to derive the Gain-Driven Tool Selection policy. This ensures that the constructed reasoning paths are derived from a controlled distribution where the efficient tool usage is empirically verified.

\textbf{Dataset Filtering and Expansion.} Table~\ref{tab:sample_counts_isr_mts} quantifies the dual effect of our pipeline. First, the Intractable Sample Removal (ISR) acts as a quality filter, discarding approximately 22\% of manipulated images where forensic tools failed to yield perceptible evidence. This step is critical to prevent the model from hallucinating explanations for undetectable forgeries. Second, the Multi-Trajectories Synthesis (MTS) expands these filtered images into a total of 34,053 diverse reasoning trajectories. This one-to-many mapping enables the model to learn adaptive investigation strategies—ranging from direct visual assessment to complex multi-tool verification—rather than memorizing a single fixed path for each image.

\textbf{Tool Usage Patterns.} Fig.~\ref{fig:tool_usage_analysis} illustrates the distribution of tool invocations within these trajectories, highlighting two key behaviors:
\begin{itemize}
    \item \textbf{Tool Selectivity:} As shown in the left panel, \textit{Error Level Analysis (ELA)} is invoked significantly more often for forged images (8,389) than for authentic ones (2,502). This confirms that ELA serves as a specific indicator for compression anomalies, whereas FFT and NPP act as broader validity checks applicable to both classes.
    \item \textbf{Reasoning Depth:} The right panel reveals that authentic images frequently trigger shorter reasoning chains (0 or 1 tool calls), suggesting that visual integrity is often established quickly. In contrast, manipulated images shift towards longer chains (1 to 3 calls), validating that our pipeline successfully synthesizes complexity-aware reasoning paths that deepen investigation when anomalies are suspected.
\end{itemize}

\subsection{Data Formulation Details}
\label{app:prompt_details}

In this section, we elaborate on the construction of the ground-truth annotations used for training. To derive the ground truth bounding boxes utilized in the objects field, we calculate the minimum bounding rectangle for each connected component within the pixel-level manipulation mask provided in the CASIA v2 and other forensic datasets. To ensure the annotations represent meaningful forensic targets rather than annotation noise, we apply a filtration strategy that discards connected components smaller than 100 pixels or constituting less than 0.05\% of the total image area. This rigorous filtering ensures that the supervision signal during the Supervised Fine-Tuning and Reinforcement Learning stages directs the model toward salient manipulation regions.

To intuitively elucidate the distinction between the conventional Visual-Textual CoT and our proposed Visual-Centric CoT, we present comparative visualization cases representing the input-output structures of both paradigms. Unlike the Visual-Textual CoT, which relies heavily on linguistic descriptions of high-level semantics, the Visual-Centric CoT explicitly incorporates tool execution codes and visual intermediates into the reasoning chain. These visualizations demonstrate how our framework grounds the verification process in tangible forensic evidence rather than ambiguous textual rationale, effectively mitigating semantic hallucinations. The exact system and user prompt templates governing these interactions are provided in \textbf{Appendix~\ref{sec:system_user_prompt}}.

\subsection{Construction of Visual-Textual CoT Data}
\label{app:textual_cot_construction}

To construct the \textit{Visual-Textual CoT} variant utilized in our ablation studies, we extend the reasoning trajectories $P_{aug}$ generated via the Multi-Trajectories Synthesis (Algorithm \ref{alg:combined_pipeline}). It is important to note that the sequence of tool invocations and the final decision path remain identical to the \textit{Visual-Only} version; the only difference lies in the insertion of linguistic rationales (wrapped in \texttt{<think>} tags) that explain the visual evidence.

To generate these high-fidelity textual rationales, we explicitly employ \textbf{Qwen-2.5-VL-72B-Instruct} \cite{Bai2025Qwen25VL} as the Teacher MLLM. It is crucial to distinguish this data generator from our target student model, \textbf{Qwen-3-VL-4B-Instruct}, utilized during the training phase. We select the 72B variant specifically for its superior visual reasoning capabilities and vast parameter scale, which enable it to synthesize intricate forensic descriptions that the smaller 4B model can subsequently learn to mimic through knowledge distillation.

To provide a precise visual reference and ensure reproducibility during the teacher's generation process, we strictly follow a standardized \textbf{Mask-Overlay Prompting Strategy}. Specifically, the binary ground-truth mask is first thresholded to isolate the manipulated region. We then apply an alpha-blending process where the manipulated region is rendered in a salient red color (RGB: 255, 0, 0) and superimposed onto the original image with a fixed transparency coefficient of $\alpha=0.5$. This configuration ensures that the specific location of the forgery is explicitly highlighted to guide the model's attention, while the semi-transparent nature of the overlay preserves the underlying textural details necessary for analyzing visual artifacts.

Addressing the concern regarding the teacher model's potential dependency on the explicit mask, we provide the following clarification:

\textbf{Concern: Does providing the ground-truth mask cause the teacher model to rely on the overlay rather than simulating genuine forensic findings?}

\textbf{Response:} We address this trade-off by drawing upon established methodologies in Visual Chain-of-Thought (V-CoT) literature \cite{Su2025OpenThinkIMGArXiv,Wang2025PixelReasonerArXiv}. Existing approaches for synthesizing reasoning trajectories typically fall into two categories:

\begin{itemize}
    \item \textbf{Autonomous Exploration via Strong Teachers \cite{Su2025OpenThinkIMGArXiv}:} This paradigm leverages the superior visual perception of large-scale models (e.g., GPT-4o) to autonomously plan action trajectories based on few-shot prompts. The teacher model functions as an autonomous planner, determining tool selection and execution sequences. While flexible, this approach relies heavily on the teacher's intrinsic ability to detect the initial anomaly without assistance. It typically requires a post-hoc filtering stage to discard invalid or hallucinatory paths.
    
    \item \textbf{Fixed-Trajectory Supervision for Specialized Tasks \cite{Wang2025PixelReasonerArXiv}:} This approach is preferred when the investigation logic is finite and well-defined. Instead of letting the model wander, it employs pre-designed prompt templates (e.g., ``Global Analysis $\rightarrow$ Tool Usage $\rightarrow$ Local Verification'') to force the teacher model into specific, high-gain reasoning patterns. By acting as a ``director,'' the framework ensures the generated data covers diverse and effective error-correction or refinement scenarios that a model might not spontaneously attempt.
\end{itemize}

\textbf{Justification for Our Approach:} We align with the \textbf{Fixed-Trajectory} paradigm for a critical reason: unlike general visual tasks, forensic artifacts (e.g., weak PRNU traces or subtle compression inconsistencies) are often imperceptible to even the strongest commercial MLLMs without tool assistance. An autonomous exploration approach would likely fail at the first step, as the teacher would not perceive the ``suspicious region'' required to trigger the correct tool. Therefore, we utilize the \textbf{Ground-Truth Mask} as a necessary ``director's cue.'' It does not serve as the final answer but rather anchors the teacher's attention, enabling it to simulate a successful detection and localization trajectory (e.g., describing the noise mismatch at that specific location) that would otherwise be impossible to synthesize.

Aligning with the \textbf{Fixed-Trajectory} paradigm, we implement the Mask-Overlay strategy. Our rationale and specific mitigation measures are as follows:
\begin{enumerate}
    \item \textbf{Necessity of Fixed Trajectories in Forensics:} Unlike general visual tasks where objects are salient, forensic artifacts (e.g., noise inconsistencies in ELA) are visually subtle and often imperceptible without precise localization. Relying solely on the teacher model's autonomous exploration poses a high risk of hallucination or missed detection, which would degrade the quality of the training data. Therefore, we adopt the fixed-trajectory strategy to ensure the teacher attends to the exact tampered region.
    \item \textbf{Mitigation via System Prompting:} To prevent the model from simply describing the "red mask," we enforce a strict negative constraint in the system prompt: \textit{"You must treat the red overlay solely as a location hint. Do NOT mention the red color or the mask itself. Focus your description entirely on the visual anomalies (texture, noise, resolution) visible BENEATH the overlay."} This ensures that while the mask guides \textit{where} to look, the generated rationale describes \textit{what} forensic features are present.
\end{enumerate}

Based on the structure of the trajectory derived in Stage 2 (Algorithm \ref{alg:combined_pipeline}), we categorize the generation process into three distinct prompt templates. The No-Tool Direct Assessment is applied when no external tools are retrieved. In cases where a single expert tool is selected, the Single-Tool Analysis is utilized. Finally, the Multi-Tool Synthesis is employed when multiple tools are involved.

\begin{itemize}
  \item \textbf{No-Tool Direct Assessment:} Applied when $P_{aug} = \{\emptyset\}$. The model is prompted to identify global semantic inconsistencies or obvious visual artifacts without external aids.
  \item \textbf{Single-Tool Analysis:} Applied when $P_{aug} = \{[t_i]\}$. The prompt focuses on interpreting the specific feature map (e.g., ELA noise patterns) returned by the selected expert tool $t_i \in T_{rank}$.
  \item \textbf{Multi-Tool Synthesis:} Applied when $P_{aug} = \{ [t_1, \dots, t_k] \}$. The model is tasked with synthesizing evidence from multiple views to form a comprehensive verdict.
\end{itemize}

The specific prompt templates corresponding to these three categories are detailed in \textbf{Appendix~\ref{sec:trajectory_synthesis_templates}}.

\section{Training Implementation and Reward Formulation}
\label{app:training_implementation_and_reward_formulation}

\subsection{Detailed Reward Formulation}
\label{app:reward_formulation}

In this section, we provide the precise mathematical definitions and implementation details for the reward components utilized in the Reinforcement Learning stage. 

\paragraph{Variable Definitions and Parsing Logic.}
Let $\mathcal{B}_{gt} = \{b_1, \dots, b_M\}$ denote the set of ground-truth bounding boxes for a manipulated image, where $M \ge 1$. Similarly, let $\hat{\mathcal{B}} = \{\hat{b}_1, \dots, \hat{b}_N\}$ represent the set of predicted bounding boxes.
To compute the rewards, we first parse the model's raw textual completion to extract the predicted label $\hat{c}$, the bounding boxes $\hat{\mathcal{B}}$, and the tool usage history $u_t$.
\begin{itemize}
    \item \textbf{Label and BBox Extraction:} We extract the content within the \texttt{<answer>...</answer>} tags. The label $\hat{c}$ is determined by keyword matching (e.g., ``fake'' or ``real''). If $\hat{c}=\text{fake}$, we identify the localization outputs encapsulated by the special tokens \texttt{<|box\_start|>} and \texttt{<|box\_end|>}. We then parse the coordinate strings within these tokens (formatted as \texttt{(x1, y1), (x2, y2)}) using regular expressions to construct the set $\hat{\mathcal{B}}$. If no valid coordinates are found, $\hat{\mathcal{B}}$ is treated as an empty set $\emptyset$.
    \item \textbf{Tool Usage Detection:} We define $u_t \in \{0, 1\}$ by scanning the conversation history. $u_t=1$ if we detect a valid \texttt{<tool\_call>} tag containing a supported function name (ELA, FFT, NPP, or Zoom-In) or a \texttt{<tool\_response>} indicating prior execution; otherwise, $u_t=0$.
\end{itemize}

\paragraph{Classification Reward ($R_{cls}$).}
This component ensures the model correctly distinguishes between authentic and manipulated images. Crucially, to mitigate semantic hallucinations where the model predicts ``fake'' without locating any actual tampering traces, we impose a validity constraint derived from the parsing logic: for manipulated samples, the reward is granted only if the prediction is accompanied by at least one valid bounding box.
\begin{equation}
    R_{cls} = \mathbb{I}(\hat{c} = c) \cdot \begin{cases} 
    1, & \text{if } c = \text{real} \\
    \mathbb{I}(\hat{\mathcal{B}} \neq \emptyset), & \text{if } c = \text{fake}
    \end{cases}
\end{equation}

\paragraph{Localization Reward ($R_{loc}$).}
To foster precise spatial reasoning, this reward evaluates the Intersection over Union (IoU) between the predicted regions and the ground truth. Since both the prediction and ground truth may contain multiple disjoint bounding boxes (i.e., $|\hat{\mathcal{B}}| \ge 1$ and $|\mathcal{B}_{gt}| \ge 1$), a simple one-to-one IoU is insufficient. 
We employ the \textbf{Hungarian matching algorithm} (also known as the Kuhn-Munkres algorithm) to solve this assignment problem:
\begin{enumerate}
    \item We construct a cost matrix $C$ of size $N \times M$, where each element $C_{ij} = 1 - \text{IoU}(\hat{b}_i, b_j)$.
    \item We apply the Hungarian algorithm to find the optimal assignment of predicted boxes to ground-truth boxes that minimizes the total cost (maximizes total IoU).
    \item The final metric, denoted as $\text{HungarianIoU}(\hat{\mathcal{B}}, \mathcal{B}_{gt})$, is calculated as the mean IoU of these matched pairs.
\end{enumerate}
This reward is applied exclusively to manipulated samples to prevent optimization conflicts on authentic images (where $\mathcal{B}_{gt} = \emptyset$).
\begin{equation}
    R_{loc} = \text{HungarianIoU}(\hat{\mathcal{B}}, \mathcal{B}_{gt}) \cdot \mathbb{I}(c = \text{fake})
\end{equation}

\paragraph{Tool Utility Reward ($R_{tool}$).}
This conditional reward is designed to cultivate an efficient investigation policy, penalizing the MLLM for invoking forensic tools when they are unnecessary (e.g., on obvious real images) or ineffective (yielding no informational gain). Based on the parsed tool usage $u_t$, a reward is deemed valid only if the tool invocation leads to a correct final verdict. For manipulated images, we impose a stricter requirement: the tool must contribute to a high-precision localization, defined by the Hungarian IoU exceeding a threshold $\tau_{\text{IoU}}$ (set to 0.5).
\begin{equation}
  R_{tool} = u_t \left( \mathbb{I}_{c=\text{real}} R_{cls} + \mathbb{I}_{c=\text{fake}} \mathbb{I}_{\text{HungarianIoU} > \tau_{\text{IoU}}} \right)
\end{equation}

\subsection{Implementation Details}
\label{app:implementation_details}

To ensure the reproducibility of our results and provide a comprehensive understanding of the experimental variables, we detail the specific configurations used for the main benchmarks and the comparative ablation studies.

\paragraph{Main Results Configuration.}
To substantiate the state-of-the-art performance reported in Table 1 and Table 2 of the main paper, we rigorously evaluated different reasoning paradigms. As discussed in the Experiments section, our preliminary investigation compared the efficacy of Visual-Textual CoT against the proposed Visual-Centric (Visual-Only) CoT. Empirical evidence demonstrated that the inclusion of intermediate linguistic rationales often introduced semantic hallucinations, whereas the Visual-Centric approach yielded superior robustness; thus, the latter was adopted as the standard configuration for all main results. Regarding the post-processing refinement, the segmentation masks are generated by the Hiera-Large variant of the Segment Anything Model 2 (SAM 2 Hiera Large), which receives the bounding box predicted by our model as a visual prompt. It is imperative to clarify that while SAM 2 serves as a refinement module for pixel-level segmentation, the fundamental region-level localization capability is acquired through our strategic tool learning pipeline. The ForgeryVCR explicitly identifies the manipulated regions and generates the bounding box coordinates that guide the segmentation process.
\begin{table*}[h]
  \centering
  \caption{Comprehensive overview of training configurations across all experiments. The table details the specific settings for the reasoning paradigms (CoT), SFT data strategies (ISR: Intractable Sample Removal, MTS: Multi-Trajectories Synthesis), RL reward components ($R_{cls}$, $R_{loc}$, $R_{tool}$), training duration, and trainable modules. ``General MLLMs'' refers to the baselines listed in Table 3 of the main paper.}
  \label{tab:experimental_configurations}
  \resizebox{\textwidth}{!}{%
  \begin{tabular}{l|c|cc|cc|ccc|c|ccc}
  \toprule
  \multirow{2}{*}{\textbf{Experiment / Method}} & \multirow{2}{*}{\textbf{Backbone}} & \multicolumn{2}{c|}{\textbf{CoT Modality}} & \multicolumn{2}{c|}{\textbf{SFT Strategy}} & \multicolumn{3}{c|}{\textbf{RL Rewards}} & \textbf{Duration} & \multicolumn{3}{c}{\textbf{Trainable Modules}} \\
   & & Visual & Textual & ISR & MTS & $R_{cls}$ & $R_{loc}$ & $R_{tool}$ & (Steps) & Aligner & Vision & LLM \\
  \midrule
  
  Base Model (Zero-shot) & Qwen3-VL-4B & $\times$ & $\times$ & $\times$ & $\times$ & $\times$ & $\times$ & $\times$ & - & - & - & - \\
  Lightweight Tool-Specific FT & Qwen3-VL-4B & $\times$ & $\times$ & $\times$ & $\times$ & $\times$ & $\times$ & $\times$ & 5 Epochs & \checkmark & \checkmark & $\times$ \\
  
  \midrule
  Fine-tuning General MLLMs & Various & $\times$ & $\times$ & $\times$ & $\times$ & \checkmark & \checkmark & $\times$ & 2,700 + 776 & \checkmark & \checkmark & \checkmark \\
  
  \midrule
  \multirow{2}{*}{Ablation: CoT Modalities} & \multirow{2}{*}{Qwen3-VL-4B} & $\times$ & $\times$ & $\times$ & $\times$ & \checkmark & \checkmark & $\times$ & 2,700 + 776 & \checkmark & \checkmark & $\times$ \\
   & & $\times$ & \checkmark & $\times$ & $\times$ & \checkmark & \checkmark & $\times$ & 2,700 + 776 & \checkmark & \checkmark & $\times$ \\
  
  ForgeryVCR* & Qwen3-VL-4B & \checkmark & \checkmark & \checkmark & \checkmark & \checkmark & \checkmark & \checkmark & 2,700 + 776 & \checkmark & \checkmark & \checkmark \\
  
  \midrule
  \rowcolor[HTML]{CCE6FF} 
  \textbf{ForgeryVCR (Ours)} & \textbf{Qwen3-VL-4B} & \textbf{\checkmark} & \textbf{$\times$} & \textbf{\checkmark} & \textbf{\checkmark} & \textbf{\checkmark} & \textbf{\checkmark} & \textbf{\checkmark} & \textbf{2,700 + 776} & \textbf{\checkmark} & \textbf{\checkmark} & \textbf{\checkmark} \\
  
  Ablation: Reward Variants & Qwen3-VL-4B & \checkmark & $\times$ & \checkmark & \checkmark & Var. & Var. & Var. & 2,700 + 776 & \checkmark & \checkmark & \checkmark \\
  Ablation: Toolbox Variants & Qwen3-VL-4B & \checkmark & $\times$ & \checkmark & \checkmark & \checkmark & \checkmark & \checkmark & 2,700 + 776 & \checkmark & \checkmark & \checkmark \\
  
  \bottomrule
  \end{tabular}%
  }
  \end{table*}

  \paragraph{Experiment-Specific Configurations.}
  To ensure transparency and reproducibility, we provide a detailed breakdown of the training configurations for all experiments in Table \ref{tab:experimental_configurations}. 
  Unless otherwise specified, all Supervised Fine-Tuning (SFT) stages are implemented using Low-Rank Adaptation (LoRA) with lora rank $r=32$, lora alpha $\alpha=64$, and a learning rate of $5\text{e-}4$. The Reinforcement Learning (RL) stages are conducted with LoRA settings of $r=128$, $\alpha=256$, and a learning rate of $5\text{e-}5$.
  Below, we elaborate on the specific settings for each experimental category, highlighting the differences in data strategies and optimization objectives.
  
  \begin{itemize}
      \item \textbf{Lightweight Tool-Specific Fine-Tuning (FT).} 
      This preliminary stage is designed to train the single-tool expert models required for the Gain-Driven Tool Selection process. As detailed in the first row of Table \ref{tab:experimental_configurations}, we employ a short training schedule of 5 epochs to prevent overfitting. Crucially, to focus on aligning visual features with forensic descriptors, we unfreeze the Vision Encoder and Projector (Aligner) while keeping the LLM backbone frozen. No complex data filtering (ISR) or synthesis (MTS) is applied here, as these models themselves serve as the filtering mechanism for subsequent stages.
  
      \item \textbf{ForgeryVCR* (Visual-Textual CoT).} 
      This variant represents the dual-modality baseline evaluated in our main comparative results. It follows the complete two-stage training pipeline, consisting of 2,700 steps for SFT and 776 steps for RL. To support its hybrid reasoning capability, the training data is constructed using both Intractable Sample Removal (ISR) and Multi-Trajectories Synthesis (MTS), containing both visual tool tokens and detailed linguistic rationales. Consequently, the model is optimized using all three reward components ($R_{cls}, R_{loc}, R_{tool}$), with all modules (Vision, Aligner, and LLM) set to trainable to accommodate the complex multimodal generation.
  
      \item \textbf{ForgeryVCR (Visual-Centric Reasoning).} 
      Our proposed Visual-Centric framework adheres to the same rigorous training schedule (2,700 SFT + 776 RL steps) and full parameter optimization (all modules trainable) as the Visual-Textual counterpart. The defining difference lies in the \textit{CoT Modality}: we explicitly exclude intermediate linguistic descriptions from the reasoning chain, forcing the model to rely solely on visual intermediates and tool tokens. This configuration maximizes the effectiveness of the Tool Utility Reward ($R_{tool}$) and minimizes semantic hallucinations, serving as the standard setting for our reported state-of-the-art results.
  
      \item \textbf{Fine-tuning General MLLMs.} 
      To benchmark against standard parameter adaptation, these baselines are fine-tuned using the identical data split and duration as ForgeryVCR. However, as they lack the forensic toolbox, they are trained without the specialized ISR/MTS strategies, using standard image-text pairs where the model directly predicts the verdict. Accordingly, the reward function during RL is modified to exclude the Tool Utility Reward ($R_{tool}$), focusing strictly on a weighted sum of Classification ($R_{cls}$) and Localization ($R_{loc}$) rewards.
  
      \item \textbf{Ablation: CoT Modalities.} 
      For the ``No CoT'' and ``Textual-Only'' variants referenced in the ablation studies, we adjust the modality columns as shown in the table. Notably, for these simplified baselines, we freeze the LLM backbone and only fine-tune the Vision Encoder and Aligner to isolate the impact of reasoning modalities. Since no tools are invoked, $R_{tool}$ is deactivated.
  
      \item \textbf{Ablation: Reward Functions.} 
      These experiments utilize the full ForgeryVCR configuration (Visual-Centric CoT, full trainable modules) but vary the specific components like $R_{cls}$, $R_{loc}$, and $R_{tool}$ during the RL stage to analyze the trade-offs between detection accuracy, localization precision and tool invocation strategy.
  
      \item \textbf{Ablation: Toolbox Variants.} 
      These runs maintain the standard ForgeryVCR pipeline and optimization settings but restrict the available tool set (e.g., removing NPP or Zoom-In) during the SFT data construction and inference phases. This isolation allows us to quantify the individual contribution of each forensic component.
  \end{itemize}

\section{Extended Experimental Analysis}
\label{app:extended_experiments}
In this section, we provide supplementary experimental results covering the Reward Weight Ablation Study, tool-usage evolution, computational cost, and robustness against post-processing degradations.

\subsection{Reward Weight Ablation Study}
\label{app:reward_ablation}

We investigate the sensitivity of the reinforcement learning process to the hyperparameters $\lambda_{cls}$, $\lambda_{loc}$, and $\lambda_{tool}$ in Eq. (5). Table~\ref{tab:reward_weight_sensitivity} summarizes the performance trade-offs. We observe that increasing the localization weight $\lambda_{loc}$ to 2.0 encourages better pixel-level grounding, but without a moderate penalty on tool usage ($\lambda_{tool}=0.5$), the model may over-invoke tools, leading to slight degradation. The combination of $\{1.0, 2.0, 0.5\}$ yields the optimal balance.

\begin{table*}[h]
    \centering
    \caption{Hyperparameter sensitivity analysis on reward weights. We investigate the impact of varying $\lambda_{cls}$, $\lambda_{loc}$, and $\lambda_{tool}$ on model performance. Best results are marked in \textbf{bold} and second-best are \underline{underlined}. The final configuration (highlighted) achieves the optimal trade-off.}
    \label{tab:reward_weight_sensitivity}
    \setlength{\tabcolsep}{8pt}
    \begin{tabular}{ccccccc}
        \toprule
        \multicolumn{3}{c}{Reward Weights} & \multicolumn{2}{c}{Image-level Detection} & \multicolumn{2}{c}{Pixel-level Localization} \\
        \cmidrule(r){1-3} \cmidrule(lr){4-5} \cmidrule(l){6-7}
        $\lambda_{cls}$ & $\lambda_{loc}$ & $\lambda_{tool}$ & F1 & ACC & F1 & IoU \\
        \midrule
        1.0 & 1.0 & 1.0 & 0.7920 & 0.7922 & 0.5716 & 0.5180 \\
        1.0 & 2.0 & 1.0 & 0.7883 & 0.7824 & \underline{0.5772} & \underline{0.5230} \\
        1.0 & 1.0 & 0.5 & \underline{0.7995} & \underline{0.8142} & 0.5475 & 0.4961 \\
        \rowcolor[HTML]{CCE6FF} 1.0 & 2.0 & 0.5 & \textbf{0.8271} & \textbf{0.8261} & \textbf{0.5881} & \textbf{0.5306} \\
        \bottomrule
    \end{tabular}
\end{table*}

\FloatBarrier
\subsection{Efficiency and Tool-Usage Analysis}
\label{app:efficiency_tool_usage}

\begin{figure}[H]
  \centering
  \includegraphics[width=\linewidth]{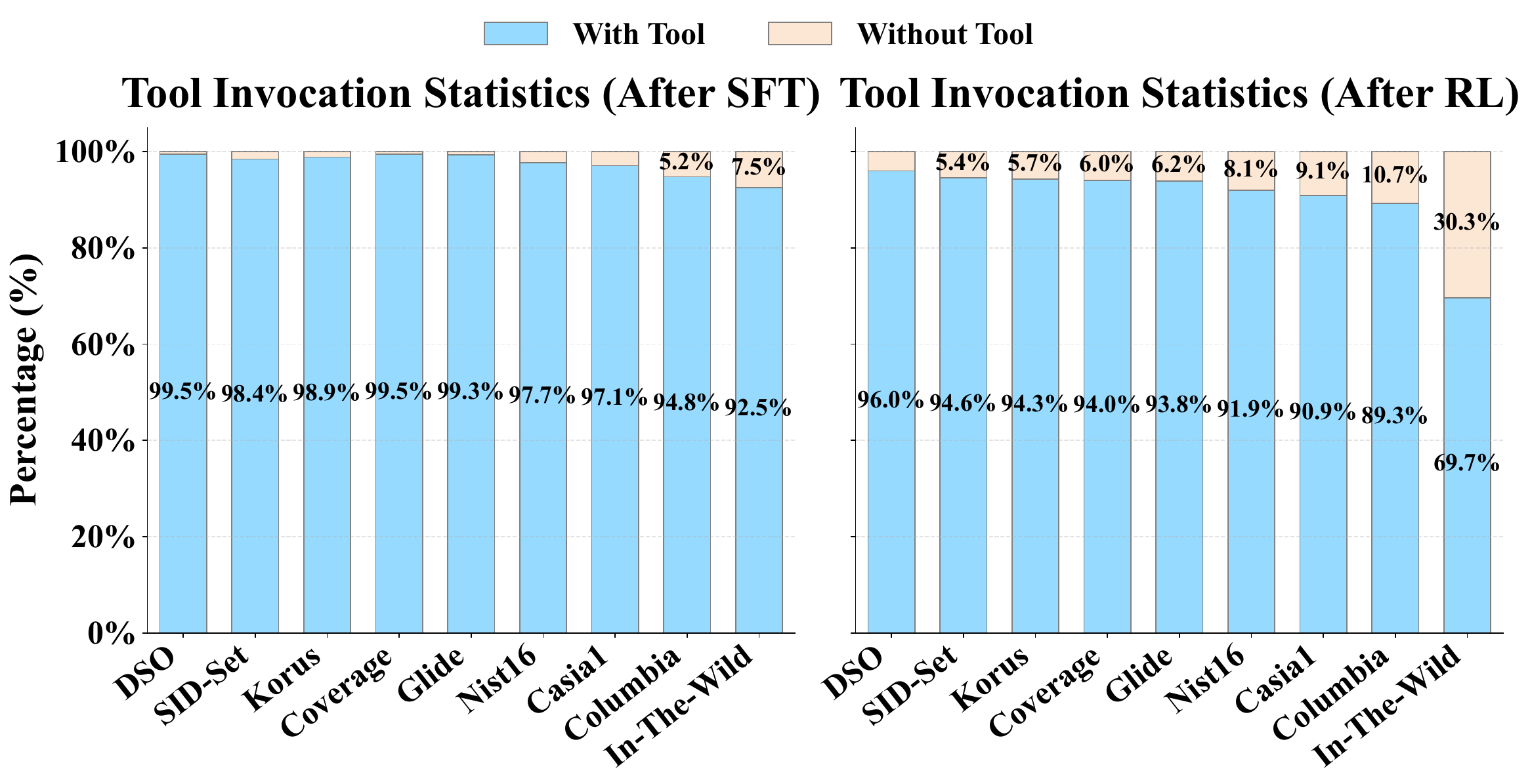}
  \caption{Evolution of tool usage ratios from SFT to RL stages.}
  \Description{Charts showing evolution of tool usage ratios from SFT to RL stages across different benchmarks.}
  \label{fig:app_tool_dependency_stats}
\end{figure}

\paragraph{Tool-Usage Evolution.}
Figure~\ref{fig:app_tool_dependency_stats} further shows the evolution of tool usage from SFT to RL across benchmarks. The SFT stage exhibits indiscriminate tool usage, notably applying Zoom-In frequently even on authentic images. The RL stage suppresses these non-informative invocations by optimizing for utility. The MLLM learns to bypass tool invocation when direct visual assessment suffices, leading to more direct inference across all benchmarks, confirming a shift from mechanical execution to an efficiency-oriented policy.

\paragraph{Computational Cost.}
We compare model parameters and inference time in Table~\ref{tab:app_computational_cost}, with all methods benchmarked on CASIA v1 with batch size 1 on a single GPU. Under a warm-model protocol averaging all 920 inputs, ForgeryVCR requires 310.51 ms per image.

\begin{table}[H]
  \centering
  \small
  \caption{Computational cost comparison.}
  \label{tab:app_computational_cost}
  \setlength{\tabcolsep}{4pt}
  \begin{tabular}{lcc}
    \toprule
    Method & Parameters (B) & Inference Time (ms) \\
    \midrule
    FakeShield & 13.37 & 6809.96 \\
    SIDA-13B & 14.01 & 412.14 \\
    SIDA-13B-description & 14.01 & 471.70 \\
    \midrule
    ForgeryVCR$^*$ & 4.44 & 337.49 \\
    \rowcolor[HTML]{CCE6FF} \textbf{ForgeryVCR} & \textbf{4.44} & \textbf{310.51} \\
    \bottomrule
  \end{tabular}
\end{table}

%


\subsection{Robustness against Post-Processing Degradations}
\label{app:robustness_degradation}

\begin{figure}[t!]
  \centering
  \includegraphics[width=\linewidth]{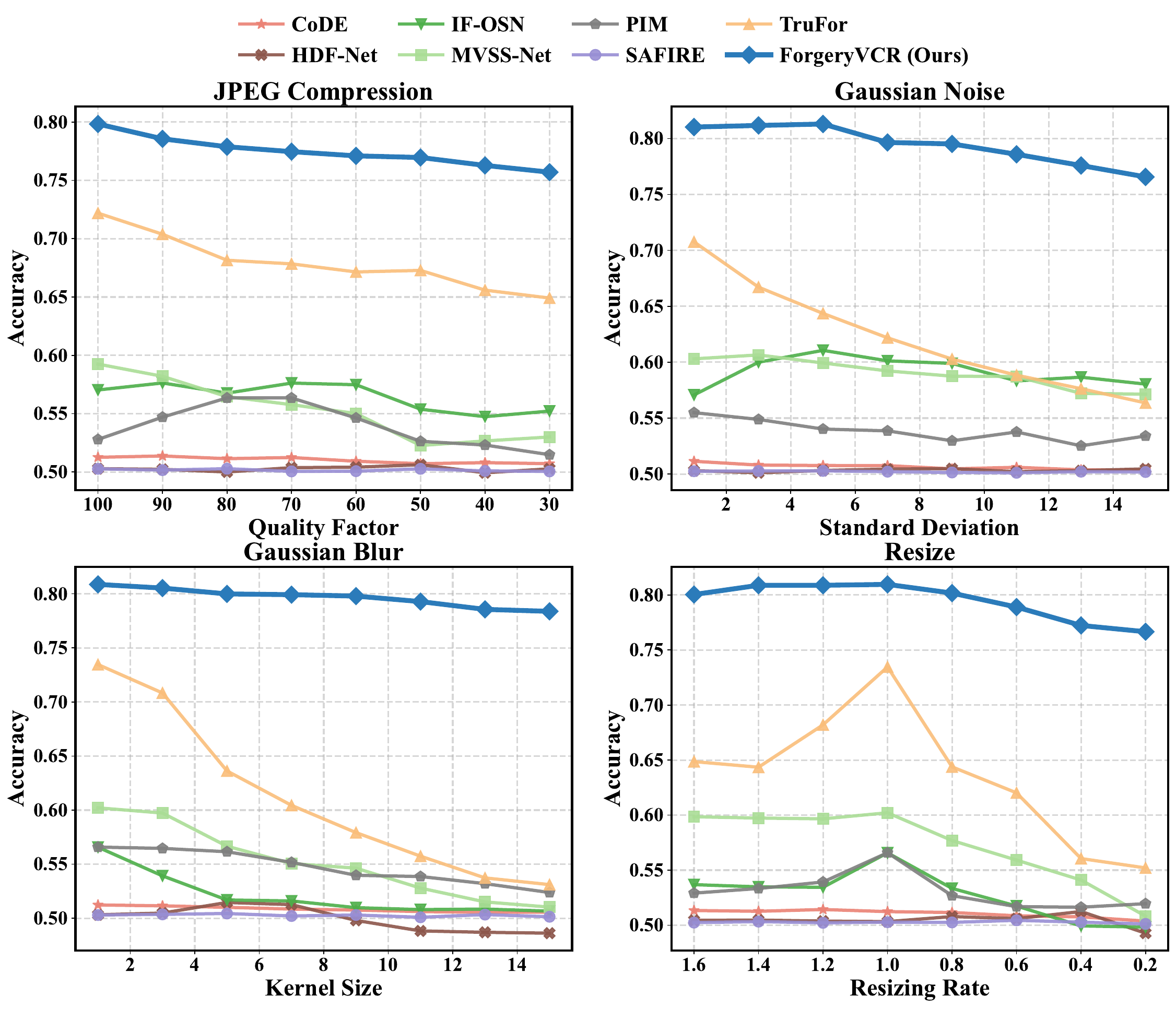}
  \caption{\textbf{Detection Robustness (Accuracy).} Comparison of image-level detection accuracy under varying degradation intensities. ForgeryVCR (Pink) exhibits superior stability compared to specialist networks.}
  \label{fig:robustness_acc}
\end{figure}

\begin{figure}[t!]
  \centering
  \includegraphics[width=\linewidth]{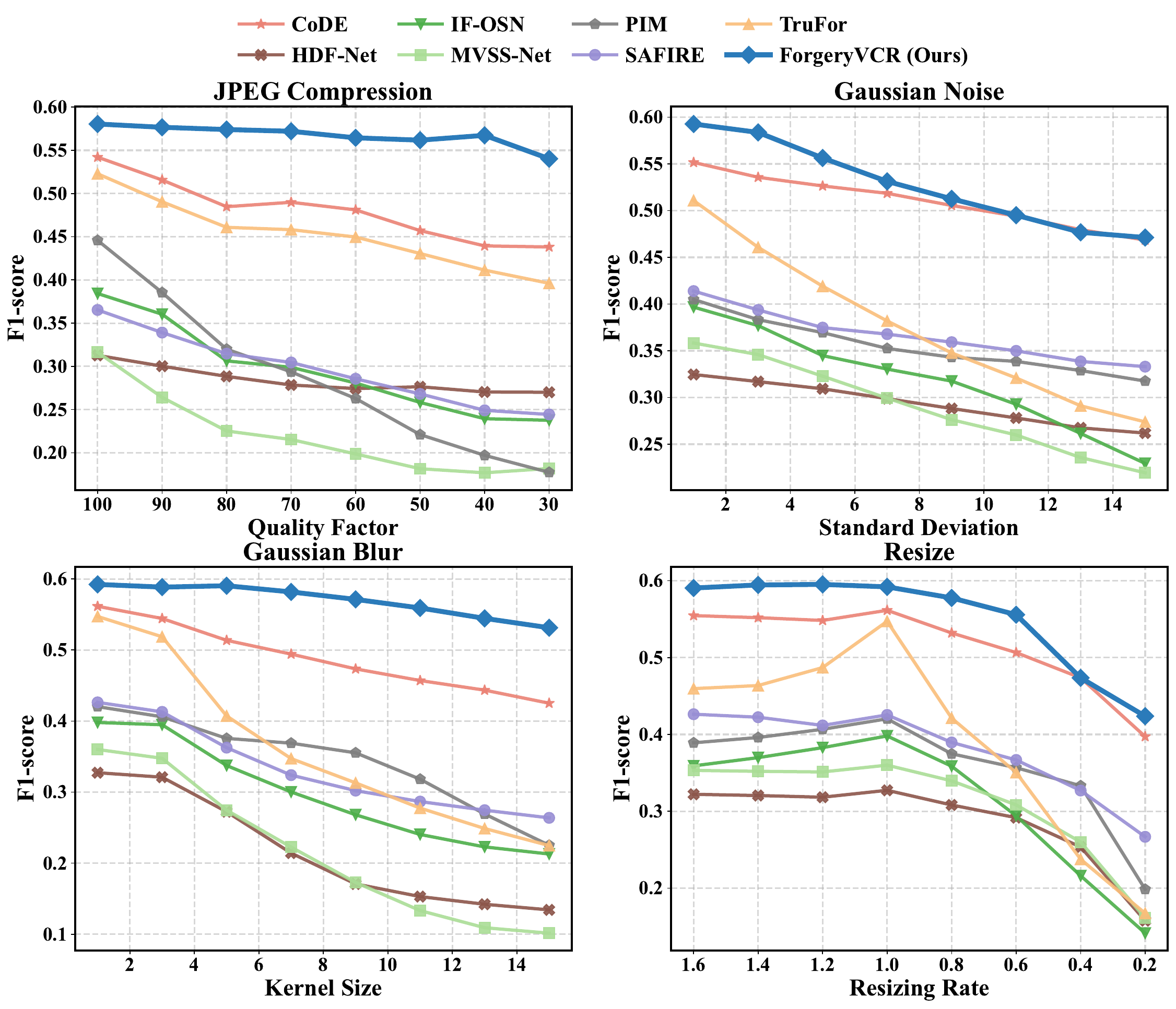}
  \caption{\textbf{Localization Robustness (F1-score).} Comparison of pixel-level localization F1-scores. Our framework maintains precise grounding capabilities even when high-frequency traces are attenuated.}
  \label{fig:robustness_f1}
\end{figure}

To evaluate the stability of ForgeryVCR in real-world scenarios, we conducted extensive robustness tests against common image degradations. We assessed both image-level detection accuracy and pixel-level localization F1-scores under varying intensities of JPEG compression, Gaussian noise, Gaussian blur, and resizing operations. The comparative results against state-of-the-art specialist networks and other baselines are illustrated in Fig.~\ref{fig:robustness_acc} and ~\ref{fig:robustness_f1}.

It is worth noting that the visual comparisons in the aforementioned figures primarily focus on traditional IFDL methods. For MLLM-based forensic approaches such as FakeShield and SIDA, a comprehensive full-scale degradation analysis was constrained by their significantly higher complexity and inference latency. Consequently, we performed a targeted comparison with these methods specifically on the CASIA v1 dataset. Table \ref{tab:mllm_robustness} details the pixel-level localization performance (F1 and IoU) under specific degradation conditions, demonstrating that ForgeryVCR maintains superior robustness even against these multimodal counterparts.

\begin{table}[b]
\centering
\caption{Quantitative robustness comparison with MLLM-based methods on CASIA v1. We report the Pixel-level Localization F1-score and IoU under JPEG compression (Quality Factors 80, 70) and Gaussian degradation (Standard Deviation 5, 10).}
\label{tab:mllm_robustness}
\small
\renewcommand{\arraystretch}{1.2}
\setlength{\tabcolsep}{3pt}
\begin{tabular}{lccccc}
\toprule
\multirow{2}{*}{Method} & \multirow{2}{*}{Metric} & \multicolumn{2}{c}{JPEG Compression} & \multicolumn{2}{c}{Gaussian Noise} \\
 &  & Param=80 & Param=70 & Param=5 & Param=10 \\ \midrule
\multirow{2}{*}{SIDA} & F1 & 0.2774 & 0.2776 & 0.2789 & 0.2611 \\
 & IoU & 0.2333 & 0.2327 & 0.2360 & 0.2195 \\ \midrule
\multirow{2}{*}{FakeShield} & F1 & 0.5647 & 0.5645 & 0.5494 & 0.5297 \\
 & IoU & 0.5026 & 0.5022 & 0.4861 & 0.4693 \\ \midrule
\rowcolor[HTML]{CCE6FF} \cellcolor[HTML]{CCE6FF} & F1 & \textbf{0.7014} & \textbf{0.6973} & \textbf{0.7054} & \textbf{0.6846} \\
\rowcolor[HTML]{CCE6FF} \multirow{-2}{*}{\cellcolor[HTML]{CCE6FF}ForgeryVCR} & IoU & \textbf{0.6771} & \textbf{0.6868} & \textbf{0.6638} & \textbf{0.5981} \\ \bottomrule
\end{tabular}
\end{table}

\begin{figure*}[t!]
  \centering
  \includegraphics[width=0.98\linewidth]{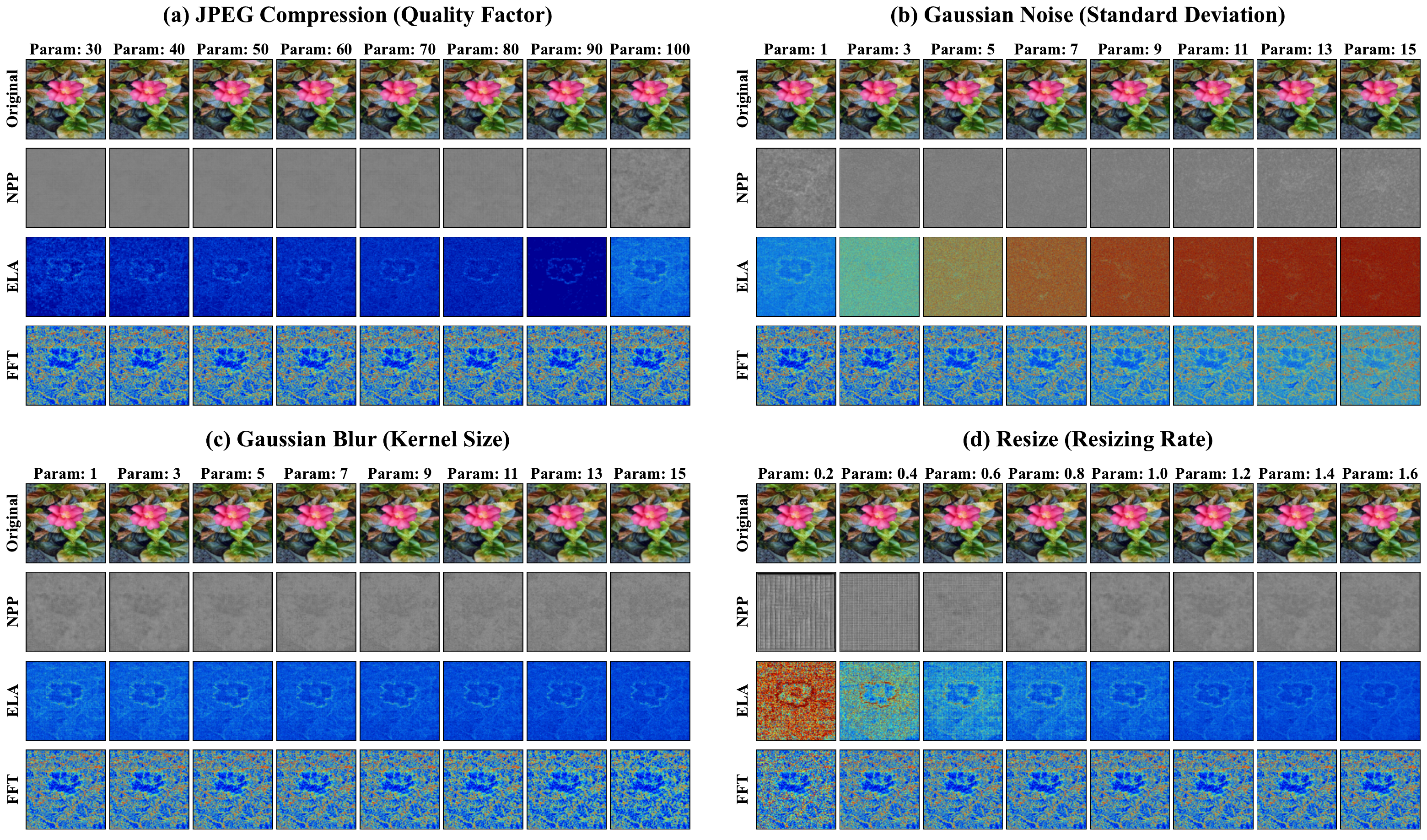}
  \caption{\textbf{Robustness analysis visualization.} To better understand the quantitative stability observed in the charts above, we visualize the actual outputs of our forensic tools under varying degradation intensities.}
  \label{fig:robustness_combined}
\end{figure*}
\subsection{Visual Mechanics of Robustness against Degradations}
\label{app:robustness_degradation_analysis}
As evidenced by the performance curves, ForgeryVCR consistently outperforms competing methods across all distortion types. We analyze the specific performance dynamics for each degradation category below:

To better understand the quantitative stability observed in Fig.~\ref{fig:robustness_combined}, we visualize the actual outputs of our forensic tools under varying degradation intensities. We present these qualitative comparisons in Fig.~\ref{fig:robustness_combined}(a) (JPEG), Fig.~\ref{fig:robustness_combined}(b) (Noise), Fig.~\ref{fig:robustness_combined}(c) (Blur), and Fig.~\ref{fig:robustness_combined}(d) (Resizing). This inspection reveals distinct failure modes and survival mechanisms for each tool, providing a physical explanation for the performance trends.

\noindent \textbf{Resilience Mechanism in JPEG Compression.}
Fig.~\ref{fig:robustness_combined}(a) illustrates the impact of lossy compression. While the distinct high-frequency noise in the \textbf{ELA} map diminishes as the Quality Factor drops (moving left in the figure), the \textbf{structural outline} of the manipulated region (the central flower) remains faintly visible as a coherent block even at lower qualities (QF=50).
\textit{Analysis:} This visual persistence aligns with the JPEG robustness curve in Fig.~\ref{fig:robustness_acc}, where our method maintains stable performance until extreme compression levels. Unlike pixel-level noise which is fragile, the \textit{block-level} artifacts captured by ELA act as a soft indicator that degrades gracefully, allowing the MLLM to sustain localization guidance.

\noindent \textbf{Vulnerability to Gaussian Noise.}
Contrary to the ideal scenario where forensic tools ``filter out'' noise, our visualization in Fig.~\ref{fig:robustness_combined}(b) reveals a critical physical limitation. As the standard deviation of Gaussian noise increases beyond $\sigma=3$, the sensitive camera fingerprints relied upon by \textbf{NPP} are completely overwhelmed, turning the output into a uniform gray field. Similarly, the \textbf{ELA} map becomes dominated by global high-frequency noise.

\textit{Analysis:} Despite this visual collapse of low-level features, the quantitative results in Fig.~\ref{fig:robustness_acc} show that ForgeryVCR (Pink line) maintains a high accuracy ($>0.70$) even at severe noise levels ($\sigma=14$), whereas competitors like TruFor degrade rapidly. This discrepancy suggests that our MLLM \textbf{adaptively shifts its reasoning}. When low-level forensic tools become uninformative (as visually confirmed in Fig.~\ref{fig:robustness_combined}(b), the model pivots to high-level semantic cues (e.g., lighting inconsistencies visible in RGB) or relies on the \textbf{Zoom-In} mechanism to inspect semantic plausibility, rather than blindly trusting the corrupted forensic maps.

\noindent \textbf{Behavior under Gaussian Blur.}
Blurring acts as a low-pass filter, detrimental to edge detection. As shown in Fig.~\ref{fig:robustness_combined}(c), the sharp boundaries in the ELA map become diffuse as the kernel size increases. However, the \textit{positional} information remains intact—the hotspot is blurred but still centered on the forgery.

\textit{Analysis:} This explains why the Localization F1-score (Fig.~\ref{fig:robustness_f1}) declines slowly rather than collapsing. Even if the forensic tool provides a coarse blob rather than a sharp outline, the MLLM utilizes this approximate location to attend to the region in the original image, refining the boundary via semantic segmentation to maintain reasonable performance.

\noindent \textbf{Signal Amplification in Resizing.}
Geometric transformations introduce specific artifacts. Most notably, Fig.~\ref{fig:robustness_combined}(d) reveals that \textbf{upsampling} (Rate $> 1.0$) actually \textit{enhances} forensic signals: the FFT tool captures distinct periodic grid patterns, and ELA shows increased contrast.

\textit{Analysis:} This visualization perfectly corroborates the unique trend observed in the Resize Accuracy curve (Fig.~\ref{fig:robustness_acc}, Bottom Right). While performance drops during downsampling (Rate $<1.0$) due to information loss, it recovers and even peaks during upsampling. The interpolation artifacts effectively turn the degradation into a strong detectable feature, which ForgeryVCR successfully leverages to achieve near-perfect detection in upscaled scenarios.

\section{Qualitative Visualization and Analysis}
\label{app:qualitative_analysis}

\begin{figure*}[!t]
  \centering
  \includegraphics[width=\textwidth]{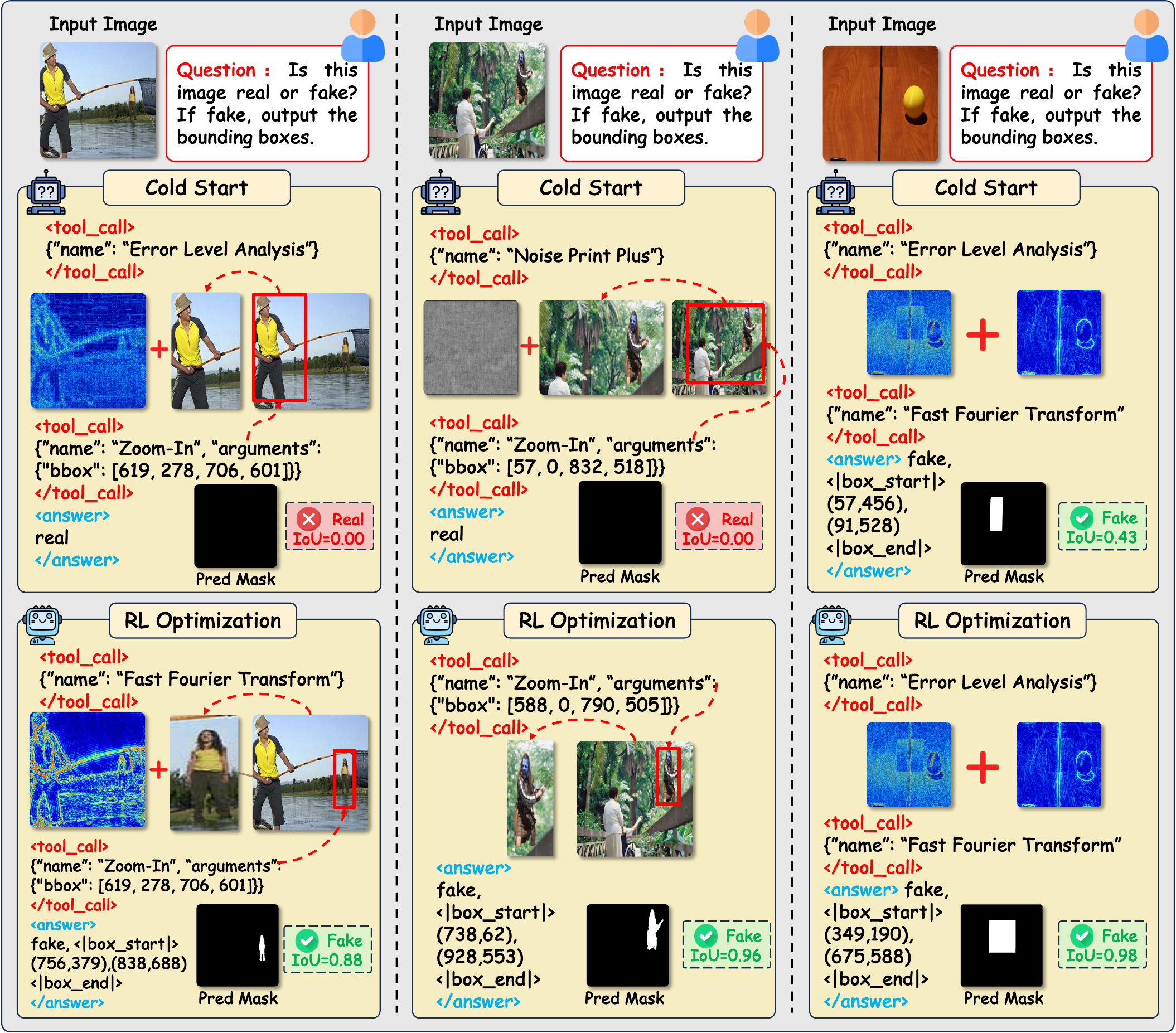}
  \caption{\textbf{Qualitative comparison of policy evolution from Cold Start (SFT) to RL Optimization.} The three columns illustrate distinct improvement behaviors: \textbf{(Left)} correcting ineffective tool selection (switching from ELA to FFT) to fix false negatives; \textbf{(Middle)} pruning redundant tools (removing NPP) to eliminate noise and improve efficiency; and \textbf{(Right)} refining spatial grounding to maximize localization precision using the same forensic cues. The face images in the left and middle figures are sourced from CASIA v1~\cite{Dong2013CASIADatabase}; the original images are from the Corel image dataset (\url{http://corel.digitalriver.com/}).}
  \label{fig:rl_correction}
\end{figure*}

A core motivation of introducing Reinforcement Learning (RL) is to optimize the tool selection policy for both efficiency and forensic accuracy. While SFT establishes the syntactic capability to invoke tools, the model often suffers from indiscriminate tool usage, leading to reasoning paths cluttered with irrelevant or noisy feature maps. In this section, we analyze three distinct evolutionary patterns observed during the transition from SFT to RL, as visualized in Fig. \ref{fig:rl_correction}.

\begin{figure*}[!t]
    \centering
    \includegraphics[width=\linewidth]{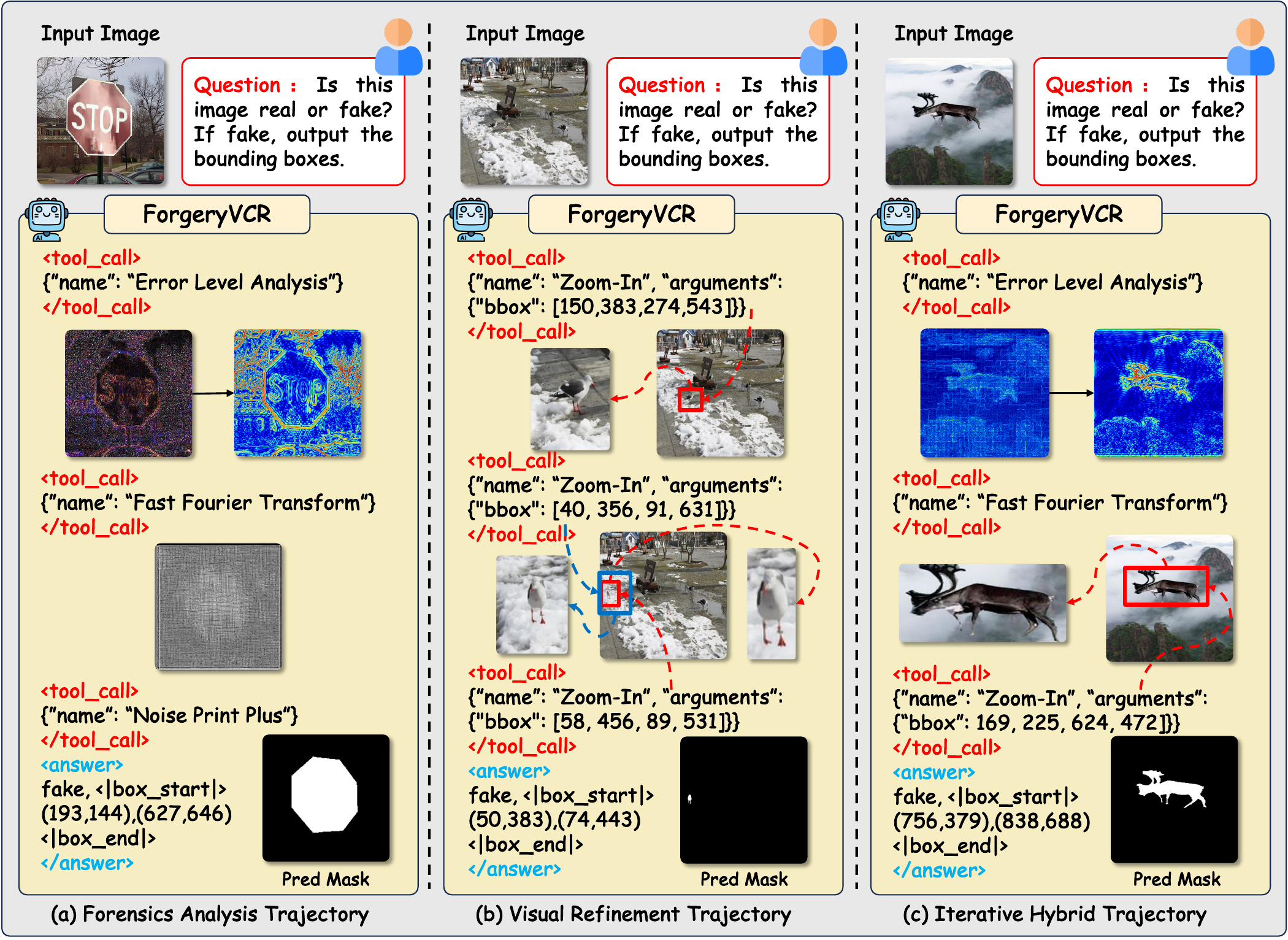}
    \caption{\textbf{Visualization of Visual-Centric Reasoning Trajectories.} We demonstrate how ForgeryVCR adaptively constructs investigation paths based on image content: (a) \textbf{Forensics Analysis Trajectory}, utilizing statistical tools like ELA, FFT, and NPP to expose hidden artifacts; (b) \textbf{Visual Refinement Trajectory}, employing the Zoom-In mechanism for multi-view fine-grained inspection; and (c) \textbf{Iterative Hybrid Trajectory}, which synergizes global forensic cues with local visual verification to generate precise localization masks.}
    \label{fig:vcot_vis}
\end{figure*}

\begin{figure*}[!t]
  \centering
  \includegraphics[width=\textwidth]{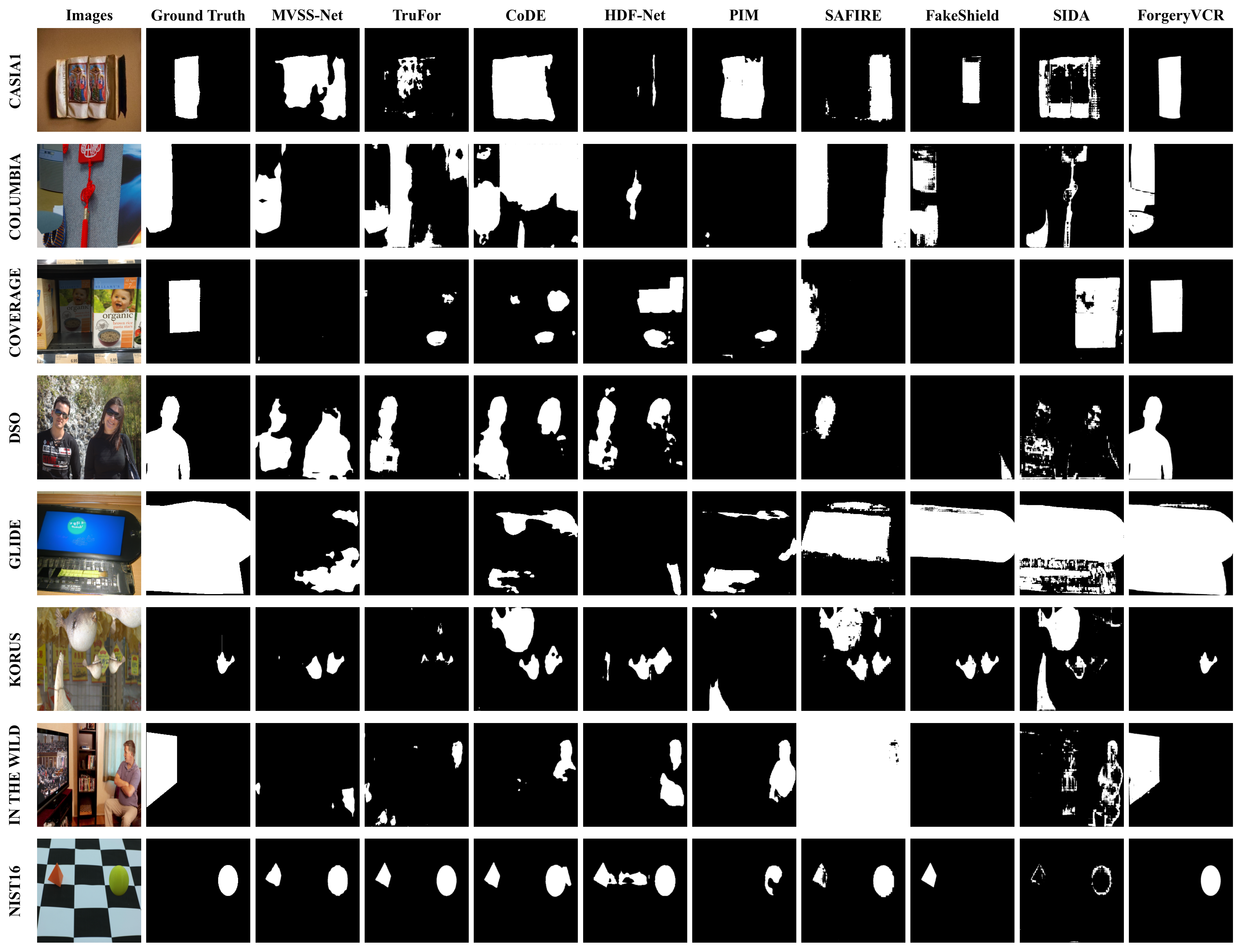}
  \caption{\textbf{Qualitative comparison of pixel-level localization masks across benchmark datasets.} We compare ForgeryVCR with representative baselines including specialist networks and forensic-tuned MLLMs. Our method consistently produces high-fidelity masks that closely align with the Ground Truth (GT), effectively suppressing background noise and accurately delineating manipulated boundaries. Original images source of DSO datasets ~\cite{DeCarvalho2013DSOTIFS} is Flickr (www.flickr.com). }
  \label{fig:mask_comparison}
\end{figure*}

\paragraph{Correction of Tool Selection.}
In the \textbf{Left column} of Fig. \ref{fig:rl_correction}, the SFT baseline initially selects \textit{Error Level Analysis (ELA)}. However, for this specific splicing manipulation, ELA fails to generate a distinct signal against the background, resulting in a False Negative ("Real") verdict. 
In contrast, the RL-optimized policy successfully abandons the ineffective tool and switches to \textit{Fast Fourier Transform (FFT)}. As seen in the generated heatmaps, the FFT visualization reveals a distinct spectral anomaly around the fisherman that was missed by ELA. This demonstrates that the MLLM has learned to strategically invoke the forensic tool that maximizes the signal-to-noise ratio for specific visual contexts, correcting the initial policy to capture valid tampering evidence (IoU 0.88).

\paragraph{Mitigation of Redundant Tool Invocation.}
The \textbf{Middle column} illustrates a case where "less is more." The SFT model employs a complex combination of \textit{Noise Print Plus (NPP)} and \textit{Zoom-In}. However, the NPP map in this specific instance appears perceptually ambiguous (a uniform gray field), which creates noise that confuses the model into an incorrect "Real" classification (IoU 0.00).

Guided by the Tool Utility Reward ($R_{tool}$), the RL-optimized model learns to prune this redundant step. It recognizes that the \textit{Zoom-In} mechanism alone provides sufficient resolution to identify the unnatural boundaries of the spliced figures in the forest. By eliminating the interference from the unnecessary forensic filter, the MLLM not only reduces inference cost but also achieves a correct "Fake" verdict with high localization precision (IoU 0.96).

\paragraph{Refinement of Localization Precision.}
The \textbf{Right column} demonstrates the impact of the Localization Reward ($R_{loc}$). Here, both the SFT and RL models correctly select the same combination of tools—\textit{Error Level Analysis (ELA)} and \textit{Fast Fourier Transform (FFT)}—which successfully highlight the tampering traces on the wooden surface. 
However, the SFT model predicts a loosely defined bounding box (IoU 0.43) that includes significant background area. Driven by the pixel-level alignment objective in the RL stage, the MLLM refines its spatial reasoning. Using the identical forensic evidence, the RL-tuned model generates a much tighter bounding box (IoU 0.98) that closely hugs the contours of the manipulated object. This confirms that RL optimizes not just the \textit{choice} of tools, but also the \textit{interpretation} of their outputs for precise grounding.

\subsection{Visual-Centric Reasoning Visualization}
\label{app:visual_centric_reasoning_visualization}

To intuitively understand the decision-making process of ForgeryVCR, we visualize three representative reasoning trajectories in Fig.~\ref{fig:vcot_vis}. Unlike previous methods that rely on vague textual descriptions, our framework actively invokes tools to materialize imperceptible traces into explicit visual evidence.

As illustrated in Fig.~\ref{fig:vcot_vis}(a), when facing invisible manipulation traces (e.g., the spliced stop sign), the model adopts the Forensics Analysis Trajectory. It sequentially invokes \textit{Error Level Analysis} (ELA), \textit{Fast Fourier Transform} (FFT), and \textit{Noise Print Plus} to reveal statistical inconsistencies that are invisible to the naked eye. 

In contrast, for high-resolution images where the manipulation is small but visually discernible, Fig.~\ref{fig:vcot_vis}(b) shows the Visual Refinement Trajectory. Here, the model bypasses global forensic tools and instead continuously triggers the Zoom-In tool to inspect the suspicious bird regions, progressively narrowing down the bounding box for exact localization.

Finally, Fig.~\ref{fig:vcot_vis}(c) presents the Iterative Hybrid Trajectory, which handles the most complex scenarios. The model first detects spectral anomalies using forensic analysis tools like ELA and FFT, and then verifies by zooming in on the specific region (the deer). This combination ensures that the final prediction is grounded in both statistical and semantic evidence.

\subsection{Qualitative Comparison with SOTA Methods}
\label{app:visual_comparisons}

To visually substantiate the quantitative localization performance reported in the main text, we provide a comprehensive comparison of predicted manipulation masks against state-of-the-art competitors in Fig. \ref{fig:mask_comparison}. The visualization spans diverse benchmarks, covering challenges such as copy-move, splicing, and complex in-the-wild manipulations. As observed in the figure, traditional deep forensic networks often struggle with background noise, producing fragmented masks with high false-positive rates. In contrast, ForgeryVCR leverages high-level semantic reasoning to filter out irrelevant texture variations, resulting in cleaner binary masks. Furthermore, compared to other MLLM-based methods that tend to generate coarse regions, our approach achieves superior boundary adherence by effectively synergizing the MLLM's spatial reasoning with the fine-grained segmentation capability of SAM 2.

\subsection{Failure Case Analysis}
\label{app:failure_cases}

The main limitation lies in fine-grained localization: when tampering leaves minimal forensic traces, the tool maps may produce no discriminative response, resulting in coarser masks. As shown in Fig.~\ref{fig:korus_failure}, in a Korus text-removal case, ForgeryVCR localizes the signboard but misses the removed text. This failure suggests that the model can identify the manipulated object while still struggling to recover subtle text-level boundaries when the available forensic evidence is weak.

\begin{center}
  \centering
  \includegraphics[width=\linewidth]{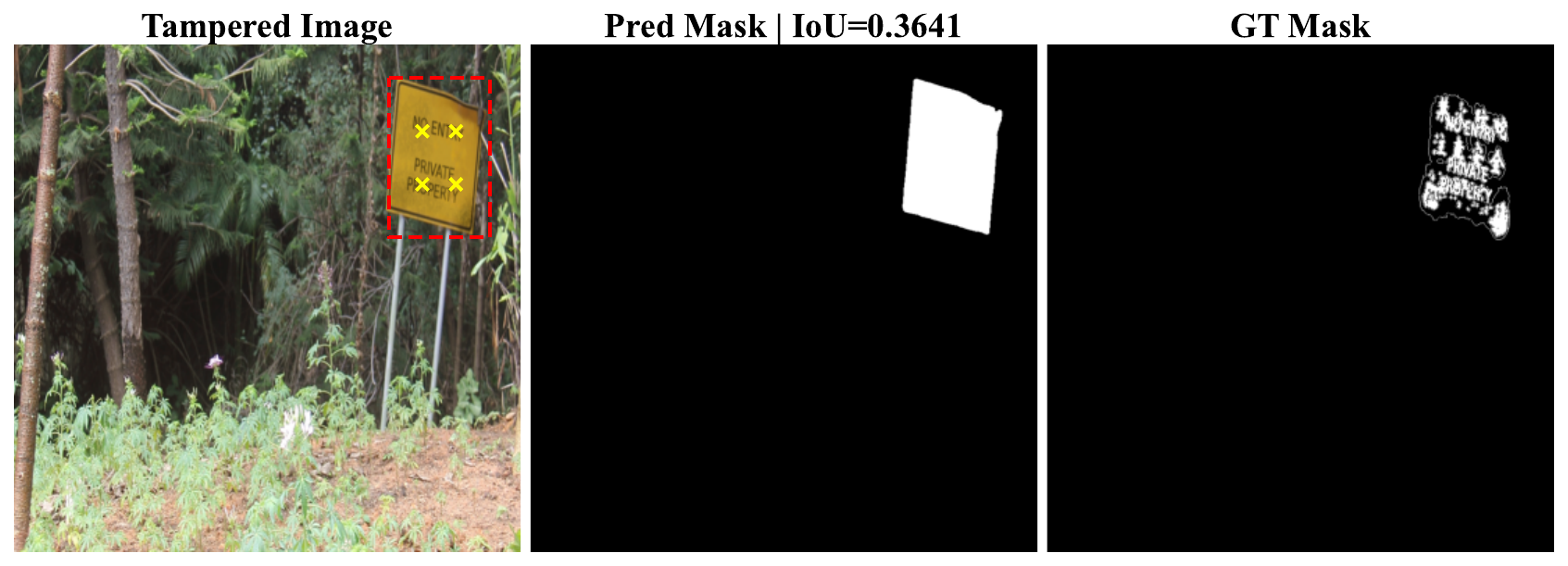}
  \captionof{figure}{Failure case on the Korus text-removal example. ForgeryVCR correctly localizes the signboard but misses the fine-grained removed-text region, resulting in an IoU of 0.3641.}
  \Description{A Korus text-removal failure case showing the tampered image with a signboard, the predicted mask covering the signboard, and the ground-truth mask highlighting the removed text.}
  \label{fig:korus_failure}
\end{center}

\section{Prompt Templates}
\label{app:prompt_templates}

\subsection{System Configuration and User Query Templates}
\label{sec:system_user_prompt}

To make the inference behavior of ForgeryVCR fully reproducible, we report the exact prompts used at deployment time. The system prompt registers the four forensic tools (\textit{Zoom-In}, \textit{ELA}, \textit{FFT}, and \textit{NPP}) as callable functions following the OpenAI tool-calling schema, instructs the model to emit each invocation as a JSON object inside \texttt{<tool\_call>} tags, and specifies the expected output format. The user prompt is an extremely concise instruction that only asks the model to determine authenticity and, if applicable, localize the manipulation, leaving all reasoning and tool-orchestration decisions to the policy learned via SFT and RL. The full \textbf{System Prompt Template} and \textbf{User Prompt Template} are shown in the boxes that follow.
\onecolumn
\subsection{Trajectory Synthesis Templates for CoT Data Generation}
\label{sec:trajectory_synthesis_templates}

To bootstrap the cold-start corpus used by SFT, we synthesize visual--textual reasoning trajectories with three complementary prompt templates, each targeting a distinct manipulation regime. \textbf{Prompt Template~1 (No-Tool Direct Assessment)} generates trajectories in which the assistant resolves the case purely through visual inspection of the original image; this covers conspicuous manipulations where forensic filters are unnecessary. \textbf{Prompt Template~2 (Single-Tool Reasoning)} enforces exactly one forensic tool invocation, producing focused trajectories that teach the model when a single statistical view (\textit{e.g.}, ELA, FFT, NPP, or Zoom-In) is already sufficient. \textbf{Prompt Template~3 (Multi-Tool Synthesis)} orchestrates a chain of tool calls and asks the assistant to reconcile evidence across heatmaps, modeling the iterative ``observe--call--re-observe'' behavior we expect at inference time. Across all templates the ground-truth overlay serves only as a hidden source of truth: the generated dialogues are constrained to never reference the overlay or address images by index, so the resulting trajectories read as natural expert investigations.

\subsection{LLM-as-a-Judge Evaluation Prompts for Hallucination Analysis}
\label{sec:hallucination_judge_prompt}

To quantify semantic hallucinations in the reasoning text of ForgeryVCR$^*$, we employ Qwen3-VL-32B-Instruct as a multimodal judge. The judge evaluates each reasoning chain across three binary dimensions: (1) \textbf{original\_visual\_grounded} — whether the CoT's visual descriptions about the original image are genuinely supported by the image; (2) \textbf{tool\_visual\_grounded} — whether the CoT's descriptions about the forensic tool maps are supported by the actual tool-map images; and (3) \textbf{final\_consistency} — whether the reasoning text is consistent with the model's final prediction and localization. The judge is provided with the original image, the forensic tool-map images, the full assistant reasoning text, and metadata about the prediction. A sample is flagged as hallucinated if any of the three scores is 0. The correction rate reported in the main paper is defined as the fraction of hallucinated samples (as judged on ForgeryVCR$^*$) where ForgeryVCR produces the correct classification while ForgeryVCR$^*$ does not. The complete \textbf{LLM-as-a-Judge Prompt Template} is shown below, after all CoT-synthesis templates.

\subsection{Data Structures for Lightweight Specialist Models}
\label{sec:specialist_data_structures}

For completeness, we also include the JSON-style data formats used to fine-tune the lightweight specialist models introduced in \S\ref{app:trajectory_synthesis_pipeline}: \textbf{Data Structure~1} corresponds to the Tool-Free Baseline, which conditions only on the original image, while \textbf{Data Structure~2} corresponds to a representative Single-Tool Expert (illustrated with the FFT high-frequency heatmap) and pairs the original image with one forensic visualization plus an expert-guided instruction. Both data structures are presented in the boxes that follow, after all CoT-synthesis and judging templates.

\medskip
\noindent\textit{All prompt templates and data structures referenced in \S\ref{sec:system_user_prompt}--\S\ref{sec:specialist_data_structures} are presented in boxes on the following pages, in the order: System Prompt, User Prompt, Synthesis Templates~1--3, the LLM-as-a-Judge Prompt, and finally Data Structures~1--2.}


\begin{tcolorbox}[
  enhanced,
  breakable,
  colback=gray!5, 
  colframe=black, 
  title=\textbf{System Prompt Template},
  title after break=\textbf{System Prompt Template (Continued)},
  fonttitle=\bfseries\small, 
  left=2mm, right=2mm, top=2mm, bottom=2mm,
  lines before break=2
]
\noindent You are a helpful assistant. \\
\\
\# Tools \\
\\
You may call one or more functions to assist with the user query. \\
\\
You are provided with function signatures within \textless tools\textgreater \textless/tools\textgreater XML tags: \\
\textless tools\textgreater \\
\noindent [ \\
\{ \\
\hspace*{1em} ``type'': ``function'', \\
\hspace*{1em} ``function'': \{ \\
\hspace*{2em} ``name'': ``zoom\_in'', \\
\hspace*{2em} ``description'': ``Zooms in on a suspicious region to check for fine-grained manipulation artifacts like inconsistent textures or noise.'', \\
\hspace*{2em} ``parameters'': \{ \\
\hspace*{3em} ``type'': ``object'', \\
\hspace*{3em} ``properties'': \{ \\
\hspace*{4em} ``bbox'': \{ \\
\hspace*{5em} ``type'': ``array'', \\
\hspace*{5em} ``items'': \{``type'': ``integer''\}, \\
\hspace*{5em} ``description'': ``The bounding box coordinates [x1, y1, x2, y2] of the region to zoom in.'' \\
\hspace*{4em} \} \\
\hspace*{3em} \}, \\
\hspace*{3em} ``required'': [``bbox''] \\
\hspace*{2em} \} \\
\hspace*{1em} \} \\
\}, \\
\{ \\
\hspace*{1em} ``type'': ``function'', \\
\hspace*{1em} ``function'': \{ \\
\hspace*{2em} ``name'': ``ELA'', \\
\hspace*{2em} ``description'': ``Performs Error Level Analysis to reveal inconsistencies in the image's compression levels. Tampered regions often exhibit distinct error levels.'', \\
\hspace*{2em} ``parameters'': \{ \\
\hspace*{3em} ``type'': ``object'', \\
\hspace*{3em} ``properties'': \{\}, \\
\hspace*{3em} ``required'': [] \\
\hspace*{2em} \} \\
\hspace*{1em} \} \\
\}, \\
\{ \\
\hspace*{1em} ``type'': ``function'', \\
\hspace*{1em} ``function'': \{ \\
\hspace*{2em} ``name'': ``FFT'', \\
\hspace*{2em} ``description'': ``Analyzes the image's frequency domain using Fast Fourier Transform. Tampering can introduce periodic artifacts or disrupt natural frequency patterns.'', \\
\hspace*{2em} ``parameters'': \{ \\
\hspace*{3em} ``type'': ``object'', \\
\hspace*{3em} ``properties'': \{\}, \\
\hspace*{3em} ``required'': [] \\
\hspace*{2em} \} \\
\hspace*{1em} \} \\
\}, \\
\{ \\
\hspace*{1em} ``type'': ``function'', \\
\hspace*{1em} ``function'': \{ \\
\hspace*{2em} ``name'': ``NPP'', \\
\hspace*{2em} ``description'': ``Analyzes the image's noise fingerprints using Noise Print Pattern. Tampered regions often show inconsistent noise variance compared to the authentic background.'', \\
\hspace*{2em} ``parameters'': \{ \\
\hspace*{3em} ``type'': ``object'', \\
\hspace*{3em} ``properties'': \{\}, \\
\hspace*{3em} ``required'': [] \\
\hspace*{2em} \} \\
\hspace*{1em} \} \\
\} \\
] \\
\textless/tools\textgreater \\
\\
For each function call, return a json object with function name and arguments within \textless tool\_call\textgreater \textless/tool\_call\textgreater XML tags: \\
\textless tool\_call\textgreater \\
\{``name'': \textless function-name\textgreater, ``arguments'': \textless args-json-object\textgreater\} \\
\textless/tool\_call\textgreater
\end{tcolorbox}

\vspace{0.2cm}

\begin{tcolorbox}[
    enhanced,
    colback=gray!5, 
    colframe=black, 
    title=\textbf{User Prompt Template},
    fonttitle=\bfseries\small, 
    left=2mm, right=2mm, top=2mm, bottom=2mm
]
\noindent \textless image\textgreater Determine if this image is real or fake. If manipulation is found, highlight the tampered regions with bounding boxes.
\end{tcolorbox}

\vspace{0.2cm}

\begin{tcolorbox}[
  enhanced,
  breakable,
  colback=gray!5, 
  colframe=black, 
  title=\textbf{Prompt Template 1: No-Tool Direct Assessment},
  title after break=\textbf{Prompt Template 1: (Continued)},
  fonttitle=\bfseries\small, 
  left=2mm, right=2mm, top=2mm, bottom=2mm,
  lines before break=2,
  parbox=false
]
\noindent {[}BEGIN OF GOAL{]} \\
You are an expert visual assistant specializing in digital image forensics. Your goal is to generate a plausible step-by-step reasoning trajectory to identify tampered regions within an image. The trajectory should be presented as a single-turn dialogue between a ``User'' and an ``Assistant'', where the Assistant performs a comprehensive visual analysis and provides a final conclusion without the need for external tools. \\
{[}END OF GOAL{]} \\
\\
{[}BEGIN OF ACTIONS{]} \\
Name: Terminate \\
Description: Concludes the analysis and provides the final answer. This action must be called at the end of the trajectory. \\
Examples: \\
Tool Called: Terminate \\
{[}END OF ACTIONS{]} \\
\\
{[}BEGIN OF TASK INSTRUCTIONS{]} \\
You are given two images: the original image and a ground truth overlay showing the exact tampered region. The overlay is your absolute source of truth for your conclusion.

Your task is to generate a single-round dialogue. \\
1. The ``User'' will ask for an analysis. \\
2. Your ``Assistant'' response must be a comprehensive analysis of the original image. \\
3. Your reasoning MUST focus on the specific object highlighted in the ground truth overlay. Describe visual inconsistencies on that object (e.g., unnatural texture, mismatched lighting, sharp edges, incorrect shadows). \\
4. Your speech must conclude with a precise final verdict, accurately identifying the tampered object and describing its location textually (e.g., ``in the bottom-left corner'', ``on top of the building''). \\
5. Finally, call the ``Terminate'' action. \\
6. \textbf{Crucially, your entire output must not mention the ground truth overlay or refer to images by number (e.g., [Image 2]). You are simulating an expert's direct analysis.} \\
7. \textbf{Your final verdict MUST NOT contain any bounding box coordinates or the ``\textless bbox\textgreater'' placeholder.} \\
{[}END OF TASK INSTRUCTIONS{]} \\
\\
{[}BEGIN OF OUTPUT FORMAT{]} \\
Your output must be a single block starting with ``{[}BEGIN OF SOLUTION{]}'' and ending with ``{[}END OF SOLUTION{]}''. The dialogue must be in a single round.

{[}BEGIN OF SOLUTION{]} \\
\# Round 1 \\
User: ``Please analyze the provided image for any signs of tampering. If you find any, explain why and identify the manipulated areas. If not, please confirm its authenticity and describe your analysis.'' \\
Assistant: ``[Your detailed analysis and final verdict here]'' \\
Tool Called: Terminate \\
{[}END OF SOLUTION{]} \\
{[}END OF OUTPUT FORMAT{]} \\
\\
{[}BEGIN OF EXAMPLES{]} \\
{[}BEGIN OF SOLUTION{]} \\
\# Round 1 \\
User: ``Please analyze the provided image for any signs of tampering. If you find any, explain why and identify the manipulated areas. If not, please confirm its authenticity and describe your analysis.'' \\
Assistant: ``I am analyzing a photograph showing a group of monkeys on a stone wall. Upon close inspection, the monkey located in the bottom-left corner exhibits clear signs of digital manipulation. Its internal texture is unnaturally smooth and lacks the fine fur detail present on the other monkeys. Furthermore, the lighting on this monkey is flat and inconsistent with the ambient light of the scene, and it casts no discernible shadow on the wall beneath it. My analysis is complete. The image is tampered, and the manipulated object is the monkey in the bottom-left of the image.'' \\
Tool Called: Terminate \\
{[}END OF SOLUTION{]} \\
{[}END OF EXAMPLES{]}
\end{tcolorbox}

\begin{tcolorbox}[
  enhanced,
  breakable,
  colback=gray!5, 
  colframe=black, 
  title=\textbf{Prompt Template 2: Single-Tool Analysis (e.g., ELA)},
  title after break=\textbf{Prompt Template 2: (Continued)},
  fonttitle=\bfseries\small, 
  left=2mm, right=2mm, top=2mm, bottom=2mm,
  lines before break=2,
  parbox=false
]
\noindent {[}BEGIN OF GOAL{]} \\
You are an expert visual assistant specializing in digital image forensics. Your goal is to generate a plausible step-by-step reasoning trajectory to identify tampered regions within an image. The core of your analysis depends on identifying \textbf{inconsistencies} rather than just visual anomalies. The final output should be a multi-turn dialogue between a ``User'' and an ``Assistant''. \\
{[}END OF GOAL{]} \\
\\
{[}BEGIN OF ACTIONS{]} \\
Name: ELA \\
Description: Generates an Error Level Analysis (ELA) heatmap to reveal discrepancies in compression history. \\
\textbf{Crucially, do not assume that bright areas are automatically tampered.} \\
\hspace*{1em}* \textbf{Normal Behavior:} Areas with complex textures (e.g., foliage, gravel, fur) naturally exhibit higher error levels (brighter noise) in ELA. Smooth areas (e.g., sky, skin) usually appear dark. \\
\hspace*{1em}* \textbf{Tampering Indicators:} Look for \textbf{mismatches}. A tampered region often breaks the continuity of the noise pattern. For example, a visually smooth object appearing suspiciously bright in ELA, or an object exhibiting a distinct noise texture (e.g., different color distribution or density) that does not match the surrounding environment's error profile.

Name: Terminate \\
Description: Concludes the analysis and provides the final answer. This action must be called at the end of every trajectory. \\
{[}END OF ACTIONS{]} \\
\\
{[}BEGIN OF TASK INSTRUCTIONS{]} \\
You are given three images: the original image, the ground truth overlay (your absolute source of truth), and an ELA heatmap.

Your task is to generate a two-round dialogue. \\
1. \textbf{Round 1:} The ``Assistant'' analyzes the \textbf{original image only}. Focus on the object highlighted in the ground truth. Describe visual suspicions regarding \textbf{consistency}: Does the object's lighting, resolution, or grain match its surroundings? (e.g., ``The object seems to have a different texture quality compared to the background''). Conclude by calling ``ELA''. \\
2. \textbf{Round 2:} The ``Assistant'' analyzes the \textbf{ELA heatmap}. \\
\hspace*{2em}* \textbf{Contextual Comparison:} Compare the suspicious area's ELA signature against the rest of the image. \\
\hspace*{2em}* \textbf{Reasoning:} Explain \emph{why} the ELA pattern is suspicious. Is it brighter than it \emph{should} be given its visual texture? Does the noise pattern look ``pasted'' on top of the background noise? \\
\hspace*{2em}* \textbf{Avoid False Positives:} Acknowledge that natural textures are noisy, but highlight that the tampered object has a \emph{distinct} or \emph{alien} noise profile compared to valid high-frequency areas. \\
\hspace*{2em}* Integrate findings to form a conclusion and call ``Terminate''. \\
3. \textbf{Crucially, do not mention the ground truth overlay or refer to images by number.} \\
4. \textbf{Your final verdict MUST NOT contain any bounding box coordinates or the ``\textless bbox\textgreater'' placeholder.} \\
{[}END OF TASK INSTRUCTIONS{]} \\
\\
{[}BEGIN OF OUTPUT FORMAT{]} \\
Your output must be a single block starting with ``{[}BEGIN OF SOLUTION{]}'' and ending with ``{[}END OF SOLUTION{]}''. The dialogue must be in two rounds.

{[}BEGIN OF SOLUTION{]} \\
\# Round 1 \\
User: ``Is this image authentic? Leave no stone unturned.'' \\
Assistant: ``[Analysis of the original image, focusing on visual inconsistencies like lighting or edge artifacts]'' \\
Tool Called: ELA

\# Round 2 \\
Assistant: ``[Analysis of the ELA heatmap. Explicitly discusses the ``mismatch'' between visual texture and error level. Conclusion describes the tampered object]'' \\
Tool Called: Terminate \\
{[}END OF SOLUTION{]} \\
{[}END OF OUTPUT FORMAT{]} \\
\\
{[}BEGIN OF EXAMPLES{]} \\
{[}BEGIN OF SOLUTION{]} \\
\# Round 1 \\
User: ``Is this image authentic? Leave no stone unturned.'' \\
Assistant: ``I am examining an image of a grassy field with a zebra standing in the center. While the scene looks mostly natural, I notice the zebra's outline seems slightly too sharp against the background grass, and the lighting on its coat doesn't perfectly match the ambient direction. This raises a suspicion that the zebra might have been inserted. I will use ELA to check for compression inconsistencies.'' \\
Tool Called: ELA

\# Round 2 \\
Assistant: ``Analyzing the ELA heatmap, I see significant noise across the entire image, which is expected given the complex texture of the grass. However, the area corresponding to the zebra exhibits a \textbf{distinctly different noise intensity and color distribution} compared to the surrounding grass. While the grass shows a uniform high-frequency scatter, the zebra stands out as a solid block of hyper-intense artifacts that doesn't blend with the environment's noise profile. This \textbf{inconsistency in error levels} confirms that the zebra originates from a different source and was pasted into this scene. The image is tampered.'' \\
Tool Called: Terminate \\
{[}END OF SOLUTION{]} \\
{[}END OF EXAMPLES{]}
\end{tcolorbox}

\begin{tcolorbox}[
  enhanced,
  breakable,
  colback=gray!5, 
  colframe=black, 
  title=\textbf{Prompt Template 3: Multi-Tool Synthesis},
  title after break=\textbf{Prompt Template 3: Multi-Tool Synthesis (Continued)},
  fonttitle=\bfseries\small, 
  left=2mm, right=2mm, top=2mm, bottom=2mm,
  lines before break=2,
  parbox=false
]
\noindent {[}BEGIN OF GOAL{]} \\
You are an expert visual assistant specializing in digital image forensics. Your goal is to generate a plausible step-by-step reasoning trajectory to identify a \textbf{fake} (manipulated) object within an image. You will conduct a comprehensive forensic investigation using visual analysis and three specialized tools (\textbf{\{t1\}}, \textbf{\{t2\}}, and \textbf{\{t3\}}) to build an unshakeable chain of evidence. \\
{[}END OF GOAL{]} \\
\\
{[}BEGIN OF ACTIONS{]} \\
Name: NPP \\
Description: Analyzes noise patterns. Fakes appear as disrupted noise fingerprints. \\
Name: ELA \\
Description: ELA Heatmap. Fakes appear as compression anomalies (unexplained brightness). \\
Name: FFT \\
Description: High-Frequency Energy Map. Fakes appear as frequency mismatches (color contrast islands). \\
Name: Terminate \\
Description: Concludes the analysis. Must be called at the end. \\
{[}END OF ACTIONS{]} \\
\\
{[}BEGIN OF TASK INSTRUCTIONS{]} \\
You are given four inputs: \\
1. The \textbf{Original Image}. \\
2. The \textbf{Ground Truth Overlay} (This is your SECRET source of truth. It highlights the exact manipulated region). \\
3. The \textbf{\{t1\} Map}. \\
4. The \textbf{\{t2\} Map}. \\
5. The \textbf{\{t3\} Map}.

Your task is to generate a \textbf{four-round dialogue} to expose the forgery.

\textbf{CRITICAL RULE: You must NEVER mention the ``Ground Truth Overlay'' or ``Mask'' in your output. You must pretend you discovered the suspicious area through your own visual and forensic analysis.}

1. \textbf{Round 1 (Visual Suspicion):} \\
\hspace*{2em}* \textbf{Internal Step:} Look at the Ground Truth Overlay to identify the specific object/region. \\
\hspace*{2em}* \textbf{External Output:} Analyze the \textbf{Original Image}. Focus your ``suspicion'' on that specific object. Describe visual inconsistencies (e.g., lighting, sharpness, grain). \\
\hspace*{2em}* \textbf{Action:} Call ``\{t1\}''.

2. \textbf{Round 2 (First Verification - \{t1\}):} \\
\hspace*{2em}* Analyze the \textbf{\{t1\} Map} focusing on the suspicious object. \\
\hspace*{2em}* \textbf{Analysis Guide:} \\
\hspace*{4em}\{t1\_guide\} \\
\hspace*{2em}* \textbf{Requirement:} Describe the anomaly (e.g., ``The map shows a distinct [feature] on the [object]...''). \\
\hspace*{2em}* \textbf{Creative Transition:} State that this is a strong indicator of manipulation. Generate a \textbf{diverse transitional phrase} to introduce the next tool. \\
\hspace*{4em}* \emph{Instruction:} \textbf{Avoid repetitive phrasing.} (e.g., ``To investigate this inconsistency further...'', ``Checking the spectral/compression domain with...'', ``Let's cross-reference this artifact...''). \\
\hspace*{2em}* \textbf{Action:} Call ``\{t2\}''.

3. \textbf{Round 3 (Second Verification - \{t2\}):} \\
\hspace*{2em}* Analyze the \textbf{\{t2\} Map} focusing on the same object. \\
\hspace*{2em}* \textbf{Analysis Guide:} \\
\hspace*{4em}\{t2\_guide\} \\
\hspace*{2em}* \textbf{Requirement:} Confirm the anomaly persists in this domain as well. Describe the specific heatmap/noise features. \\
\hspace*{2em}* \textbf{Creative Transition:} Generate a \textbf{varied closing transition} to the final tool. \\
\hspace*{4em}* \emph{Instruction:} \textbf{Be creative.} (e.g., ``For a definitive conclusion...'', ``To close the evidentiary loop...'', ``One final verification to confirm the forgery...''). \\
\hspace*{2em}* \textbf{Action:} Call ``\{t3\}''.

4. \textbf{Round 4 (Final Verification - \{t3\} \& Verdict):} \\
\hspace*{2em}* Analyze the \textbf{\{t3\} Map} focusing on the same object. \\
\hspace*{2em}* \textbf{Analysis Guide:} \\
\hspace*{4em}\{t3\_guide\} \\
\hspace*{2em}* \textbf{Requirement:} Confirm that \{t3\} also highlights the same object as anomalous. \\
\hspace*{2em}* \textbf{Verdict:} Synthesize all evidences. State clearly: ``The image is fake.'' Provide a final descriptive verdict identifying the \textbf{object} and its \textbf{location} (e.g., ``The image is fake, and the manipulated object is the car on the left''). \\
\hspace*{2em}* \textbf{Action:} Call ``Terminate''.

5. \textbf{Formatting:} Do not refer to images by number. Simulate a professional workflow. \\
{[}END OF TASK INSTRUCTIONS{]} \\
\\
{[}BEGIN OF OUTPUT FORMAT{]} \\
Your output must be a single block starting with ``{[}BEGIN OF SOLUTION{]}'' and ending with ``{[}END OF SOLUTION{]}''.

{[}BEGIN OF SOLUTION{]} \\
\# Round 1 \\
User: ``Is this image real? Leave no stone unturned.'' \\
Assistant: ``[Visual analysis raising suspicion about the specific object found in the overlay...]'' \\
Tool Called: \{t1\}

\# Round 2 \\
Assistant: ``[Analysis of \{t1\} showing anomaly... Varied transition to \{t2\}]'' \\
Tool Called: \{t2\}

\# Round 3 \\
Assistant: ``[Analysis of \{t2\} showing anomaly... Varied transition to \{t3\}]'' \\
Tool Called: \{t3\}

\# Round 4 \\
Assistant: ``[Analysis of \{t3\} showing anomaly... Final Descriptive Verdict: The image is fake...]'' \\
Tool Called: Terminate \\
{[}END OF SOLUTION{]} \\
{[}END OF OUTPUT FORMAT{]}
\end{tcolorbox}

\vspace{0.2cm}

\begin{tcolorbox}[
  title={LLM-as-a-Judge Prompt Template},
  breakable,
  colback=gray!5,
  colframe=black!60,
  fonttitle=\bfseries\small,
  fontlower=\small
]
\small
You are a strict multimodal hallucination judge for image forensics reasoning.

You will see:
\begin{enumerate}
  \item The original image.
  \item Zero or more forensic tool-map images used by the CoT model.
  \item The full assistant reasoning text from the CoT model.
  \item The model's final parsed prediction metadata.
\end{enumerate}

Judge the CoT reasoning with three binary scores only:

\textbf{1. final\_consistency:}
Return 1 if the reasoning text is consistent with the model's final prediction and predicted localization region.
Return 0 if the reasoning text discusses a different object/region/relative position than the final prediction, or if the final conclusion contradicts the reasoning.

\textbf{2. original\_visual\_grounded:}
Return 1 if the CoT's visual descriptions about the ORIGINAL image are visibly supported by the original image.
Return 0 if those descriptions appear fabricated, unsupported, or clearly inconsistent with the image evidence.

\textbf{3. tool\_visual\_grounded:}
Return 1 if the CoT's visual descriptions about the TOOL MAPS are supported by the provided tool-map images.
Return 0 if those descriptions are fabricated, unsupported, or inconsistent with the actual tool maps.

\textbf{Important:}
\begin{itemize}
  \item Judge grounding against visual evidence, not against ground truth correctness.
  \item A wrong prediction can still be grounded if the described evidence is genuinely visible.
  \item Be conservative: use 0 if the claim is not clearly supported.
  \item Output valid JSON only.
\end{itemize}

\textbf{Case metadata:} Dataset, image path, GT tampered status, GT bounding boxes, CoT predicted fake status, CoT predicted boxes, CoT IoU, and the list of CoT tool images and full assistant messages are provided inline.

\textbf{Return JSON with exactly these keys:}
\begin{verbatim}
{
  "final_consistency": 0 or 1,
  "original_visual_grounded": 0 or 1,
  "tool_visual_grounded": 0 or 1,
  "brief_reason": "one short sentence"
}
\end{verbatim}
\end{tcolorbox}

\vspace{0.2cm}

\begin{tcolorbox}[
  enhanced,
  breakable,
  colback=gray!5,
  colframe=black,
  title=\textbf{Data Structure 1: Tool-Free Baseline},
  title after break=\textbf{Data Structure 1 (Continued)},
  fonttitle=\bfseries\small,
  left=2mm, right=2mm, top=2mm, bottom=2mm,
  lines before break=2
]
\noindent \{ \\
\hspace*{1em} ``role'': ``user'', \\
\hspace*{1em} ``content'': ``\textless image\textgreater What is the status of this image, real or fake? If you detect manipulation, please provide the bounding boxes.'' \\
\}, \\
\{ \\
\hspace*{1em} ``role'': ``assistant'', \\
\hspace*{1em} ``content'': ``\textless answer\textgreater real\textless/answer\textgreater'' \\
\}
\end{tcolorbox}

\vspace{0.2cm}

\begin{tcolorbox}[
  enhanced,
  breakable,
  colback=gray!5,
  colframe=black,
  title=\textbf{Data Structure 2: Single-Tool Expert (e.g., FFT)},
  title after break=\textbf{Data Structure 2 (Continued)},
  fonttitle=\bfseries\small,
  left=2mm, right=2mm, top=2mm, bottom=2mm,
  lines before break=2
]
\noindent \{ \\
\hspace*{1em} ``role'': ``user'', \\
\hspace*{1em} ``content'': ``\textless image\textgreater\textless image\textgreater The first image is the original input, and the second is its FFT High-Frequency heatmap. In the heatmap, red or bright areas represent strong high-frequency energy... [Tool Analysis Instructions]... If the image is authentic, output \textless answer\textgreater real\textless/answer\textgreater. If manipulated, detect and output fake and the bounding box...'' \\
\}, \\
\{ \\
\hspace*{1em} ``role'': ``assistant'', \\
\hspace*{1em} ``content'': ``\textless answer\textgreater fake, \textless\textbar box\_start\textbar\textgreater x\_min,y\_min,x\_max,y\_max\textless\textbar box\_end\textbar\textgreater\textless/answer\textgreater'' \\
\}
\end{tcolorbox}
\twocolumn

\end{document}